\documentclass[10pt]{article}
\usepackage[margin=1in]{geometry}

\usepackage[utf8]{inputenc} 
\usepackage[T1]{fontenc}    
\usepackage{hyperref,graphicx,subcaption,lipsum}       
\usepackage{url}            
\usepackage{booktabs}       
\usepackage{amsfonts}       
\usepackage{nicefrac}       
\usepackage{microtype}      
\usepackage{xcolor}         
\usepackage{amssymb,amsfonts,amsmath,amsthm,amscd,dsfont,mathrsfs,bbold,pifont}
\usepackage{graphicx}
\usepackage{comment}
\usepackage{caption, subcaption, float}
\usepackage{multirow}
\usepackage{enumitem}
\usepackage{array}
\usepackage{footmisc}

\usepackage{enumitem}
\usepackage{bbm}
\usepackage{amsthm}
\usepackage{hyperref}
\usepackage{url}
\usepackage{graphicx}
\usepackage{caption}
\usepackage{subcaption}
\usepackage{booktabs}
\usepackage{multirow}
\usepackage{wrapfig}
\usepackage{makecell}
\usepackage{natbib}

\newcolumntype{P}[1]{>{\centering\arraybackslash}p{#1}}

\definecolor{mydarkblue}{rgb}{0,0.08,0.60}
  \hypersetup{ %
    colorlinks=true,
    linkcolor=mydarkblue,
    citecolor=mydarkblue,
    filecolor=mydarkblue,
    urlcolor=mydarkblue,
    }

\newcommand{\rmB}{{\mathbf{B}}}

\newtheorem{definition}{Definition}
\newtheorem{remark}{Remark}

\usepackage[parfill]{parskip}

\title{Chain-of-Sketch: Enabling Global Visual Reasoning}

\date{}
\author{Aryo Lotfi\thanks{Equal contribution.}~~, Enrico Fini\textcolor{mydarkblue}{$^*$}, Samy Bengio, Moin Nabi, Emmanuel Abbe\\
\\[-3.5mm]
Apple\\
}

\begin{document}

\maketitle

\vspace{-0.3cm}
\begin{abstract}
Modern vision models have achieved remarkable success in benchmarks where local features provide critical information about the target. There is now a growing interest in tackling tasks requiring more global reasoning, where local features do not provide significant information. Minsky and Papert put forward such tasks in 1969 with their connectivity study, exposing the limitations of the perceptron model. In this paper, we introduce an expanded set of global visual datasets involving graphs, strings, mazes, and image grids. 
We show that large vision models still struggle to learn these tasks efficiently. Similarly, state-of-the-art multi-modal LLMs perform poorly on these datasets. We explain this learning inefficiency by means of the ‘globality degree’ measure. To mitigate this, we propose a method called chain-of-sketch (CoS).
Similar to the chain-of-thought and scratchpad techniques used in language models, CoS breaks the original task into intermediate visual steps to help learn a complex task. In addition, we show that not all CoS strategies perform equally well. Our key insight is to impose a Markovian structure on the CoS frames. This leads to the introduction of `inductive CoS' which achieves better out-of-distribution generalization and performs well even with smaller models compared to non-inductive variants.\footnote{An earlier version appeared on arXiv under the title “Visual Scratchpads: Enabling Global Reasoning in Vision”.}
\end{abstract}
\section{Introduction}
Modern computer vision models, as well as text models, are often pre-trained on vast datasets encompassing much of the knowledge available on the internet. While this has led to impressive capabilities, measuring the extent to which these models perform reasoning is still under investigation. Evidence suggests that many of these models, acting as blurry, compressed versions of the Internet, excel at smooth interpolation within their encoded knowledge, but often struggle to grasp underlying logic and extrapolate robustly. Unfortunately, classical visual benchmarks are limited to tasks that can often be tackled with superficial cues and local features. Despite progress in visual and multi-modal reasoning benchmarks \citep{yue2023mmmu, yue2024mmmu, hao2025can}, there is a need for datasets that rigorously test global reasoning and multi-step visual problem-solving.  In this work, we aim to bridge this gap by exploring when and how models are capable of learning tasks that require multi-step global processing of the input.

To that end, it is crucial to define the characteristics of global visual tasks. In contrast to local tasks, where a small subset of pixels—typically organized into patches—is sufficient to achieve better-than-random accuracy, global tasks require a more holistic understanding of the entire visual scene. For example, in ImageNet classification \citep{deng2009imagenet}, a single patch containing cat whiskers significantly increases the likelihood that the model will classify the image as a cat. This reliance on local features is further exemplified by the effectiveness of drastic image cropping in object-centric datasets, where self-supervised models such as DINO \citep{caron2021emerging-dino} employ aggressive multi-crop strategies, sometimes cropping as much as 90\% of the image, which empirically improves performance.
Humans, in contrast, do not rely solely on local information; for instance, when driving a car, it is insufficient to focus only on the view directly in front of the vehicle. A competent driver must recognize multiple visual objects in the environment and consider their complex behaviors before making decisions. Yet, using such complex real-world tasks, like autonomous driving, to study learning is impractical due to their complexity and unpredictability. Instead, we need interpretable and deterministic tasks with well-defined data generation processes to assess the reasoning ability of the models.

\begin{wrapfigure}{r}{0.3\textwidth}
    \centering
    \includegraphics[width=0.75\linewidth]{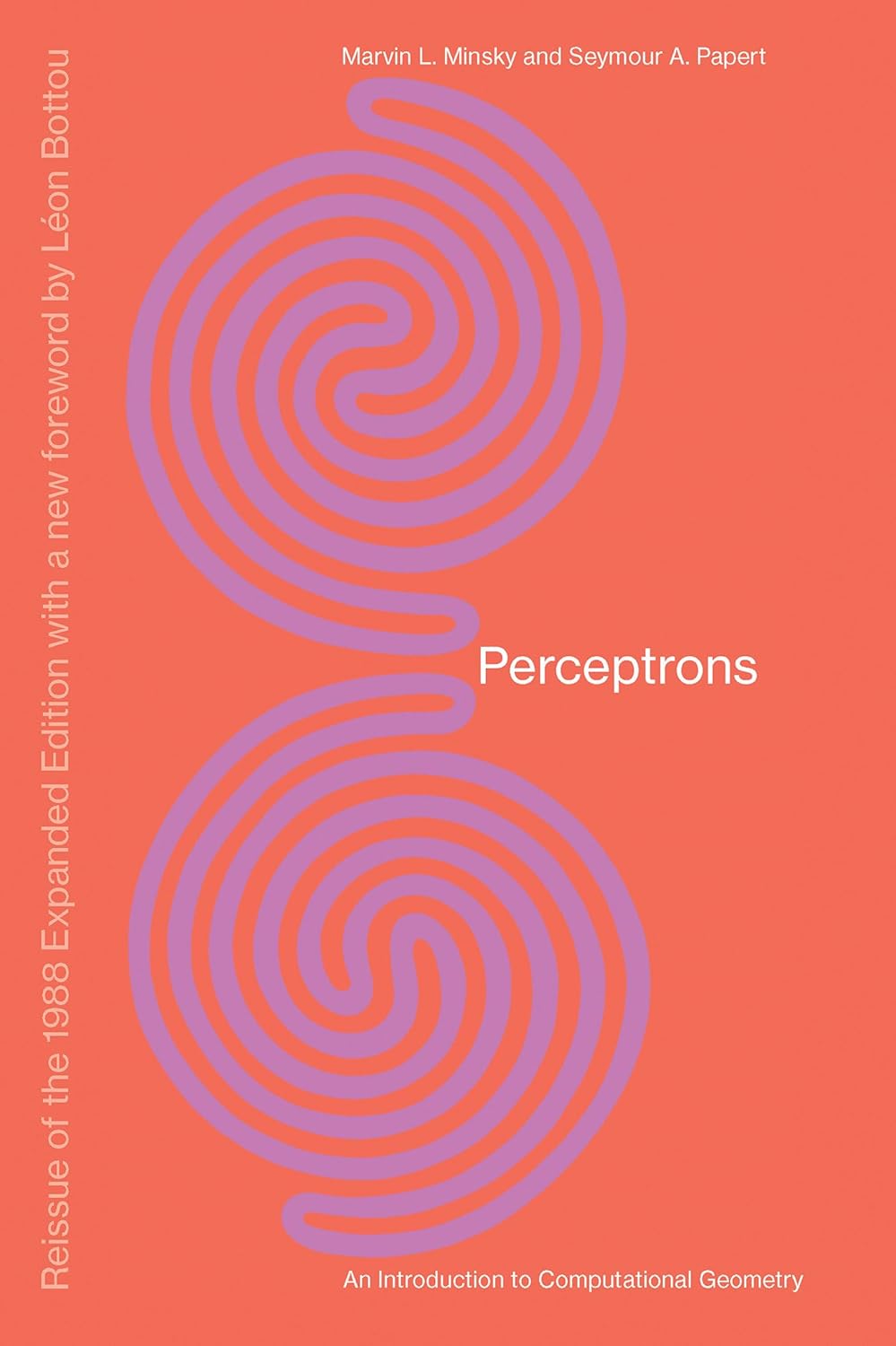}
    \caption{The cover of the 2017 edition of \textit{Perceptrons} by \citet{minsky1969perceptrons}.} 
    \label{fig:perceptrons-book-cover}
\end{wrapfigure}

To address this need for visual tasks with global multi-step reasoning, we propose five datasets, some reminiscent of the connectivity task of \citet{minsky1969perceptrons} that played a significant role in the AI winter (see Figure \ref{fig:perceptrons-book-cover}). In particular, we propose two connectivity-based tasks (cycles and strings) and two maze solvability tasks (rectangular and circular mazes). In addition, we re-purpose the pointer value retrieval (PVR) task \citep{Zhang2021PointerVR, abbe2022learning} in a visual setting. These tasks are inherently global because understanding a small portion of the input offers no meaningful information into the final label (e.g., whether the structure is connected or not). Further, their data generation processes are fully controllable and deterministic, allowing for straightforward manipulation of the task's complexity (e.g., by changing the size of the maze). These datasets enable us to simultaneously assess both reasoning and visual recognition abilities, which is the core objective of this paper. Interestingly, we find that even state-of-the-art multi-modal LLMs such as GPT-4o~\citep{hurst2024gpt}, o3, and o4-mini~\citep{openai2025o3o4mini} struggle on these datasets (see Section~\ref{sec:vllms}), barely performing better than random, which demonstrates the need for such global visual reasoning datasets.

\subsection{Contributions}
\begin{itemize}[leftmargin=*]
  \item \textbf{Exploration of locality and globality in the visual domain}: We analyze the concept of locality/globality in vision, providing a measure of the task's globality degree to better understand the learning complexity of models such as Vision Transformers (ViTs) \citep{dosovitskiy2020image}.

  \item \textbf{Datasets to investigate the limitations of global visual reasoning}: We propose a set of tasks involving graphs, strings, and mazes based on the connectivity idea and a pointer value retrieval task. These tasks are special as (1) no small subset of patches is sufficient to even weakly learn the target (and a more global multi-step reasoning is needed); (2) the complexity of the tasks can be scaled with task hyperparameters; (3)
   ViT models struggle to learn these tasks as the task complexity grows. We provide an explanation of these limitations using the globality degree measure \citep{locality}.
\item \textbf{Chain-of-sketch for global reasoning}: Similar to the chain-of-thought in text \citep{wei2023chainofthought}, we develop \textit{chain-of-sketch (CoS)} techniques to enable multi-step reasoning in vision models. CoS refers to the idea of learning intermediate visual targets that facilitate learning the original target, especially for global targets. Specifically,
    (1) we show that a single-frame CoS model can learn visual tasks that were not learnable without CoS;
    (2) we introduce a recurrent model for generating multi-frame CoS, the {\it inductive CoS}, that uses Markovian modeling of the intermediate steps and adaptive compute time at inference. We show that inductive CoS affords stronger reasoning capabilities on the above tasks by improving the OOD generalization (length generalization) and by enabling smaller models to learn despite their failure with the non-inductive CoS.  
\end{itemize}

\begin{figure*}[t]
    \centering
    \begin{subfigure}[t]{0.18\linewidth}
        \centering
        \includegraphics[width=\linewidth]{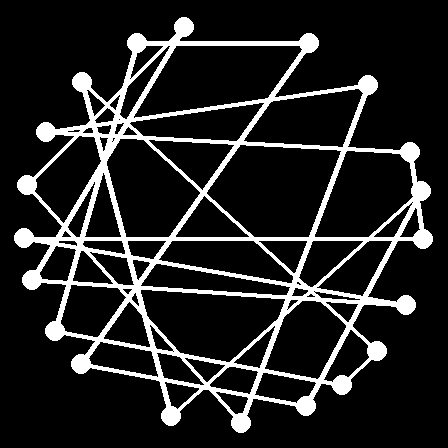}
    \end{subfigure}%
    \hspace{0.023\linewidth}%
    \begin{subfigure}[t]{0.18\linewidth}
        \centering
        \includegraphics[width=\linewidth]{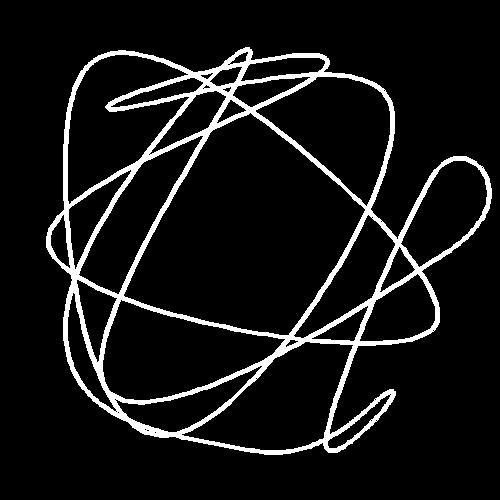}
    \end{subfigure}%
    \hspace{0.023\linewidth}%
    \begin{subfigure}[t]{0.18\linewidth}
        \centering
        \includegraphics[width=\linewidth]{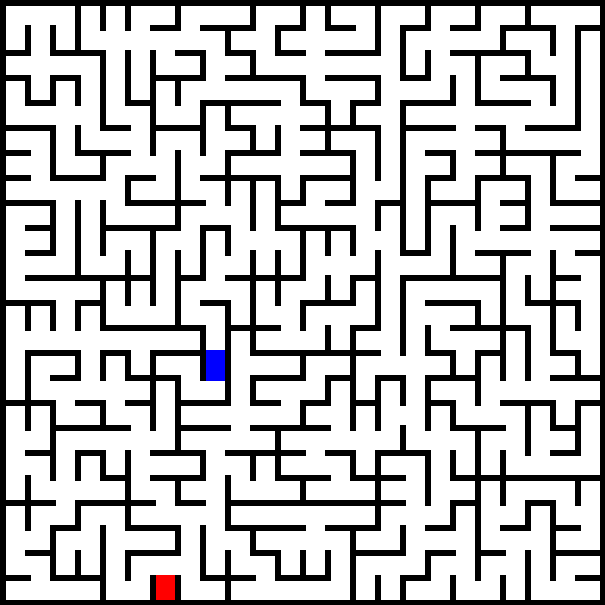}
    \end{subfigure}%
    \hspace{0.023\linewidth}%
    \begin{subfigure}[t]{0.18\linewidth}
        \centering
        \includegraphics[width=\linewidth]{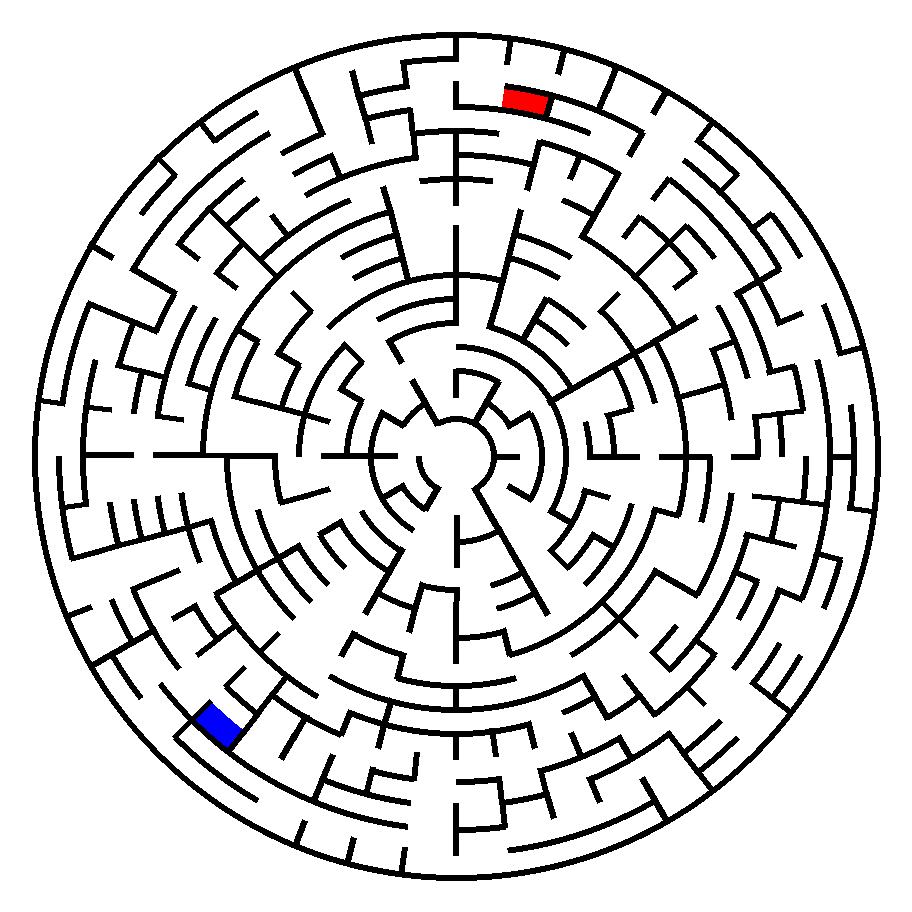}
    \end{subfigure}
    \hspace{0.023\linewidth}%
    \begin{subfigure}[t]{0.18\linewidth}
        \centering
        \includegraphics[width=\linewidth]{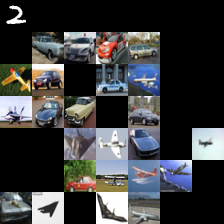}
    \end{subfigure}
    
    \begin{subfigure}[t]{0.18\linewidth}
        \centering
        \includegraphics[width=\linewidth]{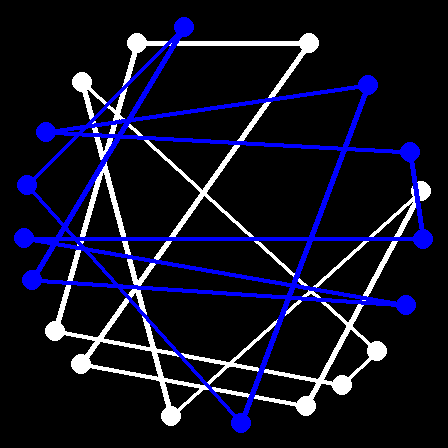}
        \caption{Cycles}
    \end{subfigure}%
    \hspace{0.023\linewidth}%
    \begin{subfigure}[t]{0.18\linewidth}
        \centering
        \includegraphics[width=\linewidth]{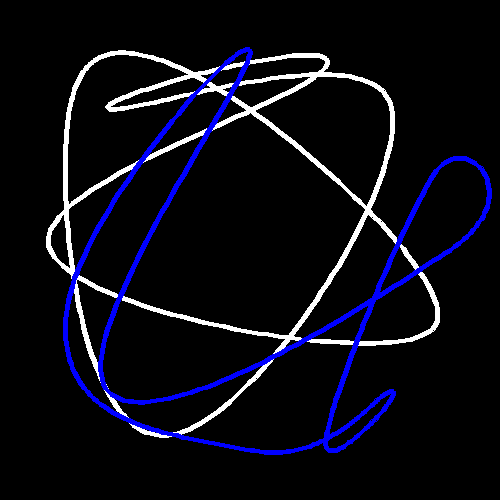}
        \caption{Strings}
    \end{subfigure}%
    \hspace{0.023\linewidth}%
    \begin{subfigure}[t]{0.18\linewidth}
        \centering
        \includegraphics[width=\linewidth]{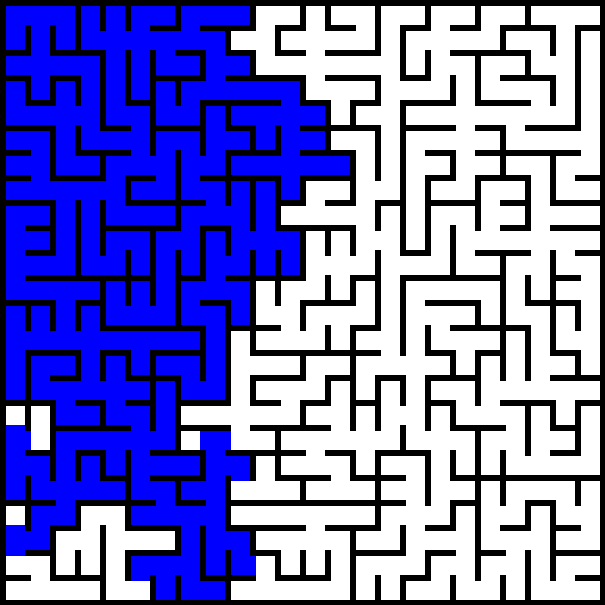}
        \caption{Rect. maze}
    \end{subfigure}%
    \hspace{0.023\linewidth}%
    \begin{subfigure}[t]{0.18\linewidth}
        \centering
        \includegraphics[width=\linewidth]{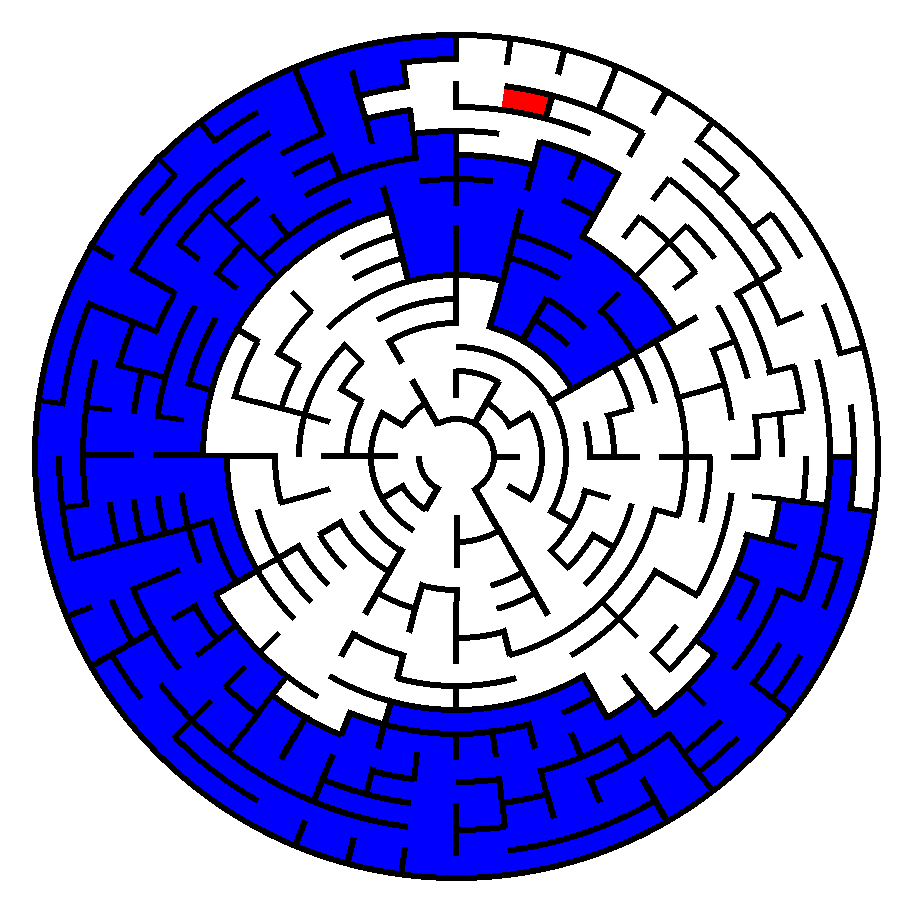}
        \caption{Circular maze}
    \end{subfigure}
    \hspace{0.023\linewidth}%
    \begin{subfigure}[t]{0.18\linewidth}
        \centering
        \includegraphics[width=\linewidth]{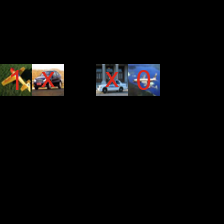}
        \caption{Image-PVR}
    \end{subfigure}
    \vspace{-1mm}
    \caption{Examples of different tasks. The first row shows the inputs; the second row shows the complete sketch (i.e., the target frame in single-frame CoS or the final frame in multi-frame CoS). In the graph and string tasks, there are one or two connected components (two shown). In the maze tasks, there are always two connected components, but the start and end nodes may or may not be connected (both cases shown). In the PVR task, the label is the parity of the airplane class in the indicated row ($0$ in this example).
    }
    \label{fig:task-examples}
\end{figure*}

\section{Global visual reasoning datasets}\label{sec:global-datasets}
Vision models have shown remarkable performance on a range of tasks, including image classification, image segmentation, and object detection.
However, mainstream vision datasets, including reasoning benchmarks described in Section \ref{related:visual-reasoning}, have two characteristics in common:
\begin{enumerate}[leftmargin=*]
    \item Local features in the image are informative. For example, if we consider an image partitioned into a set of patches, there is usually a small subset of patches that provides significant information about the target (e.g., the label).
    \item These tasks can be instinctively and instantaneously solved by humans. That is, humans do not need to ponder for long periods of time to solve these tasks. Considering the System 1 / System 2 terminology \citep{kahneman2011thinking}, these visual tasks are handled by our System 1. In general, little or no multi-step chain of entailments is necessary to solve these tasks.
\end{enumerate}

However, not all visual tasks share these characteristics. For instance, consider solving a maze, i.e., answering whether two points in a maze are connected or not. Assuming the size of the maze is large enough, humans require some reflection before solving the maze. Normally, humans would trace the paths on the maze with a pen to see where the starting point leads. Importantly, apart from trivial edge cases where the start and end locations are close, local features are not informative for the maze task. For example, if only a few patches of a maze are given, one cannot solve it with high probability. Motivated by the latter, we propose the following visual datasets in this paper:
\begin{itemize}[leftmargin=*]
    \item \textbf{Connectivity datasets.} Inspired by \citet{minsky1969perceptrons}, we consider two datasets based on the notion of connectivity. 
    \begin{itemize}[leftmargin=*]
    \item \textbf{Cycles task.} In this task, $2n$ nodes are drawn randomly (on an invisible circle) in the image. There are also $2n$ edges between these nodes that form either one cycle of size $2n$ or two cycles of size $n$. The task is to determine whether the graph is connected (one cycle, label 1) or not (two cycles, label 0). See Figure \ref{fig:task-examples} for an example. In this task, one has to reason over at least $n$ nodes and the connections between them to determine the label correctly, as any $n-1$ nodes provide no information on whether there are two cycles or one. Thus, one can simply increase the complexity of this task by increasing $n$. 
    \item \textbf{Strings task.} In order to further increase the visual complexity, we consider a dataset consisting of random strings. In each sample, there are either two closed strings or one longer closed string. The dataset generation process for these curves is similar to the cycles task above, except that in the strings we do not make the (anchor) nodes visible and also connect them using third-degree B\'ezier curves which produces continuous strings (see Figure \ref{fig:task-examples}). Similar to the cycles task, one can increase the complexity of this task by increasing the number of invisible anchor points $2n$, which leads to longer, more entangled strings. 
    \end{itemize}
    \item \textbf{Maze solvability.} We also consider a maze task in which there are always two connected components, and we have a start/source point (shown in blue) and an end/sink point (shown in red). The source and sink are in the same connected component or not equiprobably. The task is to determine whether they are connected (label 1) or not (label 0). We provide this dataset in a rectangular and a circular version to increase the visual complexity. Examples can be seen in Figure~\ref{fig:task-examples}. To adjust the complexity of maze datasets, one can modify the size of the maze and hence the number of cells, the size of the components, and the distance between the source and sink (if connected).
    \item \textbf{Image Pointer Value Retrieval (PVR).} In our image pointer value retrieval (PVR) task, similar to \citet{Zhang2021PointerVR, abbe2022learning}, each image is a grid of fixed size $n\times n$ (for $n \leq 10$) where there is an MNIST \citep{lecun2010mnist} digit acting as a pointer in the top left cell of the grid pointing to one of the next $n-1$ rows. For a fixed $k$, in each of the next $n-1$ rows of the grid, we have $k$ images where each image is sampled uniformly from the plane and car categories of the CIFAR-10 \citep{cifar} dataset. The label of an image is given by the parity of the number of occurrences of the leftmost object in the indicated row.\footnote{For even $k$, the label is equal to the parity of the number of cars, which is the same as the parity of the number of planes.} Note that the globality of this task is at least $k+1$, as one has to use the pointer and also all the $k$ images of the corresponding row to have non-zero information about the label (as any $k-1$ images of a row have no mutual information with the target since the parity function can flip depending on the single unseen image). Thus, one can also easily adjust the globality of the task by varying $k$. Figure \ref{fig:task-examples} shows a PVR task with $n=7$ and $k=4$.
\end{itemize}

For each task, there exists a natural chain-of-sketch (a single frame or a sequence of frames) that uncovers the underlying reasoning behind the label. For the maze, similar to what humans do, we can start coloring from the source cell (i.e., the cell in blue) to see which areas are reachable until reaching the sink cell or the end of the maze region, similar to performing a breadth-first search (BFS). This coloring is similar to what humans would naturally do by following the paths from the starting point to see which one (if any) leads to the sink cell. For the cycles task, we can use a similar idea: we can start by coloring one node, and then color all of the nodes that are connected to this node (i.e., either half or all of the graph). Analogously, for the strings task, the CoS would be coloring one of the strings if there are two strings or coloring the whole string if there is only one. To disambiguate which cycle/string to color, we always color the cycle/string that passes through the rightmost (anchor) node. For the PVR task, we can keep only the row indicated by the pointer and then compute the parity from left to right by checking whether each object belongs to the same class as the leftmost object, and updating the parity accordingly. See Figure \ref{fig:task-examples} for single-frame CoS examples for different tasks. 

Note that the CoS can have a single frame format where the full sketch (e.g., with all the coloring done) is shown. The CoS can also be generated in multiple frames, i.e., consecutive frames that lead to the final sketch. This CoS is again analogous to what humans do as they progressively create the full sketch. For example, this could be coloring a distance of $10$ when doing the search for the maze problems; coloring (up to) two anchor nodes in the cycles and strings tasks; and checking one additional object in the PVR task. An example of doing so for the cycles task is depicted in Figure \ref{fig:multi-frame-scratchpad}. See Appendix \ref{app:additional-figs} for CoS frames for other datasets. 

\begin{figure*}[t]
     \centering
     \begin{subfigure}[b]{0.16\textwidth}
         \centering
         \includegraphics[width=\textwidth]{figures/examples/cy20/0/0.png}
     \end{subfigure}
     \hfill
     \begin{subfigure}[b]{0.16\textwidth}
         \centering
         \includegraphics[width=\textwidth]{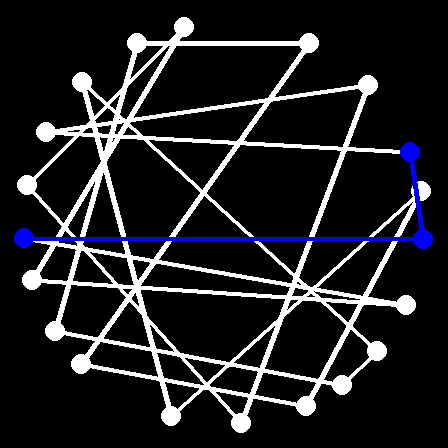}
     \end{subfigure}
     \hfill
     \begin{subfigure}[b]{0.16\textwidth}
         \centering
         \includegraphics[width=\textwidth]{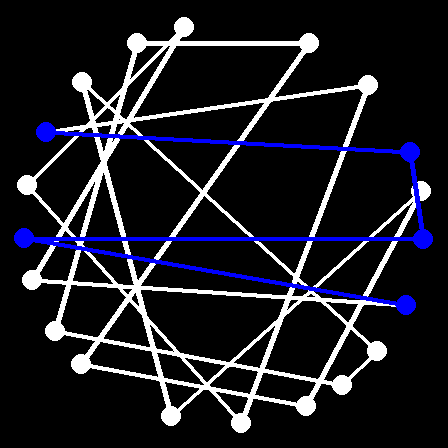}
     \end{subfigure}
     \hfill
     \begin{subfigure}[b]{0.16\textwidth}
         \centering
         \includegraphics[width=\textwidth]{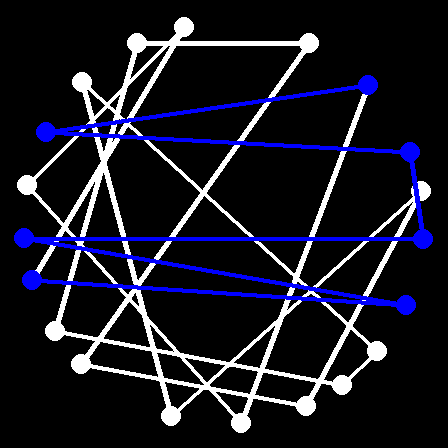}
     \end{subfigure}
     \hfill
     \begin{subfigure}[b]{0.16\textwidth}
         \centering
         \includegraphics[width=\textwidth]{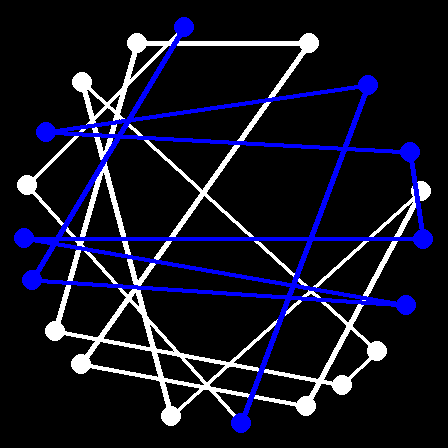}
     \end{subfigure}
     \hfill
     \begin{subfigure}[b]{0.16\textwidth}
         \centering
         \includegraphics[width=\textwidth]{figures/examples/cy20/0/5.png}
     \end{subfigure}
    \vspace{-1mm}
    \caption{An example of the cycles task with a CoS containing several frames. The input image is presented on the left, followed by different frames of the CoS from left to right, ending with the complete sketch.}
    \label{fig:multi-frame-scratchpad}
    \vspace{-2mm}
\end{figure*}

\subsection{Globality in the visual domain} \label{sec:connection-locality}
Recently, \citet{locality} proposed the notion of globality degree to explain why tasks requiring global reasoning are hard for Transformers to learn, and to understand the effectiveness of scratchpads \citep{nye2021work} and chain-of-thought \citep{wei2023chainofthought} techniques in the text domain. For input tokens $X_1, \ldots, X_n$ and output $Y$, the globality degree of a task is defined as the minimum number of tokens $k$ such that there exist $k$ tokens $X_{i_1}, \ldots, X_{i_k}$ that along with the histogram of tokens $\hat P_X$\footnote{In text, histogram refers to reporting how many times each token is appearing regardless of its position (similar to the bag of words).} provide significant information on the target $Y$, i.e., $I(X_{i_1}, \ldots, X_{i_k}, \hat P_X; Y) = n^{-O_n(1)}$ where $I$ is the mutual information. It is further conjectured, with empirical support, that the learning complexity of tasks increases with their globality degree, and Transformers can only learn tasks with a constant globality degree efficiently (using a model of polynomial size and a polynomial number of iterations). We extend this definition to vision tasks learned with models like vision Transformers (ViTs) \citep{dosovitskiy2020image}, using a finer-grained formulation that integrates the amount of information.

\begin{definition}{\bf Globality degree} (in the visual domain; see Appendix \ref{app:globality-details}). 
    Assume images are partitioned into patches $X_1, \ldots, X_n$. We define the globality degree with threshold $\alpha$ of a task as the minimum number $k$ such that there exist patches $X_{i_1}, \ldots, X_{i_k}$ that satisfy $I(X_{i_1}, \ldots, X_{i_k};Y) \geq \alpha$ where $I$ is the mutual information. 
\end{definition}
In words, this is the least number of patches $k^*$ required to obtain an $\alpha$-mutual information with the target. The higher $\alpha$, the more informative these patches are about the target, and the lower $k^*$ (for the same $\alpha$), the less global the task is. Given the arguments given by \citet{locality}, we expect the learning complexity of the task to scale with $\frac{n^{k^*}}{\alpha}$ where $k^*$ is the globality degree. Therefore, the tasks become easier to learn as they have smaller globality degrees (e.g., $k^*=O_n(1)$) and harder to learn as their globality degrees increases. We further discuss the theoretical aspects in Appendix \ref{app:globality-details}.

\begin{remark}
Note that a target being ``local'' by having a low globality degree $k^*$ for a significant $\alpha$ {\it does not mean that the target depends only on these few $k^*$ patches}. The target can still depend on all the patches. What it means is that these $k^*$ patches are sufficient in order to obtain {\it non-trivial} information about the target, i.e., the model can achieve a non-trivial accuracy using these patches and thus weakly learn the task (this weak learning may as well provide a starting point for strong learning). We further clarify this using an experiment in Appendix \ref{app:pvr_majority}.
\end{remark}

According to the definition above, vision tasks such as classical image classification are ``local'' as a few patches often provide significant information on the class (e.g., having a patch containing a dog's ear significantly increases the likelihood of predicting the dog class). Whereas in our proposed datasets, seeing a few patches often provides no information about the label. Hence, our proposed datasets have a high globality degree, or in short, are ``global''.  For example, in the maze examples, seeing just a few patches from the maze does not help determine the label. Similarly, for the cycles task of size $2n$, if one only sees the connections between $n-1$ nodes, one has no information about the label.

\begin{figure*}[htb]
    \centering
    \begin{minipage}[t]{0.48\textwidth}
        \centering
        \includegraphics[width=0.95\textwidth]{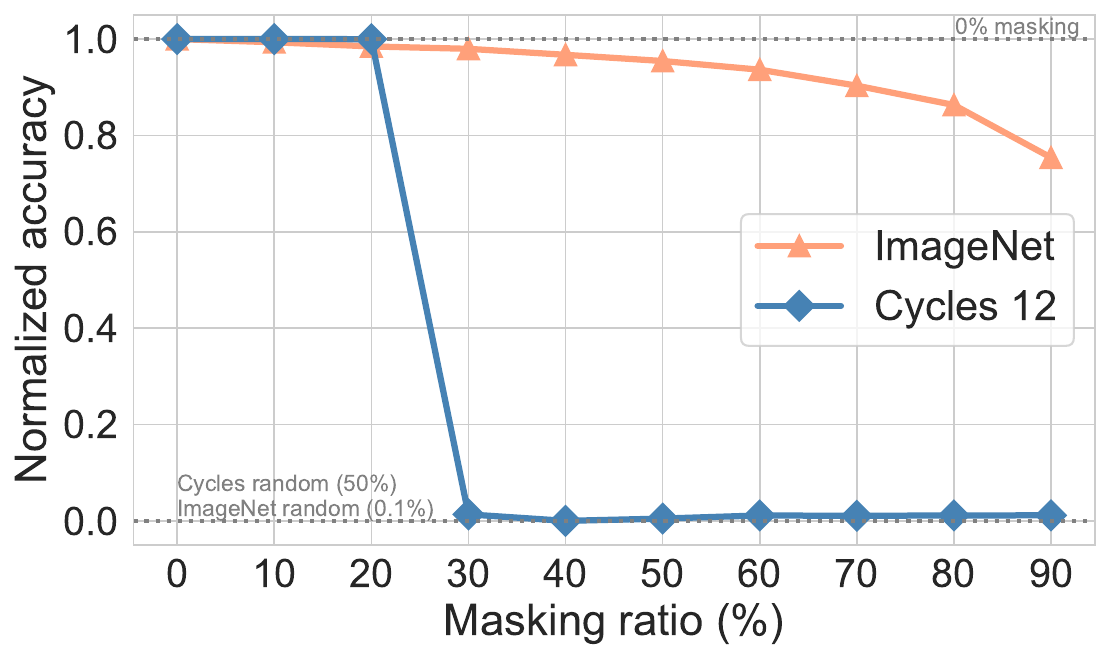}
        \caption{Experimental evidence that cycles 12 is more global compared to common computer vision benchmarks (e.g. ImageNet). Cycles 12 quickly becomes hard to learn when more patches are masked, while ImageNet remains well above random accuracy.}
        \label{fig:masking_ratio}
    \end{minipage}
    \hfill
    \begin{minipage}[t]{0.48\textwidth}
        \centering
        \includegraphics[width=0.95\textwidth]{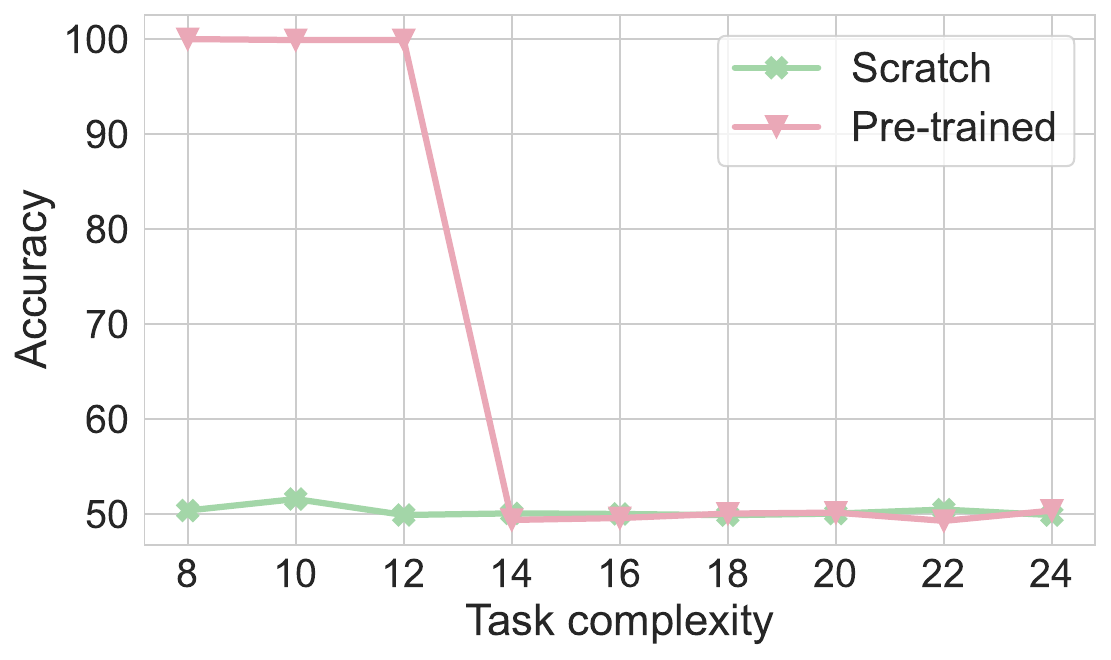}
        \caption{Comparison between training from scratch and initializing with a pre-trained model, as the task complexity varies. Pre-training is not sufficient to guarantee convergence as the complexity increases. Learning without pre-training is not possible even for simple instances.}
        \label{fig:complexity}
    \end{minipage}
\end{figure*}

\textbf{Empirical validation.} To further support our claim that mainstream vision tasks are local while our proposed tasks are not, we conduct the following experiment. For each sample in a given dataset, we mask the patches with probability $p$ at both training and inference and measure the performance of the model for different values of $p$. We perform this experiment on the cycles 12 task and ImageNet. Since the cycles 12 task is not learnable from scratch (see Figure \ref{fig:complexity}), we start from a CLIP~\citep{radford2021learning} pre-trained ViT-L/14 checkpoint~\citep{fang2023data} for both. The results are shown in Figure \ref{fig:masking_ratio}. We observe that the model demonstrates good performance on the ImageNet dataset even when 90\% of the image is masked while it cannot learn the cycles 12 task once 30\% (or more) of the image is masked.\footnote{We use min-max normalized accuracy in the plot, including the random baseline for normalization.} The latter shows that the cycles 12 task is a high globality dataset where one needs on average at least 70\% of the patches to gain minimal information on the label while ImageNet is a local task where weak learning is possible with only 10\% of the patches.

We stress that a model trained from scratch was not able to learn the cycles task even when no patch was masked. To better explain this phenomenon, in Figure \ref{fig:complexity}, we compare the performance when initializing with a pre-trained model~\citep{fang2023data} versus a model trained from scratch on the cycles task of varying size. This shows that as the task complexity increases with the number of nodes, none of the models is able to learn the cycles task, meaning that even strong internet-sourced priors in the pre-trained model are not helpful.

\subsection{Global visual tasks are hard for multi-modal LLMs} \label{sec:vllms}
In this section, we assess the difficulty of our proposed tasks for state-of-the-art multi-modal large language models (LLMs). Specifically, we selected two of our tasks for this evaluation: the cycles task and the rectangular maze\footnote{For the rectangular maze task, we slightly reduced the size of the blue and red cells in our images so that they do not touch the walls of the maze, eliminating any potential ambiguities for the model.}. These tasks are relatively natural and likely resemble examples encountered during pretraining. Moreover, they are easier than their respective counterparts, i.e., the strings task and the circular maze.\footnote{The models can potentially come up with symbolic representations for the cycles and rectangular maze tasks, whereas doing so for the strings and circular maze tasks is more challenging.}

For evaluation, we used the GPT-4o \citep{hurst2024gpt}, o3, and o4-mini \citep{openai2025o3o4mini} models and 1000 test samples per task (500 from each class). We considered two settings: (1) providing the model with a detailed prompt and asking it to solve the task, and (2) using a similar prompt along with four in-context examples (two from each class) presented as images. We disabled tool calling in both settings as our main interest is to measure the model's visual reasoning capabilities without allowing it to execute any code. The results are shown in Table~\ref{tab:llm-results}. We observe that models perform barely better than random as the tasks get harder. Interestingly, in-context learning (ICL) does not appear to be particularly beneficial in our setting. See Appendix~\ref{app:prompts} for more details and the prompts used. 

\begin{table}[h]
\centering
\begin{tabular}{l l c c c c c c}
\toprule
\multirow{2}{*}{\textbf{Task}} &\multirow{2}{*}{\textbf{Size}} &  \multicolumn{2}{c}{\textbf{GPT-4o}} & \multicolumn{2}{c}{\textbf{o4-mini}} & \multicolumn{2}{c}{\textbf{o3}} \\
\cmidrule(lr){3-4} \cmidrule(lr){5-6} \cmidrule(lr){7-8}
& & \textbf{Prompt} & \textbf{ICL} & \textbf{Prompt} & \textbf{ICL} & \textbf{Prompt} & \textbf{ICL} \\
\midrule
\multirow{3}{*}{Cycles} & 6 nodes & $61.9\%$ & $58.9\%$ & $68.9\%$ & $62.2\%$ & $63.4\%$ & $62.1\%$ \\
                            & 8 nodes & $51.6\%$ & $50.4\%$ & $55.4\%$ & $52.9\%$ &$54.7\%$ & $51.4\%$ \\
                            & 10 nodes & $49.9\%$ & $48.8\%$ & $50.5\%$ & $48.4\%$ &$49.8\%$ & $51.3\%$ \\
\midrule
\multirow{2}{*}{Maze (rect.)} & $4\times4$ & $50.6\%$ & $55.4\%$ & $55.3\%$ & $52.4\%$ & $52.4\%$ & $51.9\%$ \\
                            & $6\times6$ & $48.4\%$ & $51.8\%$ & $49.4\%$  & $51.3\%$ & $51.2\%$ & $48.6\%$ \\
\bottomrule
\end{tabular}
\vspace{0.5cm}
\caption{The accuracy of different models on our proposed tasks. Prompt columns represent the experiments where no in-context example was included while ICL is used for experiments with 4 in-context examples. Note that $50\%$ would correspond to the accuracy of a random model. Also, we have used 1000 samples for each task, giving us a 95\% confidence interval of approximately $\pm 3\%$.}
\label{tab:llm-results}
\end{table}

\subsection{Breaking globality with chain-of-sketch} \label{sec:globality-cos}
We next discuss the connection between our chain-of-sketch and the scratchpad and chain-of-thought (CoT) used in text and why it helps. The idea of CoT generally refers to training the models with intermediate steps, so that they generate both the reasoning steps and the final answer at inference. For instance, \citet{nye2021work} showed that for simple math questions training language models to output intermediate steps before the final answer leads to higher accuracy than training the model to directly output the answer. \citet{wei2023chainofthought} further developed this methodology and showed that pre-trained models can generate these intermediate steps with in-context examples and prompting as well. See related work in Section \ref{sec:related} for further discussion.

In \citep{locality}, it is shown that chain-of-thought can reduce the globality degree and, in doing so, the learning complexity as well.
More specifically, chain-of-thought can provide intermediate targets $Y_1, \ldots, Y_m$ such that $Y_m = Y$ is the final target and each $Y_i$  is of low globality given the previous intermediate targets and the input, which makes predicting them in a sequential manner easily learnable.  The same is true for our proposed datasets and CoS as each sketch frame is a low globality function of the previous frame and the label is also given by a low globality function of the final frame. For instance, each frame of our CoS in the cycles task reduces the globality degree from a quantity growing with the number of nodes to a constant, as in each step only the neighboring nodes are considered.

{\bf Single-frame CoS and hierarchical learning.} The single-frame CoS collapses all the steps of the multi-frame CoS (into the final frame). Both CoS, however, break the globality of the original target. This may not be obvious for the single-frame CoS, but it results from the fact that there are parts of the single-frame CoS target that can be predicted with low globality from the input image, i.e., weak learning of this single sketch is possible. For instance, for the cycles task, one always starts by coloring the edges adjacent to the preset vertex. The model can then learn how to progress the coloring `on its own', rather than relying on the guidance of a multi-frame CoS, and thereby turns its weak learning into strong learning. This hierarchical learning mechanism is also known as the staircase phenomenon \citep{mergedstaircase, abbe2023leap}, and is further discussed in Section~\ref{sec:staircase}. Note that the label is also a low globality function of the full sketch in the single-frame CoS.


\section{The CoS methodologies}\label{sec:methodology}
A CoS consists of adding intermediate visual steps to learn a given task. The goal is generally to have a sequence having a reduced globality compared to the original target's globality. In this paper, we consider three variants of CoS to evolve from the no-CoS model to break the globality of the target: single-frame CoS, multi-frame CoS, and inductive CoS. 
We describe these below and discuss similarities with previous work in Sections~\ref{related:reasoning_with_transformers}, \ref{related:cot-scratchpad}, and \ref{related:recurrent}.

\textbf{No-CoS baseline.} This baseline corresponds to predicting the target label directly from the input image without any intermediate step. 

We use a ViT \citep{dosovitskiy2020image} architecture with a classification token, \texttt{CLS}, for this setting. We use a linear layer on the \texttt{CLS} token features to compute the label logits and we use the cross-entropy loss function for training. As shown in Section \ref{sec:model-size-exps}, this model is not capable of learning the proposed datasets.

\subsection{Single-frame CoS}
In this case, the CoS introduces a single frame to be predicted as the intermediate target. For instance, for the cycles task it would be coloring the cycle that passes through the rightmost node, while for the maze task, it would be coloring the region reachable from the start cell of the maze. See Figure \ref{fig:task-examples} for examples of the single-frame CoS method for different tasks. 

For implementation, we keep the ViT encoder with a \texttt{CLS} as the backbone and add a linear layer to the hidden representation of the last Transformer layer to predict the sketch image. During training, we use cross-entropy loss to supervise the label and a pixel-wise mean-squared loss similar to~\citet{he2022masked, el2024scalable, fini2024multimodal} to supervise with the sketch image. In Section \ref{sec:model-size-exps}, we show that the single-frame CoS improves accuracy over the no-CoS model, and for large enough models, it may be able to learn the proposed tasks. 

\subsection{Inductive CoS}
In CoS with multiple frames, the intermediate target of the single-frame CoS is decomposed into multiple steps. Namely, rather than providing the full CoS image in one single frame, the model is provided with a sequence of frames that gradually lead to the full CoS image. For example, for the cycles task each CoS frame corresponds to coloring two edges at a time (incrementing the distance to the preset vertex). See Appendix \ref{app:dataset-generation} for the details of CoS frames and Appendix \ref{app:CoS-examples} for examples of CoS frames for different tasks.

One way to implement CoS with multiple frames is to use a similar architecture to the single-frame CoS and predict all the intermediate images in parallel. We use ``multi-frame CoS'' to refer to this preliminary model. This model serves as a useful baseline for ablation studies in Section \ref{sec:ablations}. Next, we introduce the inductive CoS method.


Inductive CoS has also multiple steps/frames, but these are learned sequentially with an inductive model. Namely, the sequence is Markovian, i.e., each new frame depends only on the previous one.
More precisely, the model has a recurrent component $\mathcal{M}$ that takes an input image (either the input image or a CoS frame)  and predicts three outputs: the next CoS frame ($\hat f$), the label ($\hat y$), and a binary halting variable ($\hat h$). This recurrent module is applied to the input image and the subsequent intermediate frames until the halting signal is triggered (or an upper limit of recurrences is reached). The predicted label at the last recurrence is the predicted label of the model. Note that generating each CoS frame depends only on the last generated frame (or the input image) and the recurrent module does not retain a history of prior frames. As a result, the model is independent of the number of sketches (frames) used in each sample.

We will see that the inductive CoS affords better OOD generalization than the single-frame and multi-frame CoS, due to its compositional structure that is independent of the number of frames a CoS has. The inductive CoS will also enable smaller models to learn when the other CoS variants do not on most considered tasks.  
    
\textbf{Training procedure.} For training, we initially used teacher forcing, which consists of providing the model with perfect frames from the training set. However, this approach creates a discrepancy during inference, where the model sees its own generated frames as input. Generated frames may exhibit a slightly different distribution as the reconstructions are not perfectly accurate. While this issue is well studied in text generation, where discrete tokens are used, it becomes more pronounced in vision tasks with continuous outputs. To mitigate this discrepancy, we use an alternated training procedure where the model sees perfect frames 50\% of the time, and generated frames the other 50\%. This ensures that the model learns to handle imperfect inputs, leading to improved performance during inference. 


\section{Experiments on OOD generalization and  model size}\label{sec:exps}
In this section, we show the performance of different methods on our proposed datasets, focusing on required model size and OOD generalization. Each of our datasets contains $10^6$ (1M) training samples. See Appendix \ref{app:training-details} for more details on the experiments. Further see Table \ref{tab:llm-results} in Section~\ref{sec:vllms} for results with multi-modal LLMs such as GPT-4o~\citep{hurst2024gpt}, o3, and o4-mini~\citep{openai2025o3o4mini} where we show these models perform only slightly better than random on our proposed tasks.

\subsection{Model size experiments}\label{sec:model-size-exps}
\begin{figure*}[t]
     \centering
     \begin{subfigure}[b]{0.32\textwidth}
         \centering
         \includegraphics[width=\textwidth]{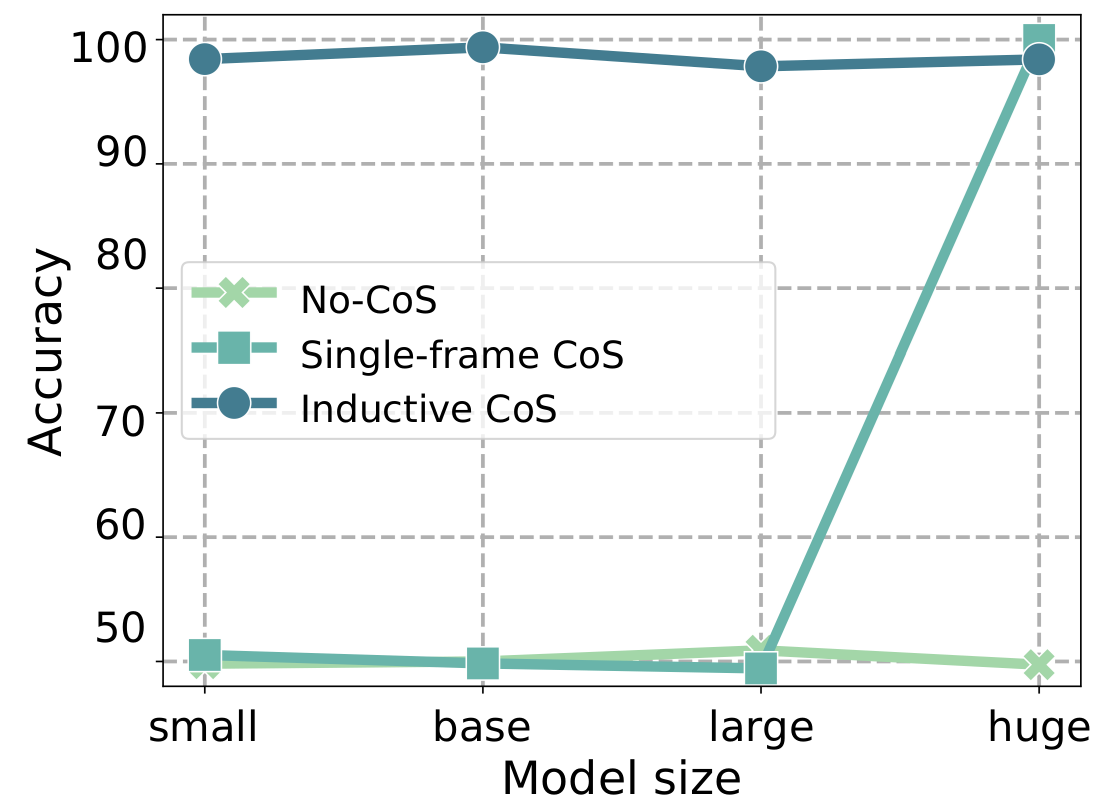}
         \caption{Cycles 24 dataset}
     \end{subfigure}
     \hfill
     \begin{subfigure}[b]{0.32\textwidth}
         \centering
         \includegraphics[width=\textwidth]{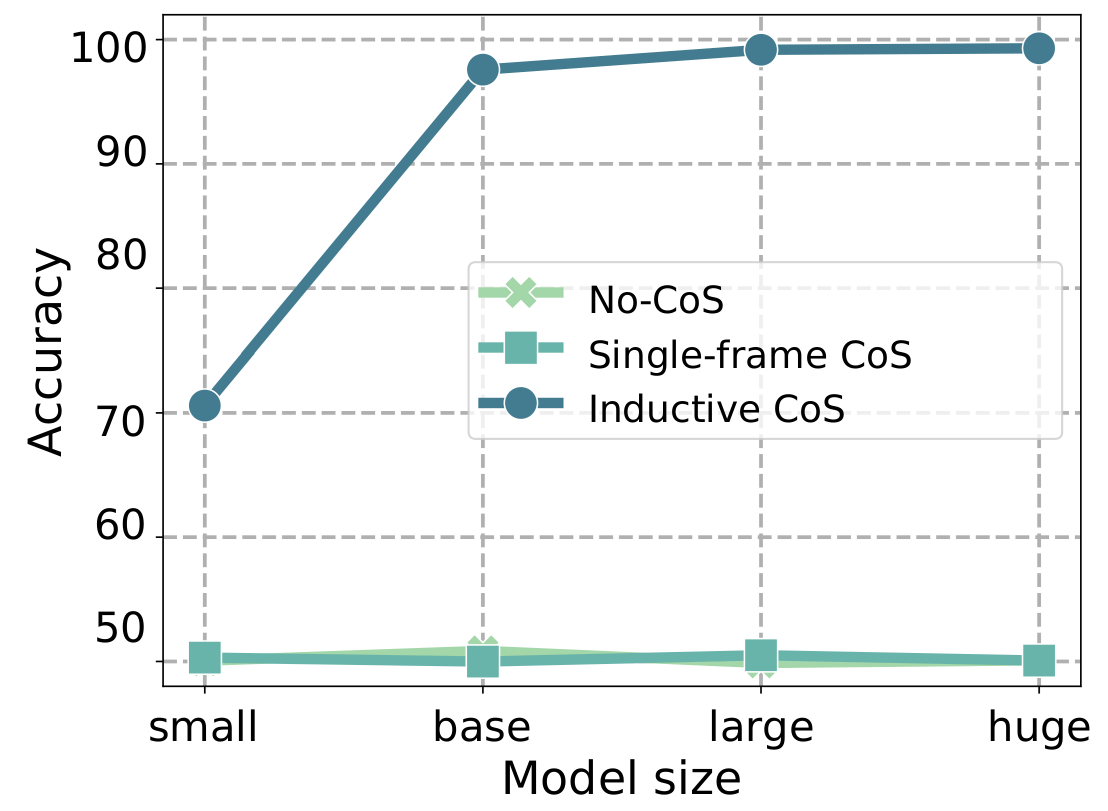}
         \caption{Strings 20 dataset}
     \end{subfigure}
     \hfill
     \begin{subfigure}[b]{0.32\textwidth}
         \centering
         \includegraphics[width=\textwidth]{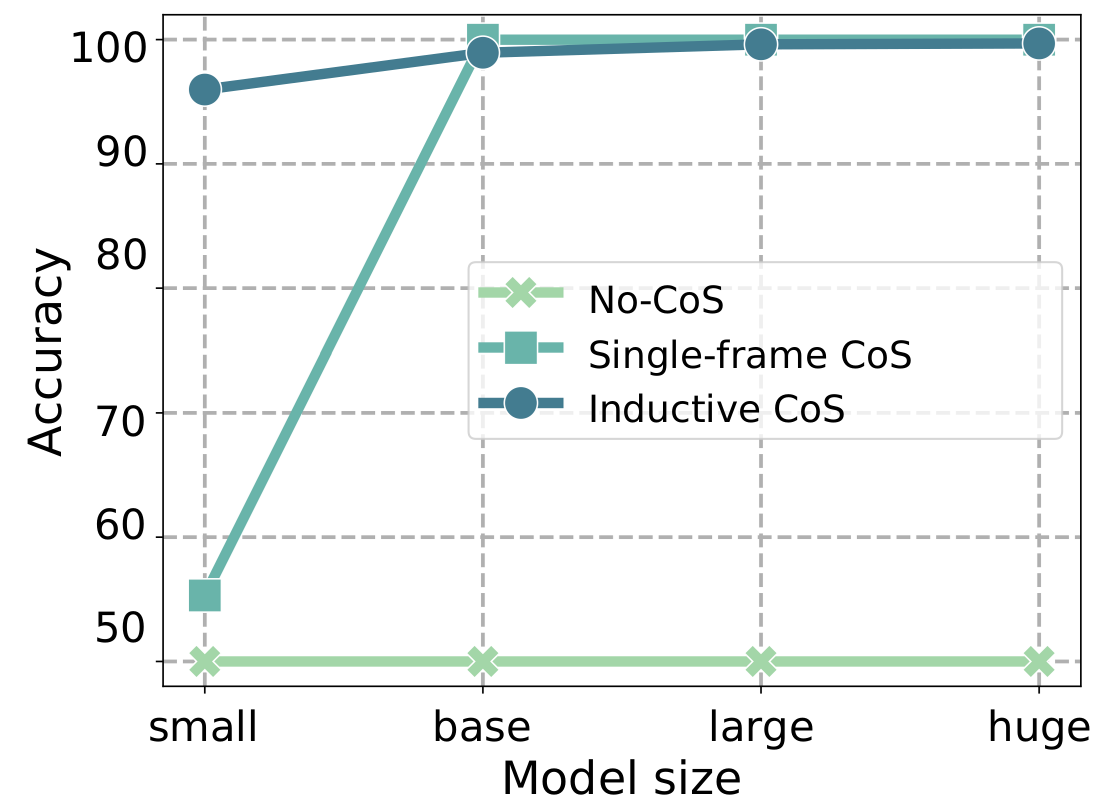}
         \caption{Maze (rect.) 32 dataset}
     \end{subfigure}
    \vspace{-2mm}
    \caption{Validation accuracy for different datasets learned by various methods and model sizes. We can see that the model without a CoS is not capable of learning any of these tasks, while for large enough models, the single-frame CoS model may be able to learn. Further, the inductive CoS model can learn all the tasks with smaller models than the single-frame CoS model.}
    \label{fig:model-size-main}
\end{figure*}

First, we compare the performance of different methods with varying model sizes on our proposed datasets. In particular, we compare the no-CoS baseline, the single-frame CoS, and the inductive CoS model used for multi-frame sketch prediction. Moreover, we use four different sizes for the ViT~\citep{dosovitskiy2020image} encoder of our models: small, base, large, and huge, which have respectively around $22M$, $86M$, $307M$, and $632M$ parameters (see Appendix \ref{app:model-implementation} for detailed specifications). The accuracy of different methods with different model sizes is shown in Figure \ref{fig:model-size-main}. It can be seen that the no-CoS baseline is not able to go beyond random accuracy for any of these tasks. On the other hand, the single-frame CoS model can learn the PVR($7\times 7, k=4$), cycles 24, maze (circ.) 16, and maze (rect.) 32 tasks for appropriately large models while it still cannot learn the strings 20 task. 
The inductive CoS model used for the multi-frame prediction, however, learns the proposed tasks even with smaller models. 
We report the results for the circular maze and PVR dataset in Appendix \ref{app:additional-exps}.

\textbf{Compute overhead.} We note that the number of parameters for the three methods is very similar. The single-frame model only adds a linear layer for predicting the CoS to the no-CoS baseline. Likewise, the inductive CoS model only adds a linear layer for predicting the halt signal to the single-frame model. However, during inference, the inductive CoS model is applied a varying number of times and uses compute adaptively depending on the complexity of the sample, and therefore usually requires a higher compute during inference compared to the single-frame CoS for a model of the same size. Nevertheless, as shown in Figure \ref{fig:model-size-main}, we can often use much smaller models with the inductive CoS model. We further discuss compute trade-offs in Appendix \ref{app:compute-overhead}.

\subsection{OOD generalization} \label{sec:ood-exps}

Next, we consider the out-of-distribution (OOD) generalization performance of different methods and show that the inductive CoS model exhibits superior OOD generalization. 
This observation is due to the fact that the inductive CoS model only learns the steps of the reasoning process, and as a result, is independent of the number of reasoning steps required---allowing it to generalize to harder problems with its adaptive compute time.

\begin{figure}[b]
    \centering
    \begin{minipage}[b]{0.37\textwidth}
        \centering
        \includegraphics[width=0.99\linewidth]{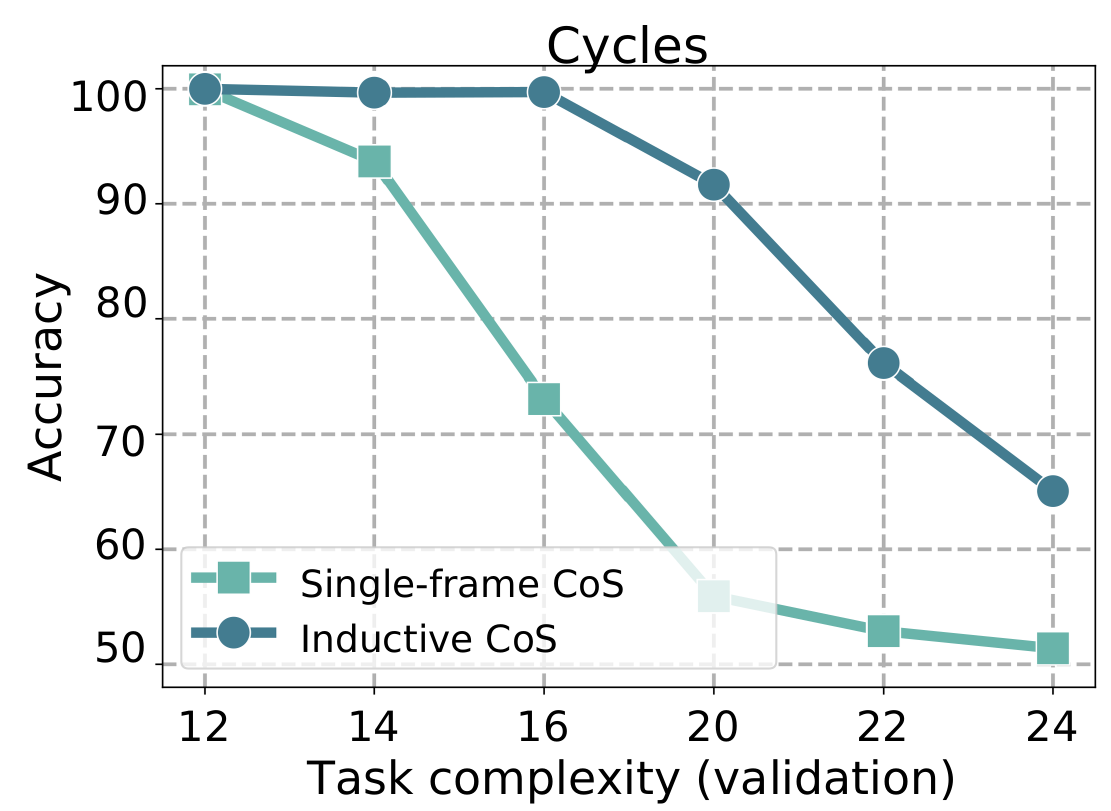}
        \captionof{figure}{The model is trained on Cycles 12 and tested on more complex instances.}
        \label{fig:ood-cycles-strings}
    \end{minipage}%
    \hfill
    \begin{minipage}[b]{0.60\textwidth} 
        \centering
        \setlength{\tabcolsep}{5pt} 
        \small
        \begin{tabular}{llcc} 
            \toprule
            \multirow{2}{*}{\textbf{Dataset}} & \multirow{2}{*}{\textbf{Method}} & \multicolumn{2}{c}{\textbf{Accuracy (\%)}} \\
            \cmidrule(lr){3-4} 
            & & \textbf{ID} & \textbf{OOD} \\
            \midrule
            \multirow{2}{*}{Maze 24 (Rect.)} & single-frame CoS & \textbf{100.0} & 54.4 \\
            & inductive CoS & 99.8 & \textbf{99.8} \\
            \midrule 
            \multirow{2}{*}{PVR Grid 7x7} & single-frame CoS & \textbf{100.0} & 48.6 \\
            & inductive CoS & 99.5 & \textbf{82.2} \\
            \bottomrule
        \end{tabular}
        \captionof{table}{Out-of-distribution (OOD) performance comparison for maze 24 (rectangular) and PVR grid 7x7 datasets. The inductive CoS demonstrates superior OOD generalization compared to the single-frame CoS across both tasks.} 
        \label{tab:ood-comparison} 
    \end{minipage}
\end{figure}

For the cycles task, we control task complexity via the number of nodes. In particular, for OOD experiments, we consider training on samples with 12 nodes and then testing on samples with a higher number of nodes and thus higher complexity. The results are visualized in Figure \ref{fig:ood-cycles-strings}. For the maze tasks, we keep the maze size the same. Instead, we create a dataset of easier samples for training (e.g., if the source and sink points are connected their distance is less than or equal to $30$) and use the main task dataset for validation. We explain the OOD training datasets for the maze tasks in more detail in Appendix \ref{app:dataset-generation}. The OOD results for the rectangular maze task are shown in Table \ref{tab:ood-comparison}. It can be seen that the inductive CoS model achieves almost perfect accuracy on OOD samples while the single-frame model performs slightly better than random. For the PVR task, we use a $7 \times 7$ grid and $k=2,3,4$ images per row for training and $7 \times 7$ grid and $k=5$ images per row for OOD samples. As shown in Table \ref{tab:ood-comparison}, the inductive model achieves $\approx 80\%$ accuracy while the OOD performance of the single-frame model is at chance level. We present more OOD experiments in Appendix \ref{app:additional-exps}.

\begin{wraptable}{r}{0.5\textwidth} 
    \centering 
    \vspace{-\intextsep} 
    \setlength{\tabcolsep}{3.5pt}
    \begin{tabular}{lcc}
        \toprule
        \textbf{Method} & \textbf{ID (\%)} & \textbf{OOD (\%)} \\
        \midrule
        inductive CoS & \textbf{100.0} & \textbf{88.2} \\
        inductive CoS (only TF) & 99.9 & 85.2 \\
        inductive CoS w/o halting & 99.9 & 80.9 \\
        multi-frame CoS & \textbf{100.0} & 64.8 \\
        single-frame CoS & \textbf{100.0} & 64.8 \\
        \bottomrule
    \end{tabular}
    \caption{Comparison of in-distribution and average OOD accuracy for CoS variants on the cycles task.}
    \label{tab:ablation-cycles-ood}
    \vspace{-\intextsep} 
\end{wraptable}
\subsection{Ablations}\label{sec:ablations}
The success of the inductive CoS model can be attributed to several factors, such as increased supervision during training, the halting mechanism, and the combination of teacher forcing (TF) and training on the model's own output distribution. In this section, we use the cycles task as a reference and provide extensive ablation experiments to quantify the contribution of each component. The computational implications of these components are also discussed in Appendix \ref{app:compute-overhead}.

\textbf{Supervision.} To show the importance of how supervision is applied—as well as its extent—we compare the inductive and multi-frame CoS models. While both methods have the same amount of target supervision, the inductive CoS exhibits much stronger OOD generalization, as shown in Table~\ref{tab:ablation-cycles-ood} (see also Appendix~\ref{app:additional_ablations}).

\textbf{Teacher forcing.} Here, we assess how much our improved training procedure contributes to performance. Specifically, we consider a baseline, ``inductive CoS (only TF),'' which uses the standard training procedure: it relies solely on teacher forcing and is not trained on the distribution of generated frames. Compared to the full inductive CoS model, this variant suffers a $3\%$ drop in performance.

\textbf{Halting.} We design an ``inductive CoS w/o halting'' baseline where we simply set a fixed large number of steps, eliminating the need for a halting signal. Compared to the full model, we observe an $\approx 7\%$ improvement in OOD generalization when dynamic halting is used.
 While the gap is significant, it is clearly smaller than the difference between inductive and non-inductive methods.

\subsection{Staircase learning phenomenon}\label{sec:staircase}
\begin{figure*}[t]
     \centering
     \begin{subfigure}[b]{0.16\textwidth}
         \centering
         \includegraphics[width=\textwidth]{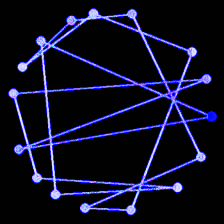}
         \caption{Iter = 2k}
     \end{subfigure}
     \hfill
     \begin{subfigure}[b]{0.16\textwidth}
         \centering
         \includegraphics[width=\textwidth]{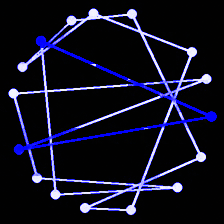}
         \caption{Iter = 6k}
     \end{subfigure}
     \hfill
     \begin{subfigure}[b]{0.16\textwidth}
         \centering
         \includegraphics[width=\textwidth]{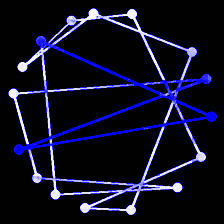}
         \caption{Iter = 8k}
     \end{subfigure}
     \hfill
     \begin{subfigure}[b]{0.16\textwidth}
         \centering
         \includegraphics[width=\textwidth]{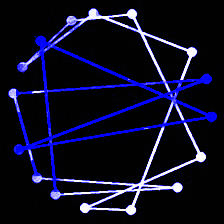}
         \caption{Iter = 10k}
     \end{subfigure}
     \hfill
     \begin{subfigure}[b]{0.16\textwidth}
         \centering
         \includegraphics[width=\textwidth]{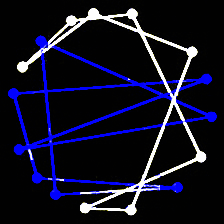}
         \caption{Iter = 13k}
     \end{subfigure}
     \hfill
     \begin{subfigure}[b]{0.16\textwidth}
         \centering
         \includegraphics[width=\textwidth]{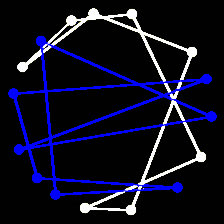}
         \caption{Iter = 50k}
     \end{subfigure}
     \vspace{-1.5mm}
    \caption{Generated sketches for an example at different stages during training. We have increased the contrast of the images for better visualization. It can be seen that the model first learns to color the rightmost node and then it goes one distance further each time during training.}
    \vspace{-1.5mm}
    \label{fig:staircase}
\end{figure*}

In our experiments with the single-frame model, we observed progressive hierarchical learning over the CoS image prediction task. Consider the cycles task as a running example. In the training set, we always color the cycle that passes the rightmost node. For sketch generations, we can observe that the model first learns to color the rightmost node. Then it learns to color the two neighbors that are connected to the initial node. Similarly, at each of the later stages of training, it learns to color roughly two more nodes (from the two sides). See Figure \ref{fig:staircase} for a visualization. 

These hierarchical learning phenomena have been previously observed and proven in theoretical settings, in particular, in the context of learning sparse Boolean functions where it is known as the staircase behavior \citep{mergedstaircase, abbe2023leap}. The staircase phenomenon states that if the target function has some hierarchical structure and is composed of different parts with different difficulties, learning easier parts first can boost learning for the harder parts. To be more precise, assume we have $n$ i.i.d. uniform Boolean variables $x_1, \ldots, x_n \in \{\pm 1\}$. It is well known that the difficulty of learning degree $k \leq n / 2$ parity function, e.g., $x_1x_2\cdots x_k$ increases as $k$ increases. In particular, if $k = \omega_n(1)$ then learning the parity function is not possible in polynomial time with regular MLPs (and statistical query methods) and learning complexity of degree $k$ parity for constant $k$ scales with $n^k$ \citep{abbe2020poly}. However, \citet{mergedstaircase} show that functions such as $x_1 + x_1x_2 + \cdots + x_1x_2\cdots x_k$ can be learned in $\tilde O(n)$ time. This is because the network can first learn the `easy' linear component $x_1$. Now, for learning $x_1x_2$ the model no longer needs to find two variables (which would scale with $\binom{n}{2}$), but it needs to only find $x_2$ since it has already learned that $x_1$ is in the support and can navigate the search space more efficiently to learn the terms like $x_1x_i$.

Considering the cycles task in the single-frame CoS again, coloring the first three nodes is a low globality function and can be learned easily. Coloring the next two nodes once the coloring of the first three nodes is learned is a low globality function (similar to $x_1\cdots x_i$ when $x_1+x_1x_2+\cdots x_1\cdots x_{i-1}$ is learned). More precisely, define $Y_k$ to be the coloring of all nodes (and edges) with a distance less than or equal to $k$ from the rightmost node ($2k+1$ nodes in total). $Y_1$ is a local target, moreover, coloring $Y_{k+1}$ correctly once $Y_k$ is learned is of constant globality degree. This staircase structure allows the model to learn $Y_1, Y_2, \ldots$ and finally the complete sketch sequentially during training as observed in Figure \ref{fig:staircase}.\footnote{For extra clarity, the visualization is enhanced in Figure \ref{fig:staircase}. Nevertheless, this behavior is also evident in the non-enhanced outputs, as shown in the third row of Figure~\ref{fig:staircase_cycles_full}.} Note that in the example of cycles task the intrinsic staircase structure of the single-frame model coincided with the CoS, however, that is not necessarily always the case. 

This example shows that the globality-degree does not satisfy the triangle inequality. In other words, we show that for input $X$ and target $Y$ and diverging globality degree (e.g., increasing number of nodes in the cycles task), there exists a single-frame sketch $X_1$ such that globality degrees of $X_1$ from $X$ and $Y$ from $X_1, X$ is constant. Thus, a single-frame CoS can make both efficient weak and strong learning (through the staircase effect) possible.

Moreover, this hierarchical learning phenomenon is not limited to the cycles task. Figure~\ref{fig:staircase_maze} demonstrates that a similar staircase behavior emerges in the more complex maze (rectangular) task. In this case, the model's behavior resembles a spreading ``cloud'' that progressively discovers contiguous areas of the maze. This is particularly noteworthy because the model is trained only on the final, fully solved maze configuration (shown in the second row of each column of Figure~\ref{fig:staircase_maze}).

In both the cycles and maze tasks, we observe a consistent pattern of the model first solving easier, more local aspects of the problem before progressively tackling more global structures. This aligns with the theoretical understanding of the staircase effect in learning sparse Boolean functions, which is now demonstrated in the visual domain.

\section{Related work}\label{sec:related}
In this section, we delve deeper into the related literature, examining it from multiple angles.
\subsection{Visual reasoning}\label{related:visual-reasoning} Different datasets have been introduced to evaluate various aspects of reasoning in the visual form. For instance, visual question answering (VQA) datasets such as \citep{vqa2015} ask questions about an image in natural language. These questions can rely on understanding the semantics in the images and basic reasoning operations such as counting. CLEVR \citep{johnson2017clevr} is a diagnostic VQA dataset made up of synthetic objects that removes spurious correlations that models exploit in traditional VQA datasets, in addition to disambiguating the types of the errors that the model can make. The reasoning operations considered in CLEVR include counting, comparison, attribute identification, and combinations of those. GQA \citep{hudson2018gqa} is another VQA dataset with real images focusing on answering compositional questions inspired by CLEVR. VCR \citep{zellers2019vcr} is focused on commonsense reasoning, asking deeper questions based on images (e.g., intentions of people and why an event is happening). 
CLEVRER \citep{Yi*2020CLEVRER:} focuses on understanding videos of CLEVR-like objects. In these videos, events such as collisions happen and different descriptive, explanatory, predictive, and counterfactual questions are asked. 
The CATER dataset \citep{girdhar2020cater} is focused on temporal reasoning where a video is given to a model and the model's task is to track a particular (potentially occluded) object throughout the video (similar to the classic cups-and-ball shuffle game).
ACRE \citep{zhang2021acre} is another dataset that aims to assess the performance of vision models in performing causal induction.
Winoground dataset \citep{thrush2022winoground} also focuses on compositional reasoning. Given two images and two captions with the same set of words, the task is to match them correctly which is shown to be very challenging for vision models. 
There are also datasets that require reasoning with a physical world model such as the Phyre dataset \citep{bakhtin2019phyre}.  Most of the aforementioned datasets rely on understanding semantics in an image, and in contrast to our proposed datasets, are easily solvable by humans. 

MathVista dataset \citep{lu2023mathvista} focuses on mathematical reasoning in the visual context. In this case, the questions are a combination of an image and text, however, the reasoning is predominantly textual, despite visual input.
Some datasets are inspired by human IQ tests such as Raven's progressive matrices \citep{santoro2018measuring-semi-raven, zhang2019raven, zhang2024far} that may be more challenging for humans compared to the classical VQA datasets, however, it is still not clear how one can increase the difficulty and the required number of reasoning steps for these datasets.  More recently, datasets such as MMMU \citep{yue2023mmmu, yue2024mmmu} and EMMA~\citep{hao2025can} have been proposed in order to assess the multi-modal reasoning performance of models. 

More visually similar to our work is the Pathfinder dataset \citep{linsley2018learning} which was introduced to show that convolutional neural networks (CNNs) cannot model long-range spatial dependencies well enough. The Pathfinder dataset in the text format was later included in the long-range arena benchmark \citep{tay2021long-lra} which aims to evaluate Transformers' ability to model long-range token dependencies. We note that our datasets do not necessarily focus on the distance between tokens (or the distance in the image) but rather the globality degree of the task and the number of reasoning steps required to solve the task. \citet{cherian2023deep} introduce vision and language tasks requiring reasoning, including graph tasks, but their reported models only achieve weak learning. In contrast, our tasks are explicitly designed for high globality, avoiding shortcuts or spurious correlations, making them impossible for large models to weakly learn without a CoS. Additionally, our tasks allow adjustable difficulty and focus purely on image classification, whereas \citet{cherian2023deep}'s dataset resembles VQA. To the best of our knowledge, the proposed datasets in this paper are unique in terms of having a scalable globality degree and number of reasoning steps while being challenging for humans as well.


\subsection{Reasoning with Transformers}\label{related:reasoning_with_transformers}
In recent years, reasoning capabilities of neural networks and in particular Transformers \citep{vaswani2017attention-transformer} have been extensively studied on a variety of topics ranging from completely synthetic symbolic datasets \citep{Zhang2021PointerVR, zhang2022unveiling} to algorithmic tasks \citep{velivckovic2022clrs} and more natural settings such as mathematical reasoning \citep{saxton2019analysing, lewkowycz2022solving-minerva}. These tasks usually have a combinatorial essence and hence an exponentially large input space which makes memorization-based learning approaches impossible for the Transformers. Another tool for assessing the reasoning abilities of neural networks is to test their OOD generalization performance to see whether they rely on superficial cues that do not work on OOD samples or rather they can compose the rules they have seen during training to generalize to OOD and often more complex examples. As a special case of OOD generalization, it has been observed that length generalization \citep{zaremba2014learning, lake2018generalization, hupkes2020compositionality}, generalizing to longer instances than what was seen during the training, is particularly challenging for Transformers even for simple arithmetic tasks such as parity, addition, and multiplication \citep{anil2022exploring-length, icml-version, lee2024teaching-arithmetic}. This challenge may be further aggravated in the settings where the input problem or its solution is longer than what the model has seen during training and hence the model has to deal with (mostly) unseen positions where it has been shown the absence or the use of different absolute or relative positional embeddings \citep{Shaw2018SelfAttentionWR, Dai2019TransformerXLAL} result in significant variations in length generalization performance \citep{kazemnejad2023impact}. Despite the efforts to understand the reasoning abilities in the symbolic domain, works in the visual domain have focused on more superficial forms of reasoning emphasizing understanding the semantics of the image. This is despite the fact that vision provides an excellent ground for OOD and length generalization experiments since one can easily depict more challenging examples with the same image resolution which removes the element of using suitable positional embeddings from the picture.

\subsection{Scratchpad and chain-of-thought}\label{related:cot-scratchpad} \citet{nye2021work} introduced the idea of scratchpads showing that training Transformers to output the intermediate reasoning steps in addition to the final solution can boost their performance on reasoning tasks such as arithmetic, math, and code understanding. 
Further, \citet{wei2023chainofthought}  show that pre-trained language models can perform step-by-step reasoning by merely seeing a few in-context examples referring to this as chain-of-thought (CoT). Later it was shown that pre-trained language models can also generate chains of thoughts only by prompting to do so \citep{kojima2023large}. \citet{locality} provide theoretical explanations on the effectiveness of scratchpads using the notion of globality concept.\footnote{In particular, for the symbolic version of the cycles task studied in \citet{locality}, it is shown experimentally that the learning complexity grows rapidly with the number of nodes ($2n$) increasing.} They also introduce a variant of the scratchpad method for multi-step reasoning problems that uses a dynamic masking technique to only attend to the input question and last step which causes the model to demonstrate superior length generalization performance. 

Moreover, there have been recent efforts to use the visual form of scratchpad and chain-of-thought in multi-modal models. In particular, visual-CoT \citep{shao2024visual-cot} takes an image with a question in the input. During the generation of the output, it first predicts a bounding box in the image that may have important information inside, and then the model focuses on that part of the image to answer the question better. This idea could be useful in cases where the answer can be given using a small part of a high-resolution image (e.g., a text written with a small font in the corner of an image). However, this work does not deal with hard reasoning tasks that require multiple reasoning steps nor produce images as scratchpad/CoT. The recent work of \citet{hu2024visual-sketchpad} introduces the notion of sketchpad. For a question (consisting of text and visual components) they use a set of visual operations and tools including 
drawing lines, placing bounding boxes with object detection models, and using Python to produce plots to generate a sketch that can potentially facilitate the reasoning process. The main difference with our works is that we focus on visual tasks that have a high globality degree and require multiple reasoning steps to solve, whereas \citet{hu2024visual-sketchpad} do not consider visual tasks that require multi-step reasoning. As a result, our approach is to use chain-of-sketches to make the tasks learnable, while in their case is to use tools (e.g., object detection or plot creation using Python) to generate images that can guide the model. As a result, in our case, the models can generate a sequence of frames that correspond to reasoning steps where each image is generated freely by the model; while the sketchpad method can only generate a single sketch in a limited manner by using a set of predefined tools and operations. 

In addition, there have been related developments in other subfields. For instance, \citet{yang2024video} and \citet{bai2024sequential} treat video as a unified interface for diverse tasks, using frame-by-frame generation for real-world decision-making applications such as robotics and self-driving cars. While there are similarities in modeling, our contributions differ significantly. We introduce high-globality image classification tasks that current vision models, even large-scale ones trained on extensive data, fail to weakly learn, revealing fundamental limitations. We demonstrate that these tasks become learnable only when using a CoS. Additionally, we extend theoretical concepts such as globality degree and establish a direct empirical link between task globality and model performance.

\subsection{Recurrent architectures}\label{related:recurrent} Several works have introduced a recurrent component into Transformer architectures \citep{dehghani2019universal, hutchins2022-block-recurrent, giannou23a-looped}. Notably, Universal Transformers \citep{dehghani2019universal} use shared weights between transformer layers and also use an adaptive computation time \citep{graves2016adaptive} by varying the number of times that the transformer layer is applied. We note that the CoS model proposed in this paper is significantly simpler than the architectures above. This is because in the proposed CoS model, due to the Markovian modeling of CoS frames, there is no sort of adaptive compute time involved at training time, and the model is simply supervised to generate the next frame given the current frame without any history (see Appendix \ref{app:model-implementation}). Further, the halting mechanism is supervised during training.


\section{Conclusions and future directions}\label{sec:conclusion}
We summarize the contributions of our work. (1) We explored the concept of locality/globality in the visual domain, extending the globality degree definition \citep{locality} to vision tasks. We then introduced five global vision tasks that have high globality and are hard to learn for ViT models despite reasonable model size scaling and pre-training. 
(2) We proposed various chain-of-sketch methodologies to break the high globality of such tasks by defining subtasks that are in the form of visual frames. We showed that training models with a single-frame CoS can make the high-globality task learnable by reducing the globality degree in a single step. However, this does not yield strong OOD performance or favorable model size tradeoffs. 
(3) We introduced the inductive CoS for learning multi-frame CoS in a Markovian manner such that each intermediate frame is learnable inductively from the previous frame. We show that the inductive CoS can learn the tasks with smaller model sizes for which the single-frame CoS fails. The inductive CoS also shows superior OOD performance and thus reasoning capabilities.

{\bf Future work.} 
With advancements in the field, we expect global reasoning to become increasingly important in the visual domain. In particular, models with text and multi-image input/output modalities will likely enable interleaved reasoning across visual and symbolic representations. This capability can support tasks involving geometry problems, visual puzzles, high-globality images such as maps, higher-order spatial relationships across video frames, and even applications like autonomous driving. A key challenge, however, is the lack of datasets containing the intermediate chain-of-sketch (CoS) frames required for model training. One promising data source is video---for example, recordings of individuals solving geometry problems or performing multi-step visual tasks. Moreover, with the growing capabilities of multi-modal models, it may soon be possible to generate CoS through in-context learning and prompting rather than relying solely on supervision during training. This process can be further enhanced using reinforcement learning, as is commonly done for chain-of-thought reasoning in text \citep{zelikman2022star, shao2024deepseekmathpushinglimitsmathematical}. The goal of this work is to take a first step toward demonstrating the necessity of chain-of-sketch methodologies for global visual reasoning.


\bibliographystyle{abbrvnat}
\bibliography{main}

\newpage
\appendix
\onecolumn 
\section{Training details} \label{app:training-details}
We first resize the input (and the CoS frames) to $224 \times 224$ resolution. We then use a patch size of $16 \times 16$ to partition the images into $196$ patches for all models before giving them to the ViT backbone of the models. The models are evaluated on 10k validation samples.   

For training, we use AdamW \citep{adamw} optimizer with weight decay $0.05$ and learning rate $0.0003$. For the learning rate, we first use a linear warm-up to increase the learning rate from $0$ to $0.0003$. Afterward, we use a cosine schedule with $3e-6$ as the end value for the rest of the training. The linear warm-up is applied for $5\%$ of the training time (e.g., $2500$ iterations if the total number of iterations is set to $50k$) and the cosine annealing is applied for the rest $95\%$ of the training time. 

Each of our experiments has been run on 8 H100 or A100 GPUs and we use a batch size of 1024 for each iteration. The whole project has an approximate total consumption of 160k GPU hours.

\subsection{Hyperparameter tuning and sensitivity}
Note that we have different settings in our experiments where we vary our methodology, model size, and dataset. This gives rise to a combinatorially large number of experiments that each require their own hyperparameter tuning which is infeasible. Nevertheless, we tried sweeps over learning rate and weight decay for some of our in-distribution settings. We found that our models and methods are relatively robust to learning rates in the range of $0.0001$ to $0.0005$ and weight decays in the range of $0.01$ to $0.1$. In particular, we observed that a learning rate of $0.0003$ and weight decay of $0.05$ work well in all of the tested settings, and therefore we use this combination for all experiments reported in this paper. Similarly, for the batch size we tried batch sizes $1024$ and $2048$. We observed that batch size $2048$ converges with a slightly smaller number of iterations, however, longer wall-clock time. Thus, we decided to use batch size $1024$ across all of our experiments.

\section{Model implementation} \label{app:model-implementation}
We use a ViT backbone for all of our methods. We use four standard sizes for the ViT model: small, base, large, and huge. ViT models differ in the number of layers, embedding dimension (hidden size), MLP size, and number of heads, see Table \ref{tab:model-size-standard} for the details. Note that these model sizes are standard \citep{dosovitskiy2020image, touvron2021training-vit-small}, further, we always use 196 patches of size $14 \times 14$ for all model sizes. 

\begin{table}[h!]
\centering
\caption{ViT model sizes and specifications}
\begin{tabular}{@{}cccccc@{}}
\toprule
\textbf{Model}   & \textbf{Hidden size} & \textbf{Number of layers} & \textbf{Attention heads} & \textbf{MLP size} & \textbf{Parameters} \\ \midrule
ViT-Small  & 384   & 12  & 6   & 1536  & $\sim$22M  \\ 
ViT-Base   & 768   & 12  & 12  & 3072  & $\sim$86M  \\ 
ViT-Large  & 1024  & 24  & 16  & 4096  & $\sim$307M \\ 
ViT-Huge   & 1280  & 32  & 16  & 5120  & $\sim$632M \\ \bottomrule
\end{tabular}
\label{tab:model-size-standard}
\end{table}

Finally, note that currently, the CoS frames in our tasks are deterministic. As a result, our image generation models are also deterministic. We expect that for more complicated tasks a random generation model for the CoS frame(s) may be more suitable. One can use different solutions in that case. For example, if there is a constant (say 2) number of possible CoS frames, the model can try to generate all these possibilities with a bipartite matching loss similar to the DETR work \citep{carion2020end-defr}. Alternatively, one can add a noise variable $z$ for the generation part to add randomness such that the output CoS image is conditioned on the input image of the model.  
Nevertheless, we emphasize that the focus of this work is on the idea of the Chain-of-Sketch and the need for it and modeling choices (e.g., CoS model) and not on image generation methods and hence we have used a simple generation method.


\subsection{Training procedure for the inductive CoS model}\label{sec:training_procedure}
Consider an input image $x = f_0$ with CoS frames $f_1, \ldots, f_T$ and label $y$ from the training set. The recurrent module $\mathcal{M}$ can be trained by teacher forcing, i.e., the model can be trained on samples of the type $f_i \to (\hat f, \hat y, \hat h) = (f_{i+1}, y, \mathbbm{1}(i+1=T))$. The issue with this training method is that the recurrent module $\mathcal{M}$ is solely trained on samples from the training distribution. However, during inference where CoS frames are not available, the recurrent module $\mathcal{M}$ will use its own generated frames as the input to itself. This discrepancy between the input distribution of the module at training and at test time could deteriorate the model's performance. We initially implemented our model with teacher forcing training described above and observed that the model can learn all the tasks rather well. The issue, however, is that, especially at the beginning of training, the predicted frames are not guaranteed to be close enough to the training distribution to perform well during inference. Hence, we decided to use the following alternative approach. We provide a frame $f_i$ from the training set to the model to get the predicted next frame $\hat f_{i+1}$ along with the predicted label and halt variables $\hat y_{i+1}, \hat h_{i+1}$. We then provide the predicted CoS frame to get the next frame $\hat f_{i+2}$ along with the next prediction for the label and halt variable $\hat y_{i + 2}, \hat h_{i + 2}$. Finally, we compute the loss for all $\hat f_{i+1}, \hat y_{i+1}, \hat h_{i+1}, \hat f_{i+2}, \hat y_{i + 2}, \hat h_{i + 2}$ and their corresponding ground truth values in the training set. Note that we consider $\hat f_{i+1}$ an independent input for the model and no gradient is backpropagated through it. As a result, during training the model's input comes from both the training distribution and the distribution of the generated frames of the model itself. We found that this method gives a considerable increase in the training speed of our models and decided to use this method for our experiments. 

We note that this problem of discrepancy between training distribution and generation distribution during inference has been previously observed in settings such as text generation in recurrent neural networks (RNNs) and reinforcement learning, for instance in \citep{bengio2015scheduled} which proposed a scheduled sampling approach as follows: for each token, they sample it either from the train distribution with probability $\epsilon$ or from the model itself with probability $1-\epsilon$ and use a schedule (e.g., linear or exponential) to reduce $\epsilon$ during training. We note that our setting is simpler as the modeling in our setting is Markovian and each CoS frame is only generated based on the previous one and not the whole history in contrast to RNNs. Hence, our simplified approach of having a fixed rate of samples from the training and generation distributions worked well. 
    
We also note that one could use a large predetermined number of steps instead of using a halting mechanism. For this, one needs to supervise the model such that if the final frame is given to the model, the model outputs the same final frame without changes.

\subsection{Compute overhead for different methods} \label{app:compute-overhead}
Taking as a reference the No-CoS baseline, we can look at the compute overhead of each proposed model. The single-frame model only adds a linear layer that is used to map the latents to the reconstructed pixels. This has a negligible impact on compute, as the backbone is a lot bigger. In terms of training time, this does not produce noticeable overheads. Similarly, the multi-frame CoS adds a linear layer for each frame of the CoS, making the compute overhead dependent on the length of the trajectory. This, together with the fact that the data loader needs to load all the target sketches for each sample, ends up causing a large increase in training time (e.g., 20x longer for maze rect. 32 dataset). On the other hand, the inductive CoS model does not need additional projection layers, therefore it does not increase the number of parameters. In the case of ``CoS (only TF)'' the training time is roughly the same as the single-frame model as it is trained with pairs of sketches. However, in the full inductive CoS model that also forwards generated frames, the training time doubles w.r.t. the single-frame model. At inference time, no-CoS, single-frame and multi-frame CoS have the same complexity, while the inductive CoS model scales according to the size of the problem at hand, which is a desirable property. Moreover, this overhead is completely offset by the ability to use smaller models.

\section{Dataset generation} \label{app:dataset-generation}
For each task, we generate a dataset with 1M training samples and 10k validation samples. For both validation and training sets half of the samples have 0 and half have 1 as the label meaning that the baseline accuracy for this dataset is $50\%$.  It is important to note that these datasets are generated in a way that minimal spurious correlations are introduced, otherwise, the model might have used those correlations for weak learning and achieving better-than-random accuracies. We explain the generation algorithm for each of the datasets below. 
\subsection{Cycles task}
The cycles task consists of $2n$ nodes and $2n$ edges such that the $2n$ edges either form a cycle of size $2n$ or two cycles of size $n$. The label for the former is 1 (connected) and for the latter is 0 (disconnected). 

For the cycles task, we generate images of size $448 \times 448$. We further choose the nodes randomly on an invisible circle with a radius of 220. Constraining the nodes to be on an invisible circle ensures that no three points are (almost) collinear. In this case, each node on the circle can be specified by its angle $\theta$. We also ensure that every two nodes are at least $\epsilon$ radians apart on the circle. To generate the points, we select $n-1$ random numbers between $0$ and $2\pi - n\epsilon$ and then sort them: $x_1 \leq x_2 \leq \ldots \leq x_{n-1}$. We also select a parameter $\beta$ randomly in $[0, 2\pi]$. Finally, we define the points to be 
\begin{equation*}
    \theta_1 = \beta, \theta_2 = \beta + x_1 + \epsilon, \theta_3 = \beta + x_2 + 2\epsilon, \ldots, \theta_n = \beta + x_{n-1} + (n-1)\epsilon.
\end{equation*}
One can easily check that $\theta_{i+1} - \theta_{i} = \epsilon + (x_i - x_{i-1}) \geq \epsilon$ (where we take $x_0=0$). 
Also, $\theta_n = \beta + x_{n-1} + (n-1)\epsilon \leq  \beta + (2\pi -\epsilon) = \theta_1 + 2\pi - \epsilon$ showing that each two consecutive points have a minimum distance of $\epsilon$ radians on the circle.

\paragraph{Chain-of-sketch.} For the multi-frame CoS of the cycles task, we first color the rightmost node in blue for the first frame. At each later frame, we color (at most) two more nodes/edges from both sides. In other words, the $k+1$th frame includes all the nodes/edges with a distance less than or equal to $k$ from the rightmost node colored in blue. Consequently, the last CoS frame which is the same as the single-frame CoS for this task colors the cycle that passes through the rightmost node in blue (whether the label is 0 or 1). We note that this resembles what humans would naturally do by following one of the cycles (with a pen for instance). 

\paragraph{OOD samples.} For the OOD experiments, we simply use the cycles tasks with a different number of nodes for out-of-distribution evaluation. We note that currently, we only generate the cycles task datasets with up to 24 nodes. We believe one has to increase the image resolution for a larger number of nodes to still keep the task visually meaningful.
\subsection{Strings task}
The generation process of the strings task is similar to the cycles task. We have $2n$ invisible nodes (called anchor nodes) and these $2n$ nodes are connected with $2n$ 3rd-degree B\'ezier curves such that we have either two strings (label 0) or a single string (label 1), equiprobably. For this task, we also generate images of size $448 \times 448$ and choose the anchor points on an invisible circle of radius $200$ with the same process described for the cycles task. 

Next, we explain how B\'ezier curves are drawn. To specify a $k$th degree B\'ezier curve between points $A$ and $B$ one needs to first define $k-1$ control points $C_1, \ldots, C_{k-1}$. To simplify the notation, we define $P_0 = A, P_1=C_1, \ldots, P_{k-1}=C_{k-1}, P_k=B$. In this case, the B\'ezier curve is given by
\begin{equation*}
    \rmB(t) = \sum_{i=0}^k \binom{k}{i}(1-t)^{k-i}t^i P_i = (1-t)^k P_0 + k(1-t)^{k-1}tP_1 + \cdots + k(1-t)t^{k-1}P_{k-1} + t^kP_k 
\end{equation*}
for $t \in [0,1]$. In particular, the cubic B\'ezier curves between points $A$ and $B$ with control points $C_1, C_2$ is given by 
\begin{equation*}
    \rmB(t) = (1-t)^3A + 3(1-t)^2tC_1 + 3(1-t)t^2C_2 + t^3 B \quad t \in [0,1].
\end{equation*}
We need to specify two control points for each B\'ezier curve. We also want the curve to look continuous to have smooth strings and as a result, we need the first derivative of the curve to be well-defined. Note that the derivative of the cubic B\'ezier curve above is given by 
\begin{equation*}
    \rmB(t)' = 3(1-t)^2(C_1 - A) + 6(1-t)t(C_2-C_1)+ 3t^2(B-C_2) \quad t \in [0,1].
\end{equation*}
More specifically, we need to ensure that the derivatives are the same at the points that two B\'ezier curves meet, i.e., at $t=0$ and $t=1$ where the derivative is equal to $\rmB(0)' = 3(C_1 - A)$ and $\rmB(1)' = 3(B - C_2)$ respectively. To define these points, further assume that points $A, A'$ and $B, B'$ are connected with cubic B\'ezier curves (i.e., we want a continuous curve that passes through $A', A, B, B'$). Also, to disambiguate the control points, we use notation $C_{1}(X, Y)$ the first control point for the B\'ezier curve between $X$ and $Y$ (similarly for $C_2$). Given the derivatives computed above, we need to ensure that $C_1(A, B) - A = A - C_2(A', A)$ and $B - C_2(A, B) = C_1(B, B') - B$. To satisfy these conditions we take 
\begin{equation*}
    C_1(A, B) = A + \alpha(B-A'),\;\; C_2(A, B) = B - \alpha(B'-A),
\end{equation*}
for a constant value of $\alpha$. One can easily check that defining the control points with the equation above makes the first derivative of the curve well-defined and the curve continuous. For instance, to check the continuity at $A$ we have 
\begin{equation}
    C_1(A, B) - A = \alpha(B-A') = A - (A-\alpha(B - A')) = A - C_2(A', A).
\end{equation}
For our datasets, we use the value $\alpha=0.25$ as we find it empirically to produce suitable samples.

\paragraph{Chain-of-sketch.} In order to generate the multi-frame CoS for the strings task, we use a similar procedure to what we do for the cycles task. We first color the rightmost anchor node. At each of the later frames, we extend the colored string from both sides by going to the next anchor nodes. Thus, the $k+1$th frame colors the string that passes through the rightmost anchor node up to the anchor nodes that have distance $k$ from the rightmost starting anchor node. Analogous to the cycles task, the last CoS frame (equivalently the single-frame CoS) for this task colors the string that passes through the rightmost node in blue. This is also similar to what humans would do by following one of the strings. 

\paragraph{OOD samples.} Similar to the cycles task, we simply use strings task of different sizes (number of anchor points) for the OOD experiments. Also, as one increases the number of anchor points, one has to increase the image resolution to keep the task feasible to solve. 
\subsection{Maze tasks}
First, we explain the logic shared by both the rectangular and circular mazes. Afterward, we discuss the specifics of these two versions. Our mazes always have two parts a source/start cell colored in blue and a sink/end cell colored in red. The source and sink cell are either in one component (label 1) or not (label 0). Both rectangular and circular mazes can be viewed as graphs where each cell is a graph node and two nodes are connected if they are adjacent and there is no wall between them. We first note that each of our maze components is a tree, which ensures that all cells in one component are connected by a unique path. To generate our maze samples, we first generate a maze that has a single fully connected component where any two cells are connected by a unique path (the corresponding graph is a tree). Then we select the start and the end cells, and finally, we add a wall to the maze to break the maze into two components. We will next explain each of these parts in more detail. 

There are several algorithms for generating a maze with one component. These algorithms differ in their generation speed, the average length of the paths in the maze, and the branching factor of the maze which specifies the average number of branches in the paths of the maze. Considering these factors, we have decided to use Kruskal's algorithm \citep{kruskal1956shortest} for generating the mazes. Kruskal's algorithm starts with a maze where all possible walls are drawn. Then, at each step, the algorithm selects a wall randomly and removes it if the two neighboring cells of this wall were not previously connected. This algorithm is continued until the maze is fully connected. For the start point of the maze, we select one of the cells adjacent to the first wall selected by the algorithm. We then compute the distances of all the cells to the start cell and in particular the maximum distance $d_{\max}$. Then we uniformly choose the target distance in $[d_{\max} - 20, d_{\max}]$, and select the end cell such that its distance from the start cell is equal to the target distance. This approach ensures that the distance between the start and end cells is random and also large enough to make the maze challenging. Finally, we insert a wall in the maze to make two components. If the label is 0 we put this wall in the unique path that connects the start and end cells, otherwise if the label is 1 we put the wall such that the path between the start and end cells remains intact. In addition to that condition,  we select the wall that minimizes the difference in the size of the two resulting components (i.e., our goal is to have components of the same size ideally).

\paragraph{Chain-of-sketch.} To generate the multi-frame CoS of the maze datasets we basically simulate a breadth-first search (BFS) from the start cell. We start from the start/source cell and for each CoS frame, we color any cell that is at a maximum distance of $10$ from the previously colored cells until we reach the end of the maze component or the solution is found. Note that adding cells of distance $1$ at each step would have resulted in too many frames. What we do for generating the CoS frames is similar to BFS. In particular, if we define $d_{\mathrm{target}}$ equal to the distance to the end cell if they are in the same component and the maximum distance from the start cell otherwise, then the $k$th CoS frame colors all cells within distance $\min\{10k, d_{\mathrm{target}}\}$ from the start cell (note that we end the search once the target is reached or the whole component is explored). In this case, also, the single-frame CoS is the same as the final CoS frame in the multi-frame CoS. 

\paragraph{OOD samples.} Generating OOD examples for the maze datasets is more challenging than the cycles and strings datasets since one cannot simply change the maze size as it will cause resolution inconsistencies. Thus, for the maze dataset, we use the same maze size for the training set and OOD samples. Instead, we use \textit{easier} samples for training and use the normal maze task dataset described above for OOD evaluation. To generate easy samples, we choose our target distance between the start and the end cell uniformly from $[10, 30]$ which is significantly smaller than $[d_{\max} - 20, d_{\max}]$ used for the main dataset where $d_{\max}$ was the maximum distance from the start cell (see above). The latter ensures that the number of CoS frames required to solve the task when the nodes are connected is less than or equal to $3$ during training. Further, instead of trying to split the maze into two components of the same size, we try to add the wall such that the size of the component that includes the start cell is closest to $\frac{30}{d_{\max}}(\frac{\mathrm{number\;of\;cells}}{2})$. By doing the latter, we make sure that the search space seen during training (size of the component including the start cell) is smaller than the main dataset, and hence samples are easier. 

Next, we explain details specific to rectangular and circular mazes. 
\subsubsection{Rectangular maze specifics}
Rectangular mazes are primarily specified by a number $n$ which indicates the number of rows and columns of the maze resulting in $n^2$ cells. E.g., maze (rect.) 32 has 1024 cells. Also, note that each cell in the rectangular maze has at most 4 neighbors.
\subsubsection{Circular maze specifics}
Circular mazes are organized into a number of concentric rings and are primarily specified by the number of rings. The zeroth circle only includes the center of the maze and is not counted in the number of rings. The first ring contains 6 cells. For each of the next rings the number of cells is kept fixed or is doubled. Also note that the center cell in the circular maze has 6 neighbors and other cells can also have up to 5 neighbors. 

\subsection{Image PVR tasks}
In all of the experiments in this paper, we used $7 \times 7$ grids. First, we randomly select one MNIST digit as the pointer from $\{1,2,3,4,5,6\}$. Next, for each of the subsequent rows, we put $k \leq 7$ CIFAR-10 images such that each image is equiprobably sampled from the airplane or car classes. $k$ is a fixed number of a specific task. The target is the result of an aggregation function applied to the row indicated by the pointer. For the main experiments of the paper, we use parity as the aggregation function, i.e., whether the number of occurrences of the first image (from left in the row) in the row is even (label 0) or odd (label 1). For experiments in Appendix \ref{app:pvr_majority}, we also use majority as the aggregation function where the label is 1 if the object appearing first (from left in the row) is in strict majority.  

\textbf{Chain-of-sketch.} We only define CoS for the parity aggregation function. For chain-of-sketch, we first hide all parts of the image except the indicated row as the first image of the chain. Next at each step, we compute the parity cumulatively: we check whether the object we are checking is the same as the first object or not, if it is, we update the parity and write the updated number on that image, otherwise, we put an 'x' on the image. We continue this until all images in the row are processed. 

\textbf{OOD samples.}
For the train distribution, we simply mix the PVR task for the number of images per row $k=2,3,4$. For the OOD evaluation set, we check the models on the task for $k=5$.

\subsection{Prompts and details of experiments with multi-modal LLMs}
\label{app:prompts}

In our experiments with GPT models, we tried various image resolutions: $256\times256$, $512\times512$, $1024\times1024$, and $2048\times2048$, as well as both high and low resolution settings of the OpenAI API. In our experiments, using high-resolution mode on $512\times512$ images seemed to work best, and we report those results in this paper. We also experimented with different prompt formulations and found that detailed prompts tended to improve performance. The exact prompts are provided in Table~\ref{tab:prompts}. We also note that tool calling was disabled in our API calls and the reported results. Also we used the following checkpoints of the models: \texttt{gpt-4o-2024-08-06}, 
\texttt{o4-mini-2025-04-16}, and \texttt{o3-2025-04-16}. 

\begin{table}[htbp]
\centering
\begin{tabular}{p{3cm} p{12.6cm}} 
\toprule
Setting & Prompt \\
\midrule
Cycles with no in-context examples & This image contains \texttt{<NODES>} nodes which are shown by large white filled dots. There are \texttt{<NODES>} edges between the nodes. Does the image contain a single closed cycle or two separate closed cycles? (A cycles is a closed path which passes through a number of nodes so when there is a single cycle all nodes are reachable from each other and when answer is 2 there are two separate connected components in the graph.) Provide your reasoning and end your answer with ``output: 1'' or ``output: 2''. \texttt{<task img>} \\
Cycles with 4 in-context examples & Each image contains \texttt{<NODES>} nodes which are shown by large white filled dots. There are \texttt{<NODES>} edges between the nodes. Does the fifth (last) image contain a single closed cycle or two separate closed cycles? (A cycles is a closed path which passes through a number of nodes so when there is a single cycle all nodes are reachable from each other and when answer is 2 there are two separate connected components in the graph.) For instance, in the first two images, there are two separate cycles and in the third and fourth images there is a single cycle. Provide your reasoning and end your answer with ``output: 1'' or ``output: 2''. \texttt{<img ex.1><img ex.2><img ex.3><img ex.4><task img>} \\
Maze (rect.) with no in-context examples & In the maze, is there a path between the blue square to the red square which doesn't cross any walls (shown by thick black lines)? Try to reason. For example, you can start from the blue square and try to see whether you reach the red square or not. End your answer with a Yes or No. \texttt{<task img>} \\
Maze (rect.) with 4 in-context examples & You are given 5 maze images where the task is to determine whether the blue and red square are connected or not, i.e., if there's a path not blocked by walls (shown by thick black lines). For example, in the first two images there is no valid path between the blue and red cells. While in the third and fourth images there is a path. Now, for the fifth image, are the blue and red cells connected? Provide your reasoning and end your answer with a Yes or No. \texttt{<img ex.1><img ex.2><img ex.3><img ex.4><task img>} \\
\bottomrule
\end{tabular}
\vspace{0.5cm}
\caption{Prompts used for the results of Table \ref{tab:llm-results}. \texttt{<NODES>} is replaced by the number of nodes in the image. Also, \texttt{<img ex.1>...<img ex.4>} are the in-context image examples and \texttt{<task img>} is the task image.}
\label{tab:prompts}
\end{table}

\section{Additional experiments} \label{app:additional-exps}

\subsection{PVR experiments clarifying the concept of globality}\label{app:pvr_majority}



\begin{figure}[h]
    \centering
    \includegraphics[width=0.6\linewidth]{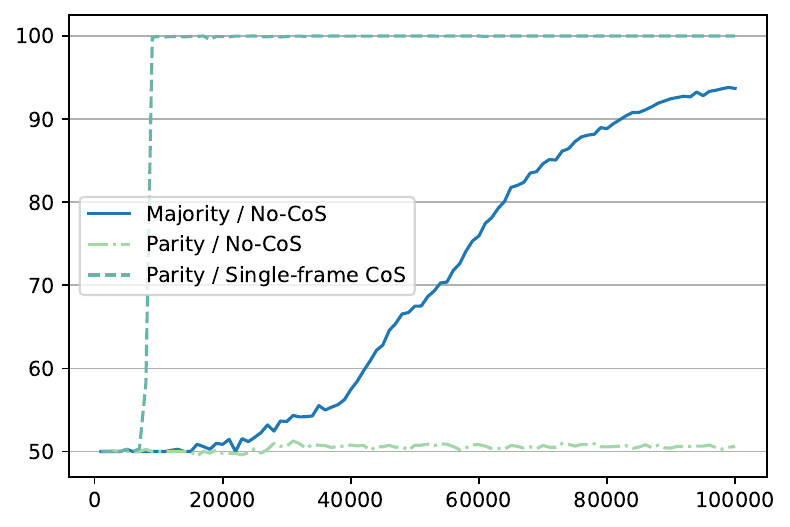}
    \caption{PVR with majority function is learnable by the ViT-B model without CoS. Whereas, PVR with parity function is not learnable by ViT without CoS regardless of the model size.}
    \label{fig:pvr-majority}
\end{figure}

To further clarify the notion of globality, we consider the following variant of the PVR task: instead of using the parity function on the indicated row, we use the majority function on the selected row, such that the label is 1 if the class that appears first (from left) is in the (strict) majority and $0$ otherwise. Parity and majority both depend on all of the inputs. However, the majority function, in contrast to the parity function is not a global function, as even seeing one element of the majority provides non-trivial information on the output. This is while seeing all except one member of a parity function provides no information as the last member can equiprobably change the outcome. To see this better in action we check if the same PVR task in $7\times 7$ grid with 4 numbers per row and majority function aggregation is learnable or not and we compare it with the parity function. In Figure \ref{fig:pvr-majority} we see that the majority variant of the task is learnable without CoS, whereas PVR with the parity aggregation is only learnable when (at least) a single-frame CoS is used. Here, we have conducted the experiments using the ViT-B model size, but we have also confirmed that the PVR task with parity aggregation remains hard to learn without CoS even if ViT-L or ViT-H are used. 

\subsection{Additional model scaling experiments on maze (circular)}
\begin{figure}[htb]
    \centering    \includegraphics[width=0.5\textwidth]{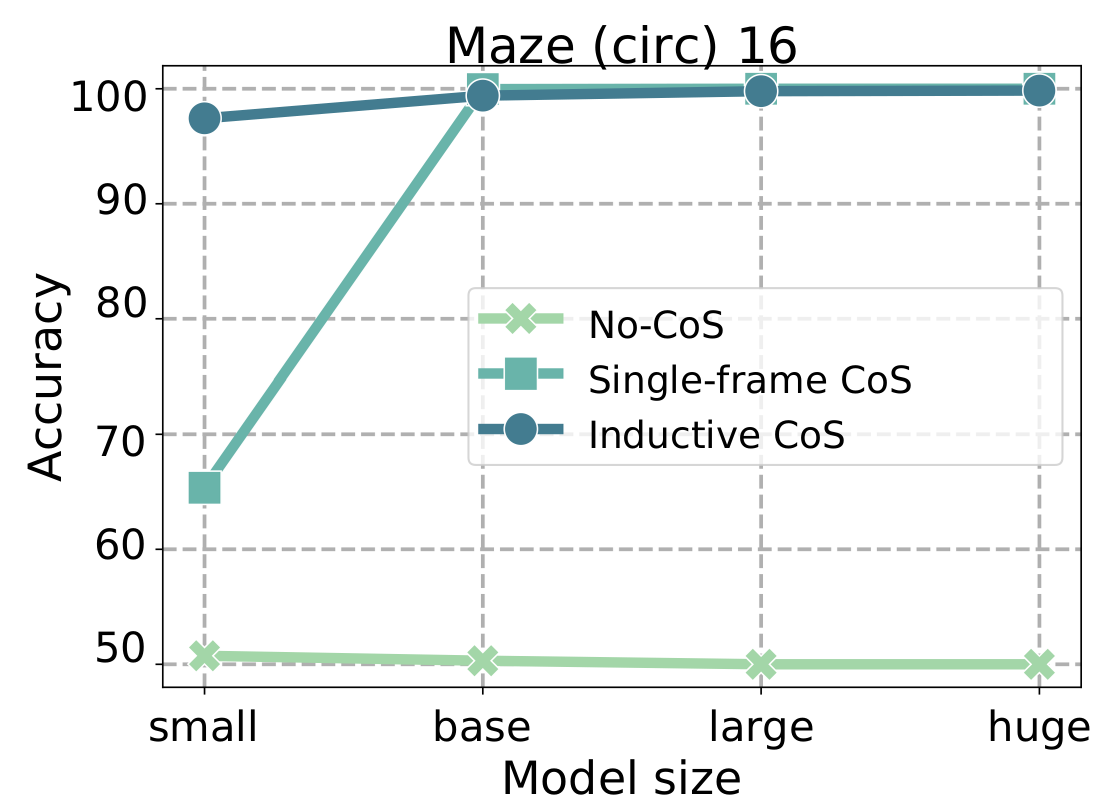}  
    \caption{Maze (circular) 16 model size experiments. The model behavior is similar to the maze (rectangular) dataset. CoS is on par with Single-frame for B, L and H but has a significant advantage on S.}
    \label{fig:maze_circ_16_ms}
\end{figure}

In the model scaling experiments conducted on the maze (circular) dataset in Figure~\ref{fig:maze_circ_16_ms}, we observe a similar behavior to that seen on maze rectangular. For larger model sizes (Base, Large, and Huge), both the CoS and the single-frame achieve near-perfect accuracy. However, the CoS model particularly shines when it comes to smaller models. With the ViT-Small model, the CoS approach significantly outperforms the single-frame, yielding a performance improvement of more than $30\%$. This indicates the effectiveness of the inductive method in handling resource-constrained settings.

\subsection{Relationship between model size and OOD generalization}
\begin{figure}[htb]
    \centering
    \includegraphics[width=0.9\textwidth]{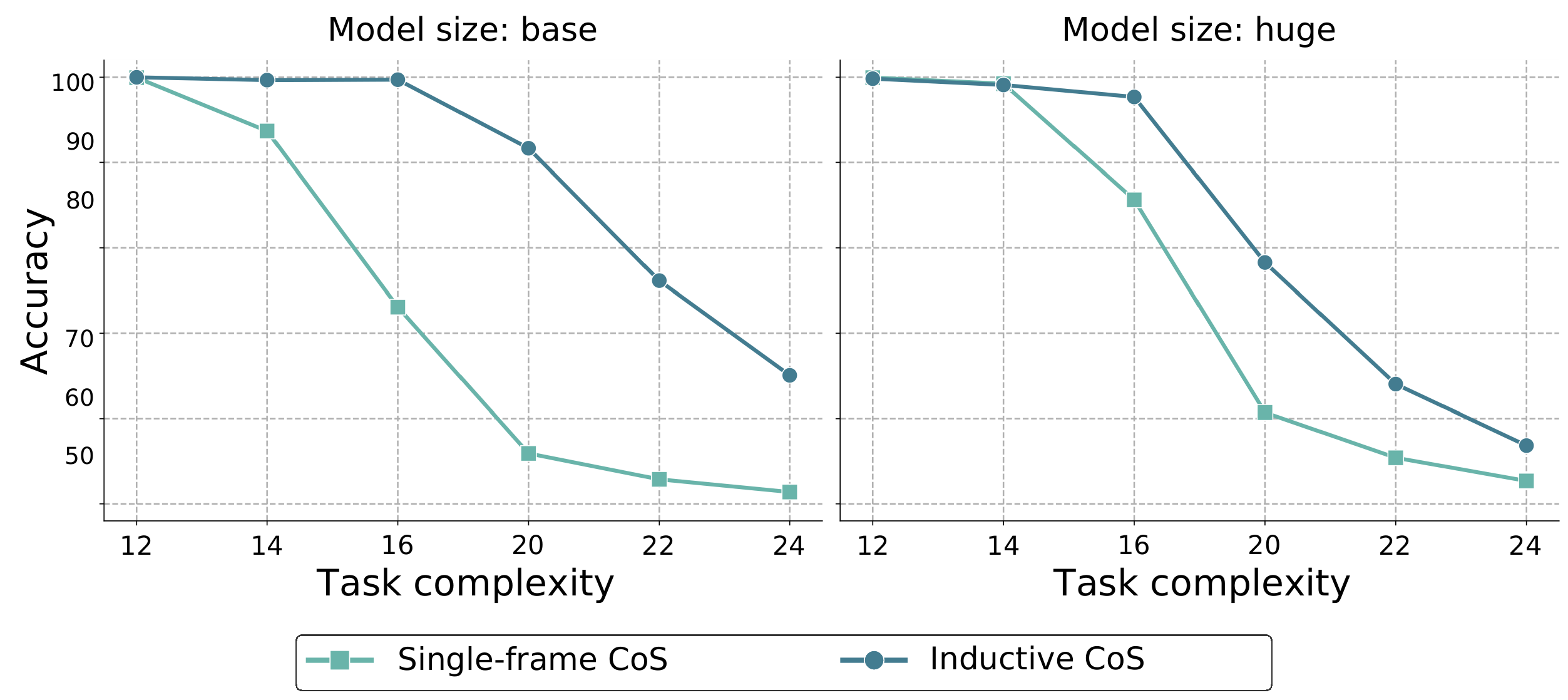}  
    \caption{OOD experiments with model size scaling}
    \label{fig:ood_scaling}
\end{figure}

The plot in Figure~\ref{fig:ood_scaling} presents the OOD generalization performance for models of different sizes (B and H) trained on task complexity 12 and tested on more complex tasks ranging from 14 to 24. Notably, the inductive CoS model consistently outperforms the single-frame CoS across the entire range of task complexities, irrespective of model size. This trend holds true for both the base and huge models, although the performance gap between the two approaches seems to decrease as model size increases. This suggests that the single-frame model can somewhat benefit from larger models. However, as shown in the main paper, a key advantage of the CoS lies in its ability to improve performance by using more compute at inference time, enabling smaller models to perform well. We hypothesize that the diminishing gap in performance with the huge model might be attributed to it being more data-hungry. Since the CoS model sees two steps per iteration for each sample, it may be more prone to memorization, suffering from the additional exposure to images during training.

\subsection{Additional OOD experiments on maze (circular) and strings datasets}
\begin{figure}[htb]
    \centering
    \begin{minipage}{0.56\textwidth}
         \centering
         \includegraphics[width=0.9\textwidth]{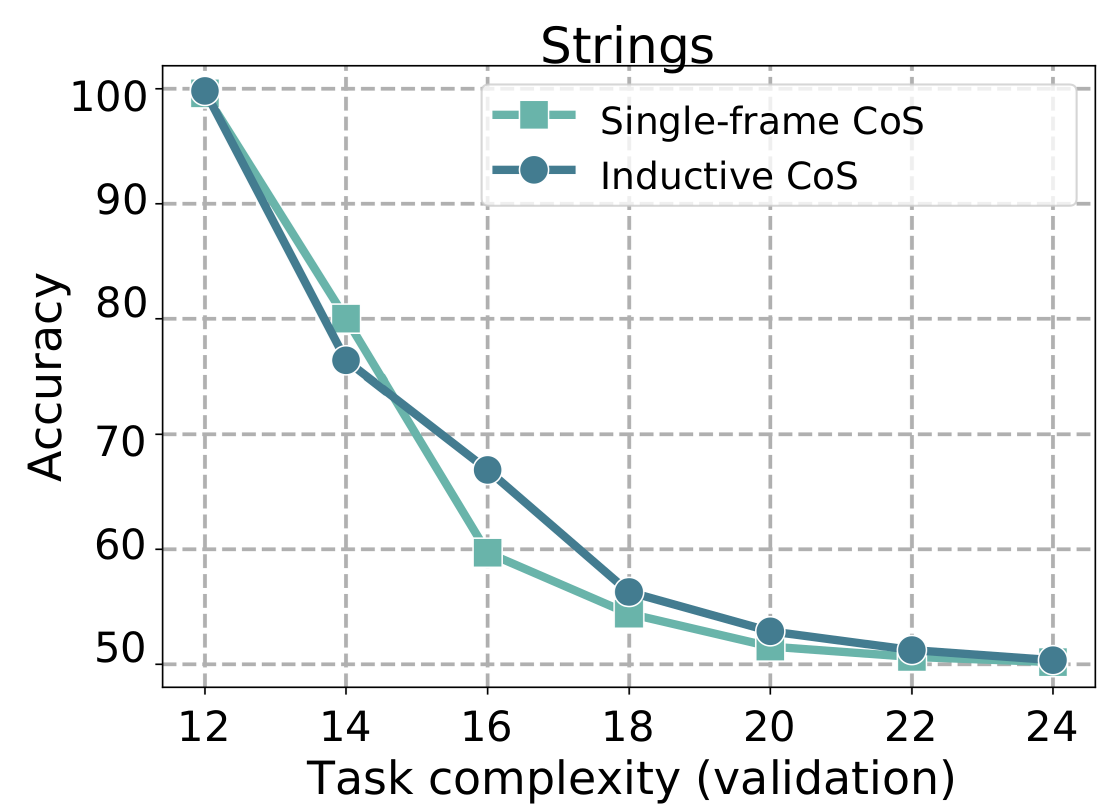}  
         \caption{OOD experiments where the model is trained on strings 12 and tested on more complex strings tasks.}
        \label{fig:ood_strings_12_12}
    \end{minipage}
    \hfill
    \begin{minipage}{0.43\textwidth}
        \centering
        \small
        \begin{tabular}{lcc}
        \toprule
        \multirow{2}{*}{\vspace{-0.5em}\textbf{Method}} & \multicolumn{2}{c}{\textbf{Accuracy (\%)}} \\
        \cmidrule(lr){2-3}
        & \textbf{ID} & \textbf{OOD} \\
        \midrule
        single-frame CoS & \textbf{100.00} & 62.99 \\
        inductive CoS    & 99.98  & \textbf{96.88} \\
        \bottomrule
        \end{tabular}
        \captionof{table}{In-distribution (ID) and out-of-distribution (OOD) performance on the maze (circular) dataset for different methods. The inductive CoS outperforms the single-frame model in the OOD setting.}
        \label{tab:ood_maze_circ}
    \end{minipage}
\end{figure}

Similar to the experiments presented in the main paper, on the maze (circular) dataset (see Table~\ref{tab:ood_maze_circ}), both the inductive and single-frame CoS achieve near-perfect performance on in-distribution (ID) tasks. However, for out-of-distribution (OOD) tasks, the inductive CoS model significantly outperforms the single-frame CoS, achieving 96.88\% accuracy compared to 62.99\%. This trend mirrors the results observed on the maze rectangular dataset, where OOD generalization is again much better for the inductive method. For the strings dataset (see Figure~\ref{fig:ood_strings_12_12}), the pattern slightly differs. The strings dataset is a more challenging dataset overall, as established in the main paper, which makes OOD generalization particularly difficult. Nonetheless, the CoS model consistently performs better than the single-frame CoS, especially on more complex OOD tasks, with the exception of size 14, the simplest OOD task in this setting.



\subsection{Additional ablations for the multi-frame CoS}\label{app:additional_ablations}
In the main paper, we discussed the factors contributing to the success of the inductive CoS model, including increased supervision during training, the halting mechanism, and the integration of teacher forcing with training on the output distribution. In this section, we present additional experiments to evaluate the impact of the multi-frame supervision. We introduced a multi-frame CoS model, which, while similar to the single-frame model, features multiple heads for predicting several CoS frames.

\begin{figure}[htb]
    \centering
    \includegraphics[width=0.5\textwidth]{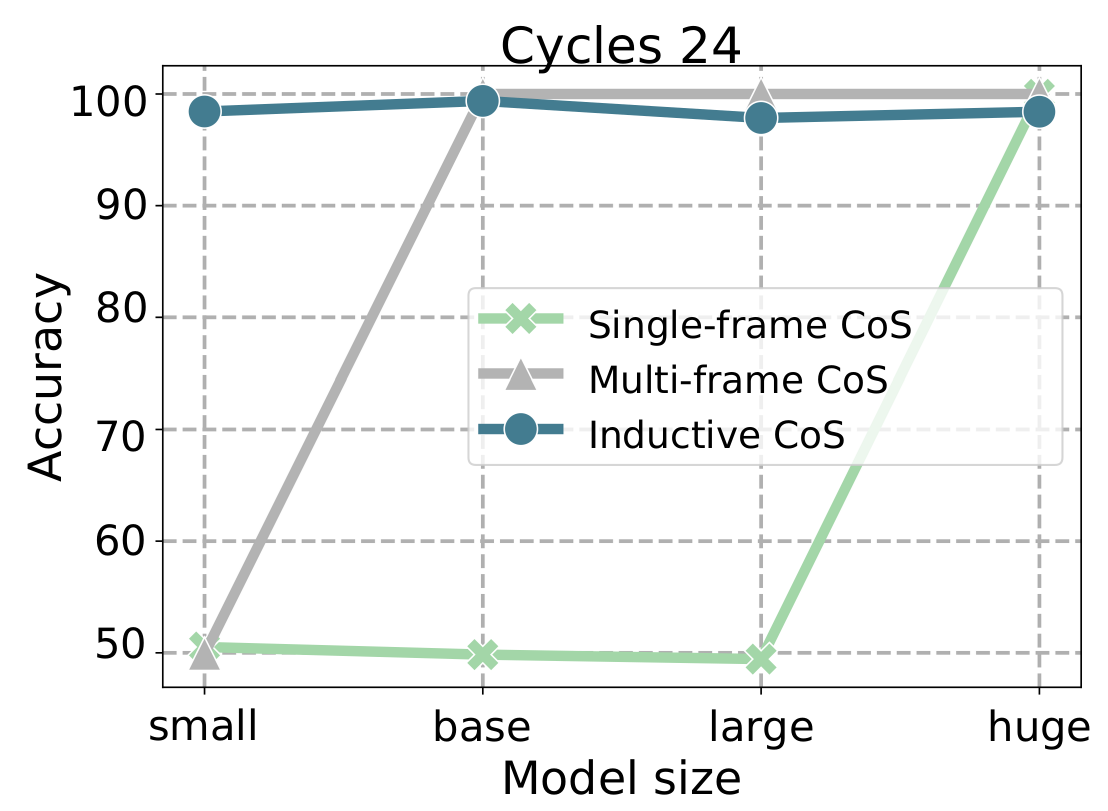}  
    \caption{Scaling parameters, single-frame vs. multi-frame vs. inductive CoS.}
    \label{fig:scaling_multi_frame}
\end{figure}

Our previous findings indicated that the multi-frame CoS did not yield any performance gains on OOD samples compared to the single-frame model. However, there is a scenario where the multi-frame approach proves beneficial: it aids convergence for smaller models (Base and Large). As illustrated in Figure~\ref{fig:scaling_multi_frame}, the inductive CoS converges across all model sizes (Small, Base, Large, and Huge), while the single-frame CoS only converges for the Huge model, indicating a greater computational demand to find the solution. The multi-frame model mitigates this issue by facilitating convergence in Base and Large models, suggesting that while it may still struggle with OOD, as noted in the main paper, it may help learning in-distribution. This improvement can be attributed to the presence of additional frames, which provide better guidance on the path to reaching the solution.


\section{Globality details}\label{app:globality-details}
Our main regime of interest, the regime with significant patches (which we studied in the main text), assumes that patches are large enough, e.g., $P \times P$ sized-patch with $P=\sqrt{n}$, such that only a unique ordering of the patches is valid (if one permutes the patches the new sample does not belong to the distribution's support w.h.p. as in many computer vision tasks of interest). In this case, the histogram part should not be inserted, as is done in our current definition. We used this assumption for defining globality degree in visual domain, which we consider to be the right regime to better understand the targets of interest.  In the small-patch regime where the patches have a constant size, i.e., $P \times P$ sized-patch with $P=O_n(1)$, one would also need to update slightly the definition of the globality degree in order to capture the fact that targets that are permutation invariant may be more easily learned by the Transformer after dropping the positional embeddings, which results in adding the histogram of the patches to the mutual information as was done in the globality degree for text by \citet{locality}.

 Further, note that to define our asymptotic quantities properly, we assume that the number of patches $n$ scales (e.g., as the size of the maze increases, we need a higher resolution image to solve it)\footnote{The number of patches in ViT models is usually constant as the images are resized unless the image is so fine-grained that a higher resolution is required, e.g., when the number of graph nodes diverges.}. In this case, the requirement would be to have $\alpha=n^{-O_n(1)}$ and $k^*=O_n(1)$ in order for the whole learning complexity to be polynomial in $n$. (Here by learning with polynomial complexity, we mean learning with polynomial model size in polynomial time.)

\section{Additional figures} \label{app:additional-figs}


\subsection{Chain-of-sketch examples for other tasks}\label{app:CoS-examples}
\begin{figure}[htb]
     \centering
     \hfill
     \begin{subfigure}[b]{0.16\textwidth}
         \centering
\includegraphics[width=\textwidth]{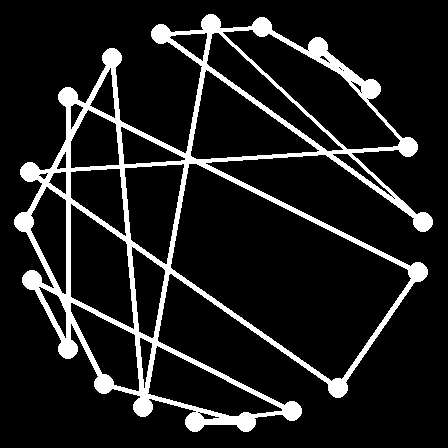}
     \end{subfigure}
     \hfill
     \begin{subfigure}[b]{0.16\textwidth}
         \centering
\includegraphics[width=\textwidth]{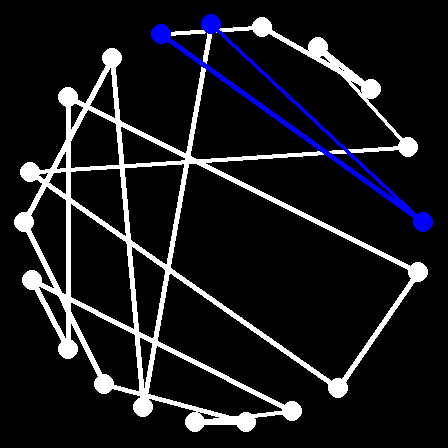}
     \end{subfigure}
     \hfill
     \begin{subfigure}[b]{0.16\textwidth}
         \centering
\includegraphics[width=\textwidth]{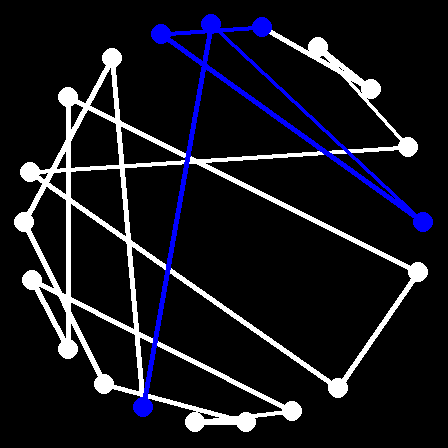}
     \end{subfigure}
     \hfill
     \begin{subfigure}[b]{0.16\textwidth}
         \centering
\includegraphics[width=\textwidth]{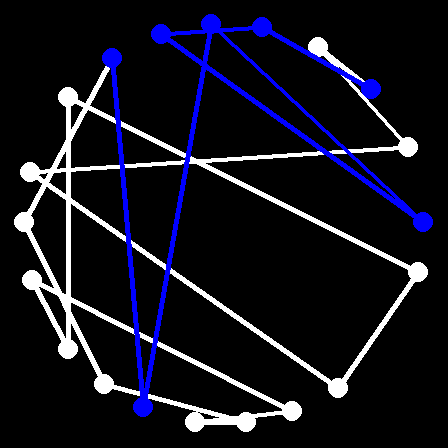}
     \end{subfigure}
     \hfill
     \begin{subfigure}[b]{0.16\textwidth}
         \centering
\includegraphics[width=\textwidth]{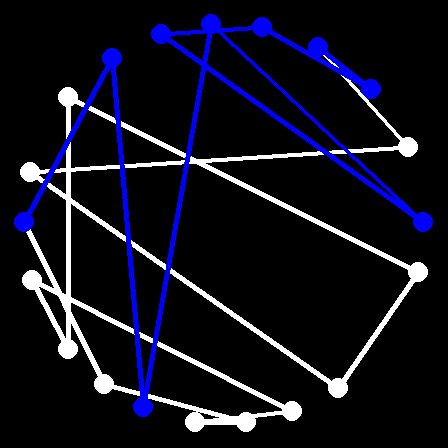}
     \end{subfigure}
     \hfill
     \begin{subfigure}[b]{0.16\textwidth}
         \centering
\includegraphics[width=\textwidth]{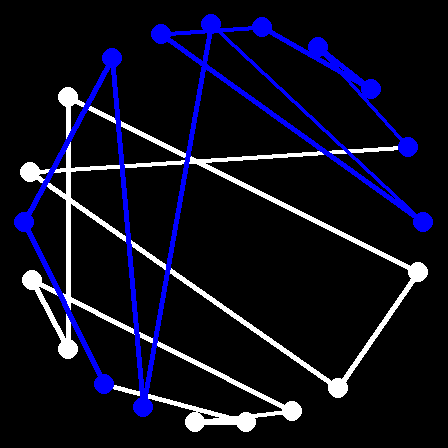}
     \end{subfigure}
     \hfill
     \begin{subfigure}[b]{0.16\textwidth}
         \centering
\includegraphics[width=\textwidth]{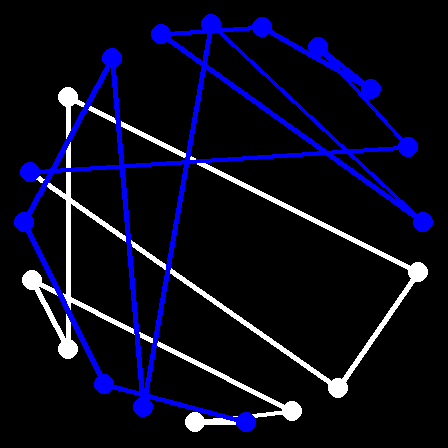}
     \end{subfigure}
     \hfill
     \begin{subfigure}[b]{0.16\textwidth}
         \centering
\includegraphics[width=\textwidth]{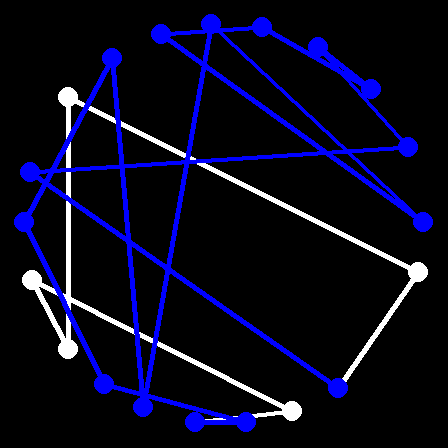}
     \end{subfigure}
     \hfill
     \begin{subfigure}[b]{0.16\textwidth}
         \centering
\includegraphics[width=\textwidth]{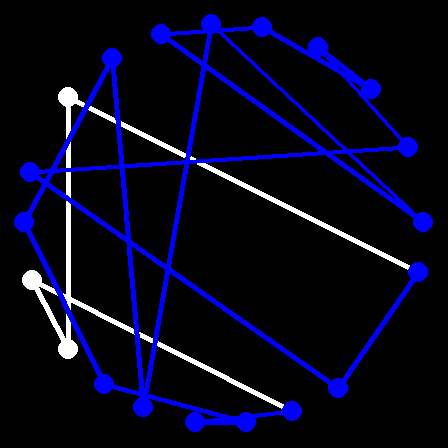}
     \end{subfigure}
     \hfill
     \begin{subfigure}[b]{0.16\textwidth}
         \centering
\includegraphics[width=\textwidth]{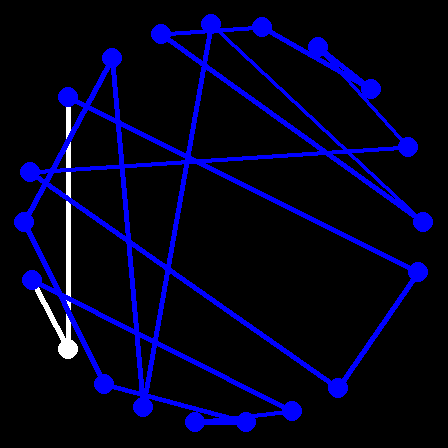}
     \end{subfigure}
     \hfill
     \begin{subfigure}[b]{0.16\textwidth}
         \centering
\includegraphics[width=\textwidth]{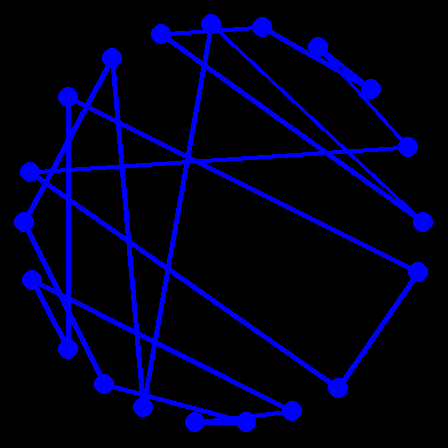}
     \end{subfigure}

    \vspace{-1mm}
    \caption{Example of sketches for the cycles 20 dataset, connected cycles.}
    \label{fig:scratchpad_cycles20}
    \vspace{-2mm}
\end{figure}

\begin{figure}[htb]
     \centering
     \hfill
     \begin{subfigure}[b]{0.20\textwidth}
         \centering
\includegraphics[width=\textwidth]{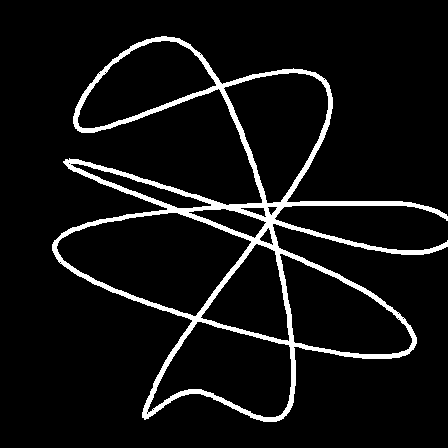}
     \end{subfigure}
     \hfill
     \begin{subfigure}[b]{0.20\textwidth}
         \centering
\includegraphics[width=\textwidth]{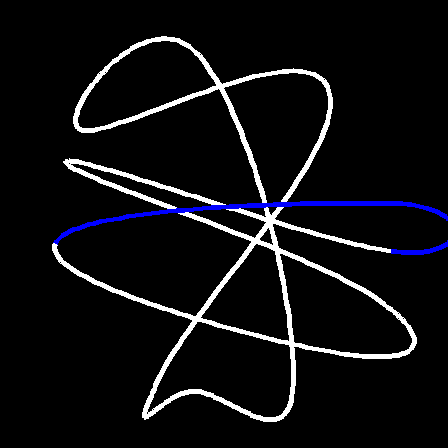}
     \end{subfigure}
     \hfill
     \begin{subfigure}[b]{0.20\textwidth}
         \centering
\includegraphics[width=\textwidth]{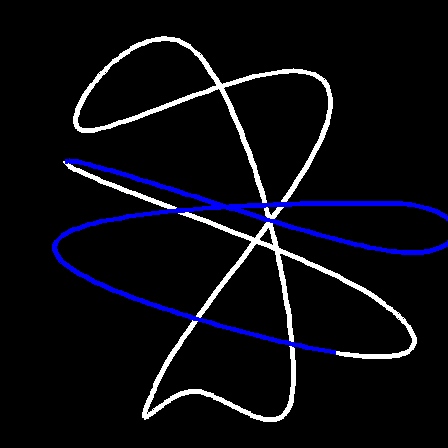}
     \end{subfigure}
     \hfill
     \begin{subfigure}[b]{0.20\textwidth}
         \centering
\includegraphics[width=\textwidth]{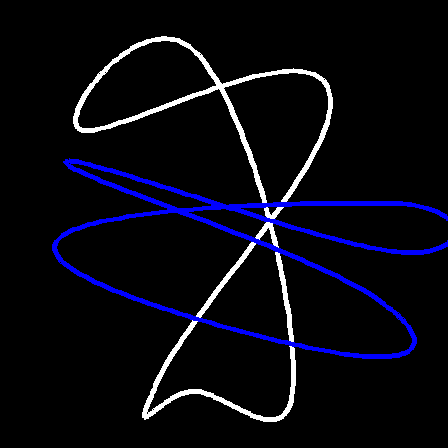}
     \end{subfigure}
    \vspace{-1mm}
    \caption{Example of sketches for the strings 12 dataset, disconnected strings.}
    \label{fig:scratchpad_strings12_disconnected}
    \vspace{-2mm}
\end{figure}

\begin{figure}[htb]
     \centering
     \hfill
     \begin{subfigure}[b]{0.13\textwidth}
         \centering
\includegraphics[width=\textwidth]{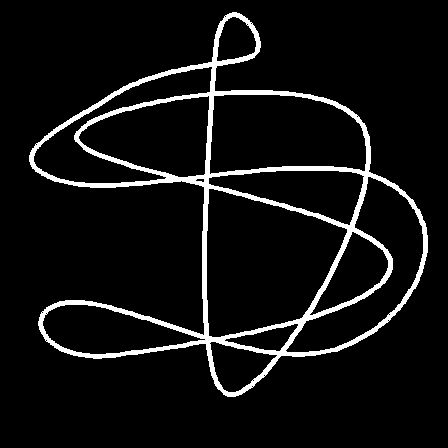}
     \end{subfigure}
     \hfill
     \begin{subfigure}[b]{0.13\textwidth}
         \centering
\includegraphics[width=\textwidth]{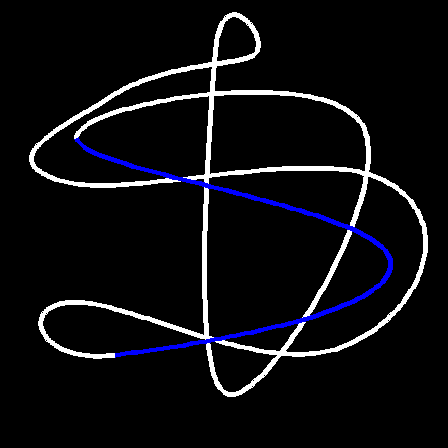}
     \end{subfigure}
     \hfill
     \begin{subfigure}[b]{0.13\textwidth}
         \centering
\includegraphics[width=\textwidth]{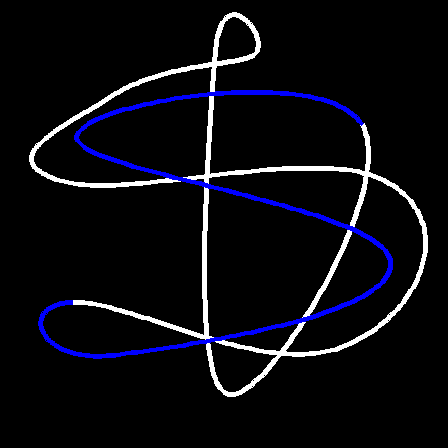}
     \end{subfigure}
     \hfill
     \begin{subfigure}[b]{0.13\textwidth}
         \centering
\includegraphics[width=\textwidth]{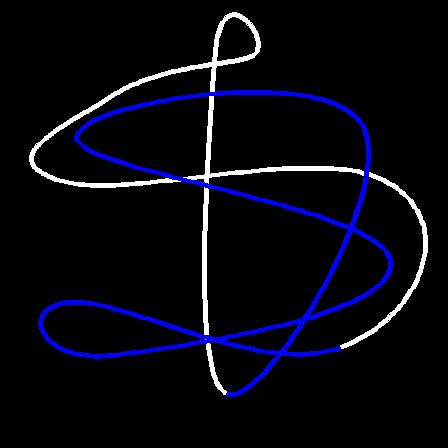}
     \end{subfigure}
     \hfill
     \begin{subfigure}[b]{0.13\textwidth}
         \centering
\includegraphics[width=\textwidth]{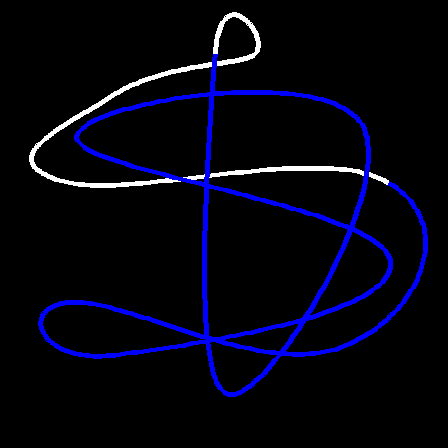}
     \end{subfigure}
     \hfill
     \begin{subfigure}[b]{0.13\textwidth}
         \centering
\includegraphics[width=\textwidth]{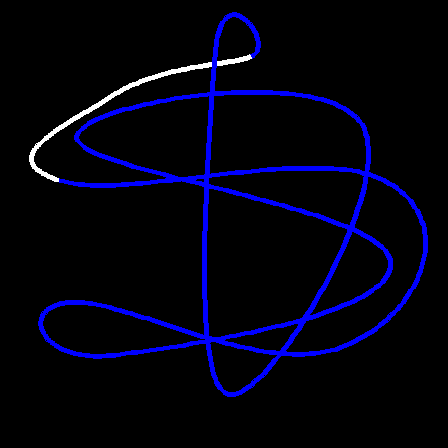}
     \end{subfigure}
     \hfill
     \begin{subfigure}[b]{0.13\textwidth}
         \centering
\includegraphics[width=\textwidth]{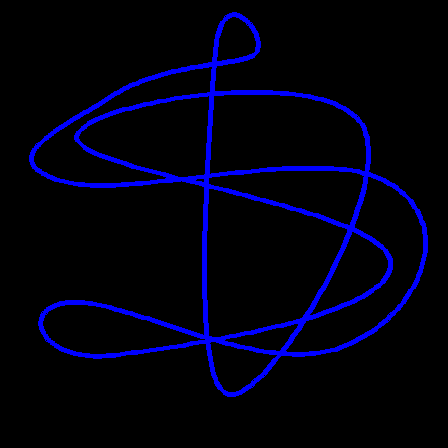}
     \end{subfigure}
    \vspace{-1mm}
    \caption{Example of sketches for the strings 12 dataset, connected strings.}
\label{fig:scratchpad_strings12_connected}
    \vspace{-2mm}
\end{figure}

\begin{figure}[htb]
     \centering
     \hfill
     \hfill
     \begin{subfigure}[b]{0.14\textwidth}
         \centering
\includegraphics[width=\textwidth]{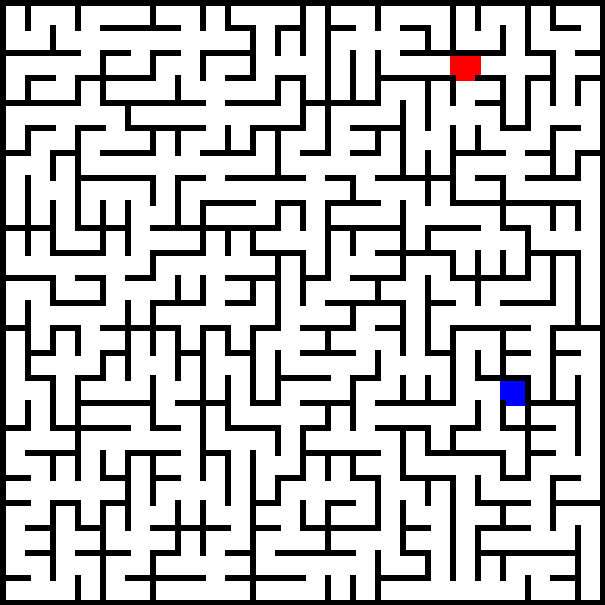}
     \end{subfigure}
     \hfill
     \begin{subfigure}[b]{0.14\textwidth}
         \centering
\includegraphics[width=\textwidth]{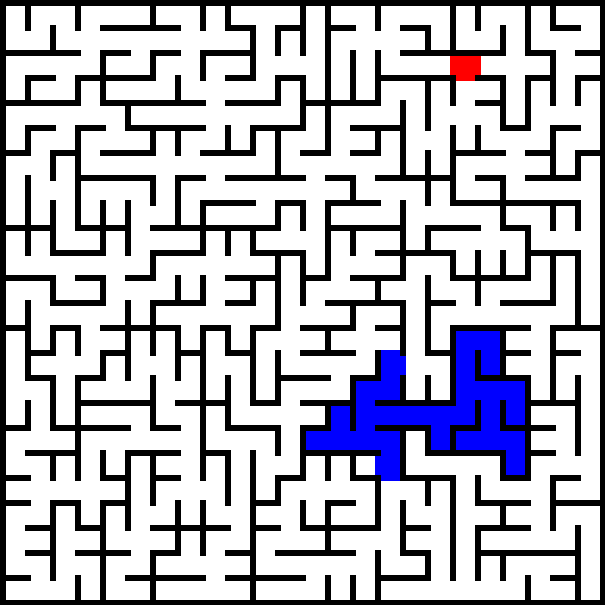}
     \end{subfigure}
     \hfill
     \begin{subfigure}[b]{0.14\textwidth}
         \centering
\includegraphics[width=\textwidth]{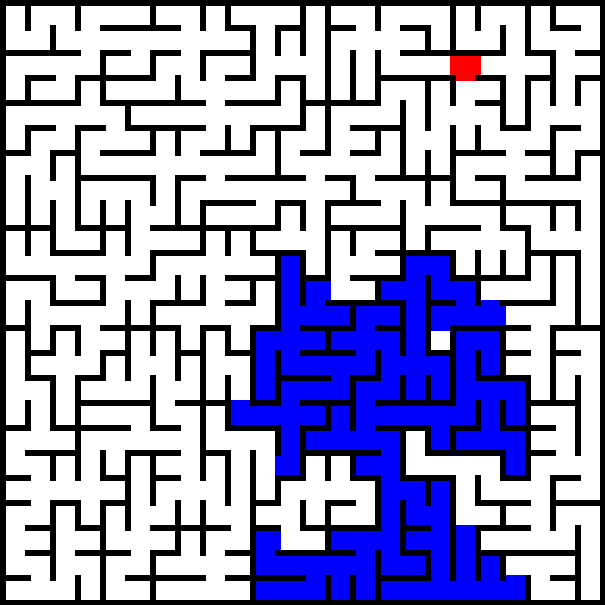}
     \end{subfigure}
     \hfill
     \begin{subfigure}[b]{0.14\textwidth}
         \centering
\includegraphics[width=\textwidth]{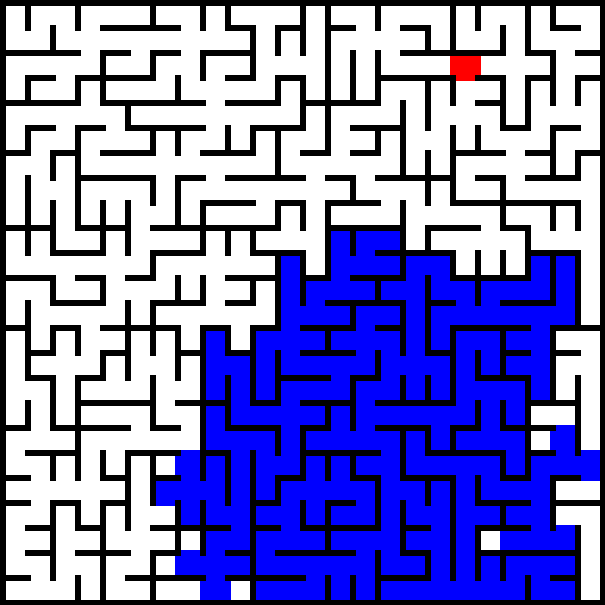}
     \end{subfigure}
     \hfill
     \begin{subfigure}[b]{0.14\textwidth}
         \centering
\includegraphics[width=\textwidth]{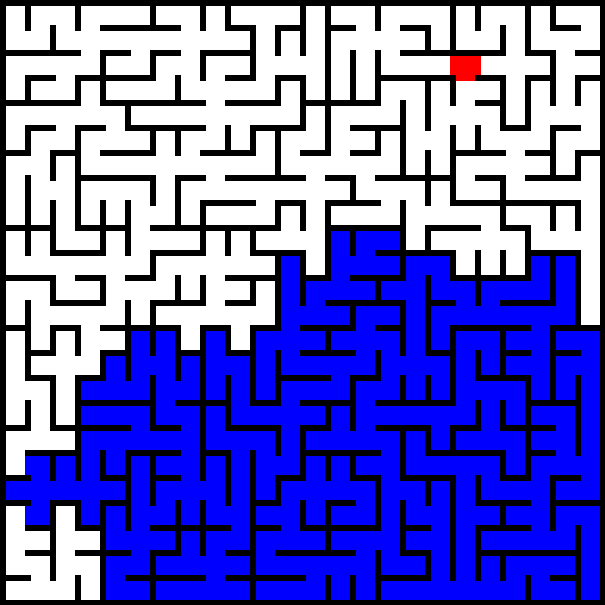}
     \end{subfigure}
     \hfill
     \begin{subfigure}[b]{0.14\textwidth}
         \centering
\includegraphics[width=\textwidth]{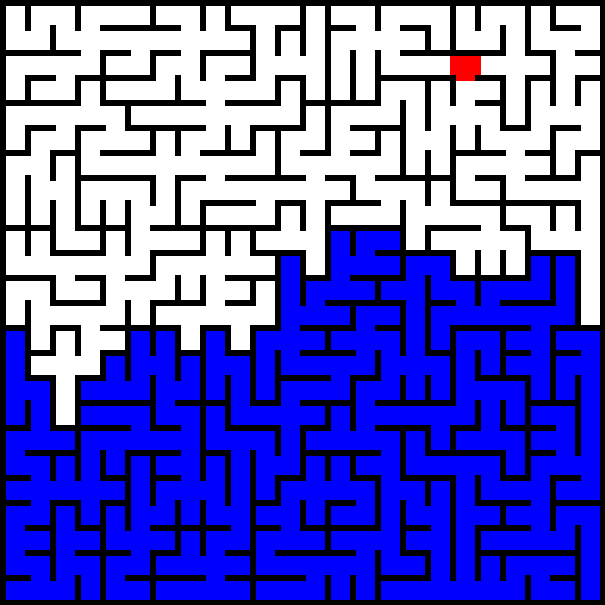}
     \end{subfigure}
    \vspace{-1mm}
    \caption{Example of sketches for the maze (rectangular) 24 dataset, non-solvable maze.}
\label{fig:scratchpad_mzr24_non_solvable}
    \vspace{-2mm}
\end{figure}

\begin{figure}[htb]
     \centering
     \hfill
     \hfill
     \begin{subfigure}[b]{0.14\textwidth}
         \centering
\includegraphics[width=\textwidth]{figures/examples/mzr24/1/000.png}
     \end{subfigure}
     \hfill
     \begin{subfigure}[b]{0.14\textwidth}
         \centering
\includegraphics[width=\textwidth]{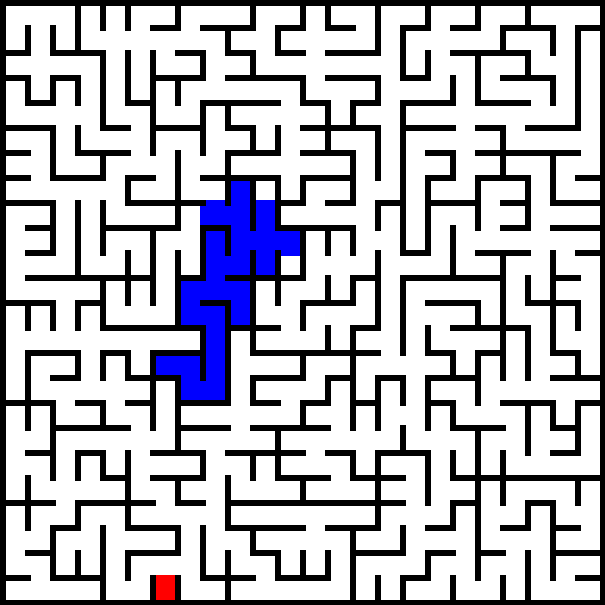}
     \end{subfigure}
     \hfill
     \begin{subfigure}[b]{0.14\textwidth}
         \centering
\includegraphics[width=\textwidth]{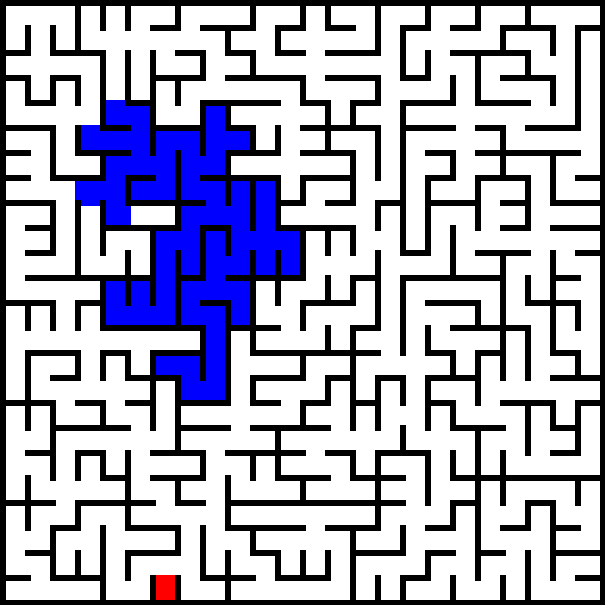}
     \end{subfigure}
     \hfill
     \begin{subfigure}[b]{0.14\textwidth}
         \centering
\includegraphics[width=\textwidth]{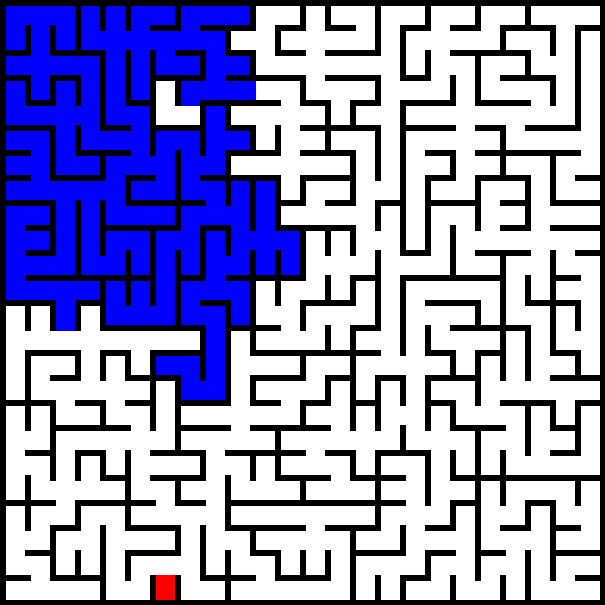}
     \end{subfigure}
     \hfill
     \begin{subfigure}[b]{0.14\textwidth}
         \centering
\includegraphics[width=\textwidth]{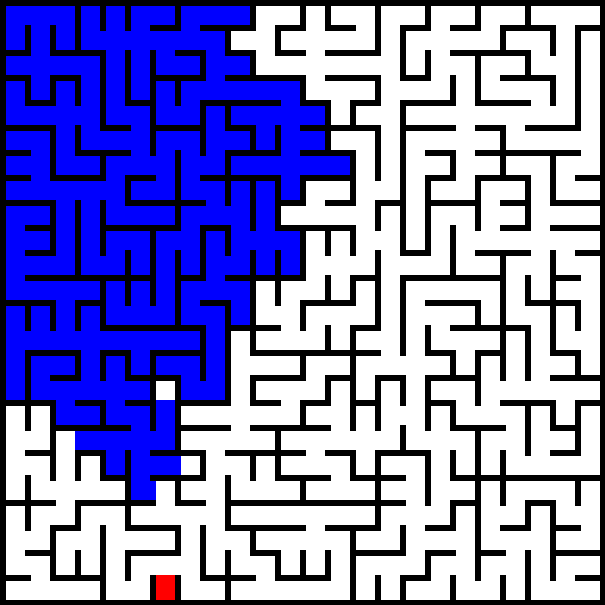}
     \end{subfigure}
     \hfill
     \begin{subfigure}[b]{0.14\textwidth}
         \centering
\includegraphics[width=\textwidth]{figures/examples/mzr24/1/005.png}
     \end{subfigure}
    \vspace{-1mm}
    \caption{Example of sketches for the maze (rectangular) 24 dataset, solvable maze.}
\label{fig:scratchpad_mzr24_solvable}
    \vspace{-2mm}
\end{figure}

\begin{figure}[htb]
     \centering
     \hfill
     \begin{subfigure}[b]{0.13\textwidth}
         \centering
\includegraphics[width=\textwidth]{figures/examples/mzc16/0/000.png}
     \end{subfigure}
     \hfill
     \begin{subfigure}[b]{0.13\textwidth}
         \centering
\includegraphics[width=\textwidth]{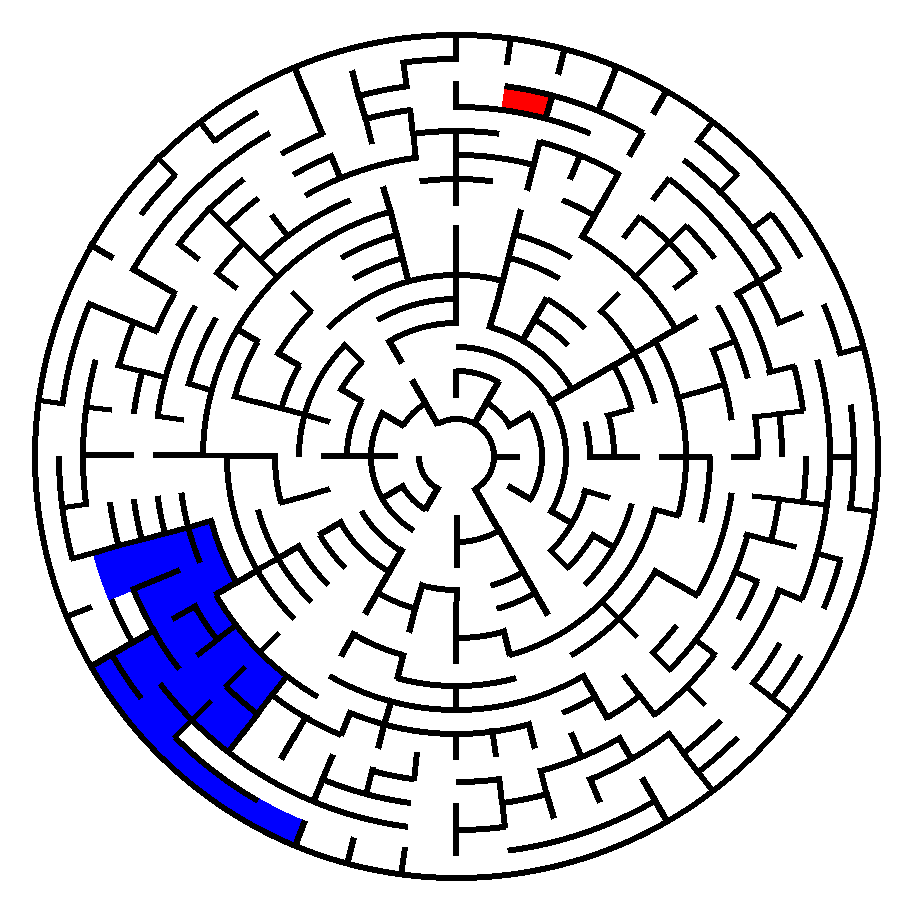}
     \end{subfigure}
     \hfill
     \begin{subfigure}[b]{0.13\textwidth}
         \centering
\includegraphics[width=\textwidth]{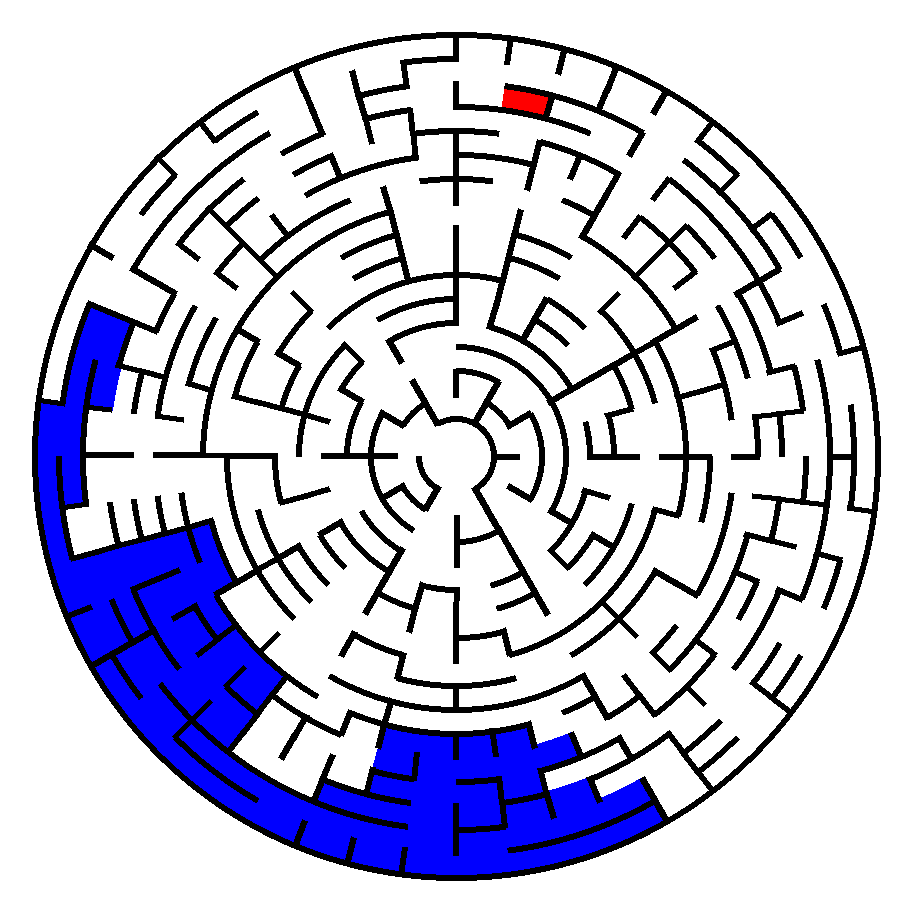}
     \end{subfigure}
     \hfill
     \begin{subfigure}[b]{0.13\textwidth}
         \centering
\includegraphics[width=\textwidth]{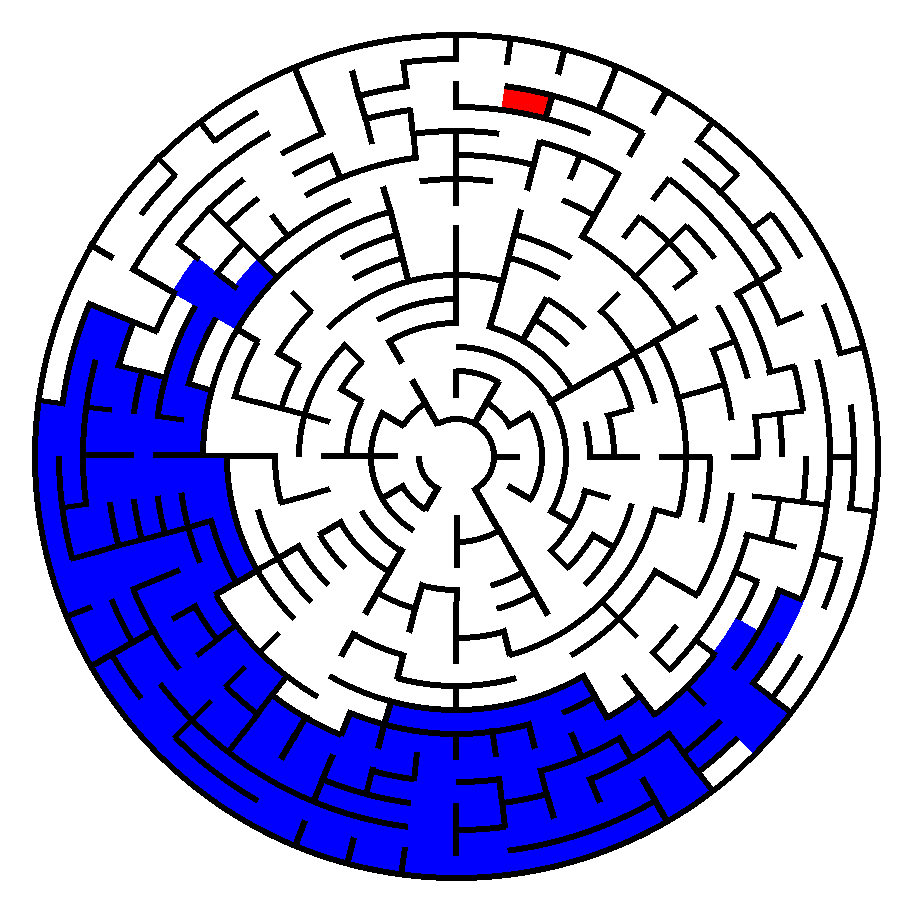}
     \end{subfigure}
     \hfill
     \begin{subfigure}[b]{0.13\textwidth}
         \centering
\includegraphics[width=\textwidth]{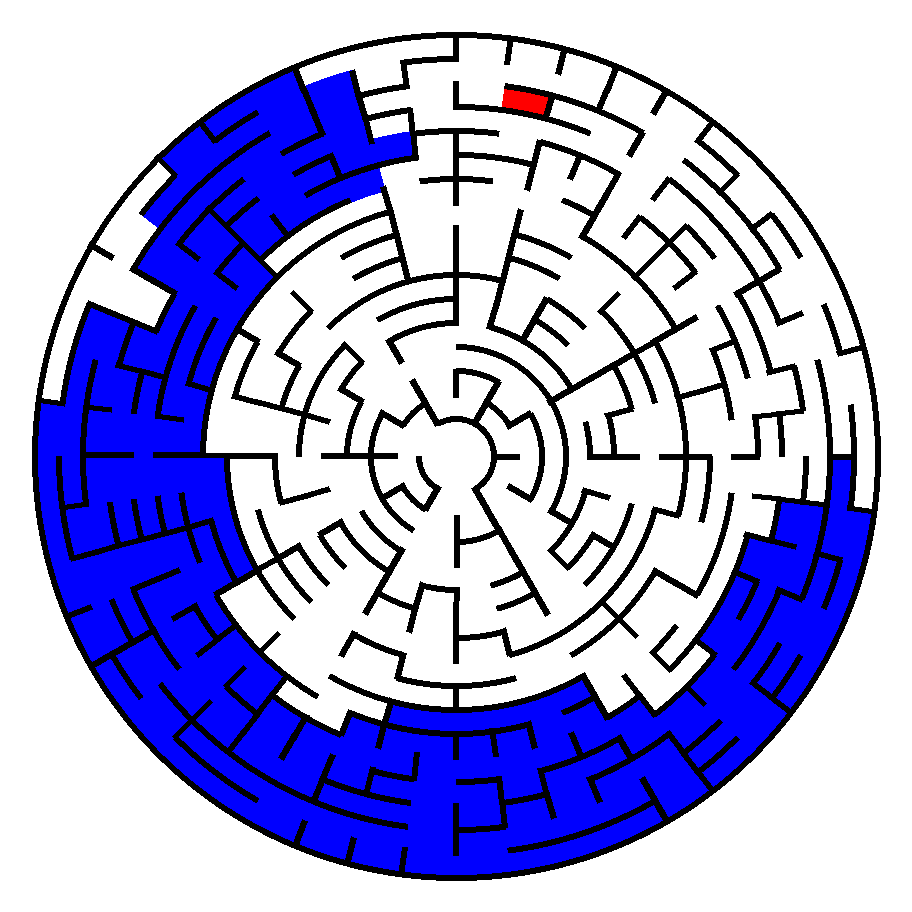}
     \end{subfigure}
     \hfill
     \begin{subfigure}[b]{0.13\textwidth}
         \centering
\includegraphics[width=\textwidth]{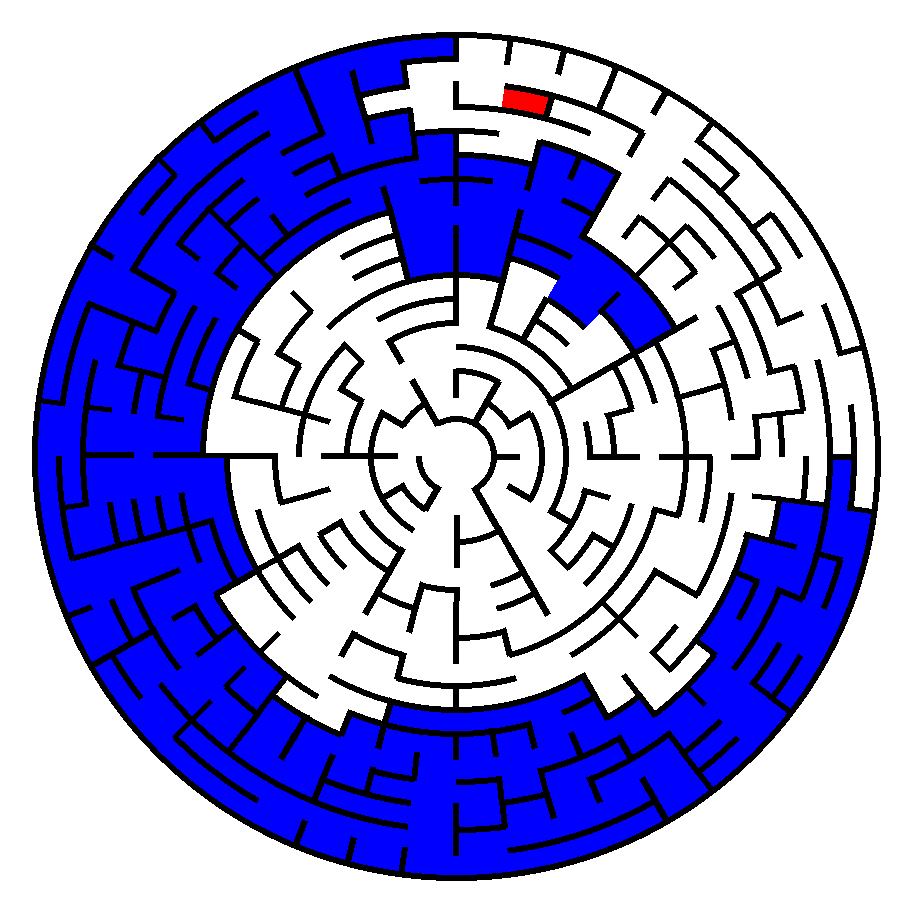}
     \end{subfigure}
     \hfill
     \begin{subfigure}[b]{0.13\textwidth}
         \centering
\includegraphics[width=\textwidth]{figures/examples/mzc16/0/006.png}
     \end{subfigure}
    \vspace{-1mm}
    \caption{Example of sketches for the maze (circular) 16 dataset, non-solvable maze maze.}
\label{fig:scratchpad_mzc16_non_solvable}
    \vspace{-2mm}
\end{figure}

\begin{figure}[htb]
     \centering
     \hfill
     \begin{subfigure}[b]{0.13\textwidth}
         \centering
\includegraphics[width=\textwidth]{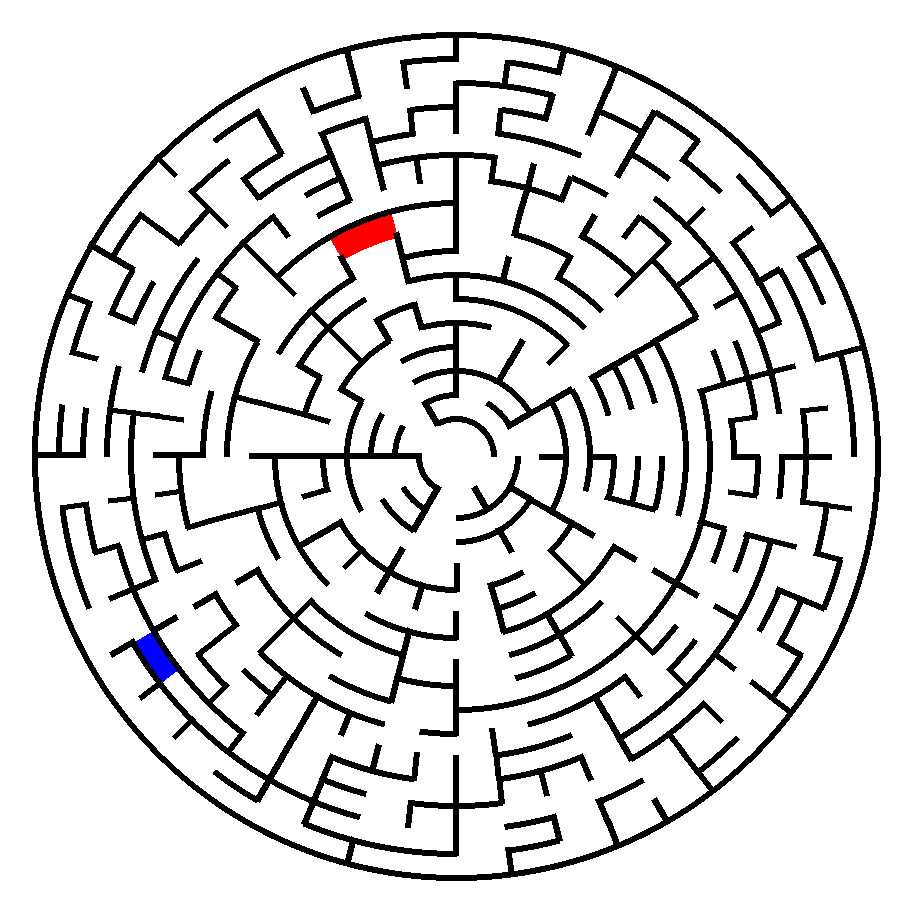}
     \end{subfigure}
     \hfill
     \begin{subfigure}[b]{0.13\textwidth}
         \centering
\includegraphics[width=\textwidth]{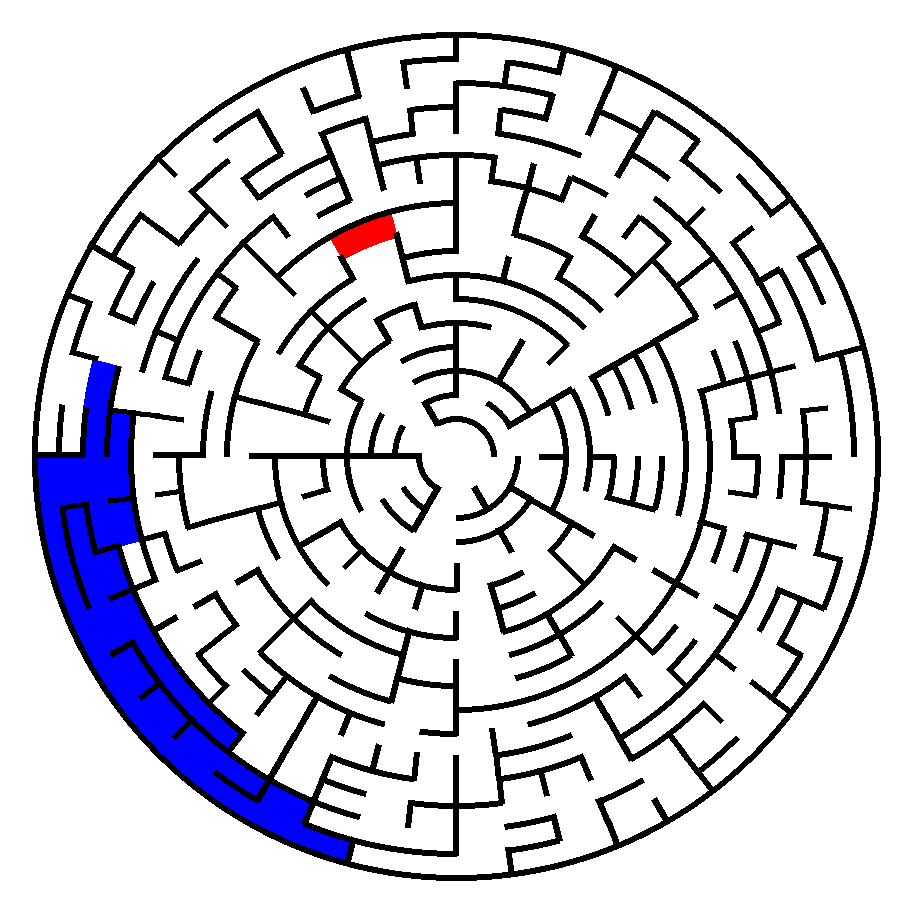}
     \end{subfigure}
     \hfill
     \begin{subfigure}[b]{0.13\textwidth}
         \centering
\includegraphics[width=\textwidth]{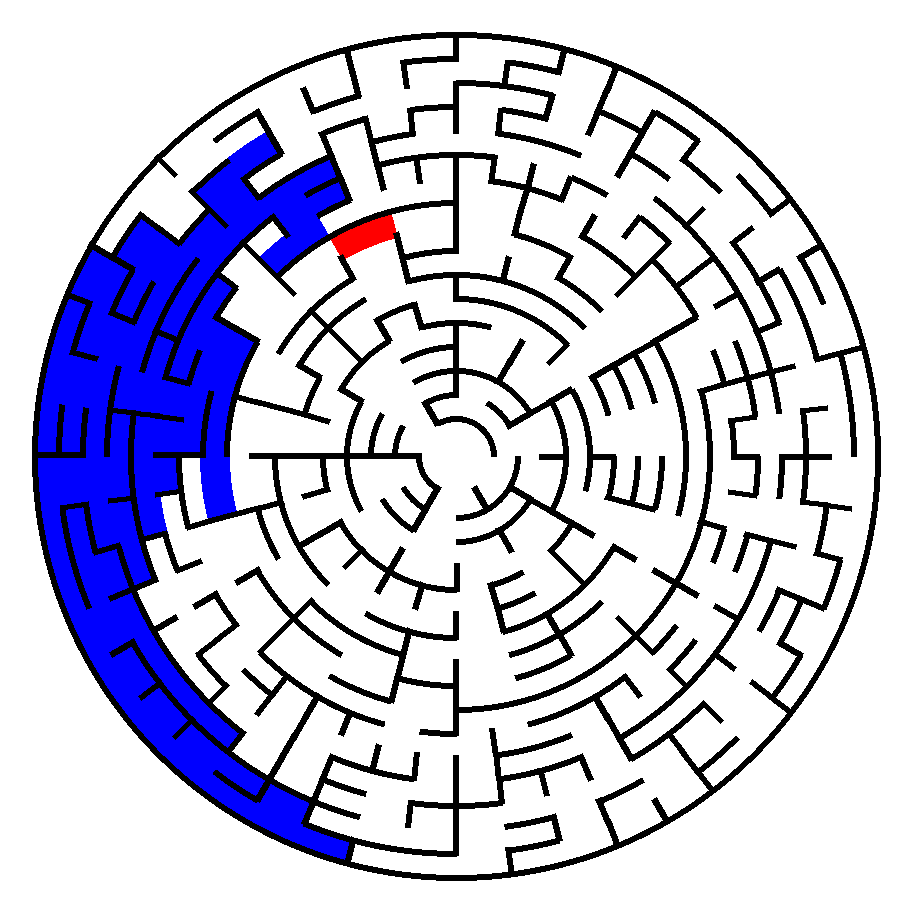}
     \end{subfigure}
     \hfill
     \begin{subfigure}[b]{0.13\textwidth}
         \centering
\includegraphics[width=\textwidth]{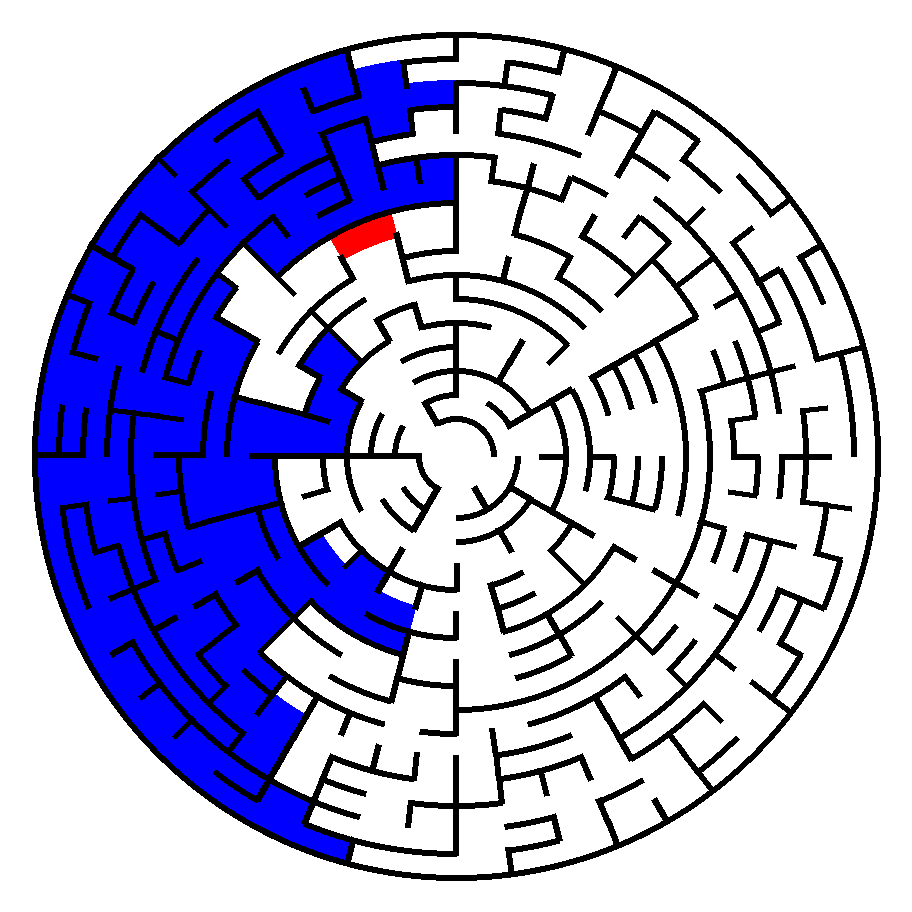}
     \end{subfigure}
     \hfill
     \begin{subfigure}[b]{0.13\textwidth}
         \centering
\includegraphics[width=\textwidth]{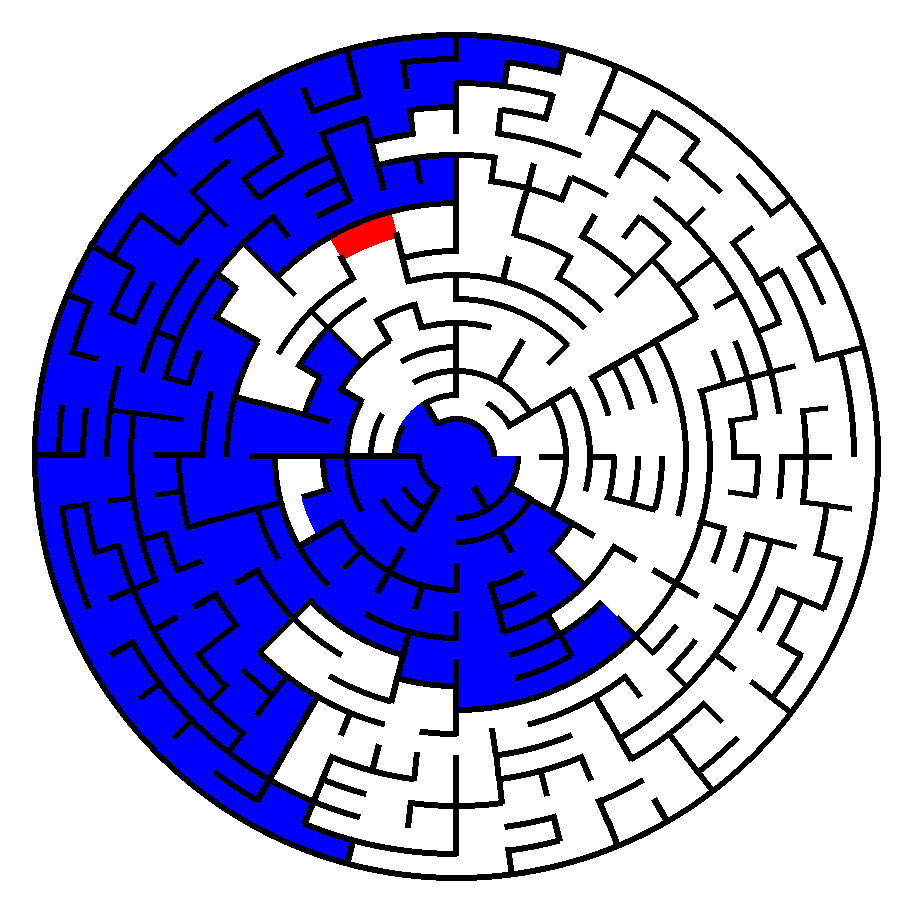}
     \end{subfigure}
     \hfill
     \begin{subfigure}[b]{0.13\textwidth}
         \centering
\includegraphics[width=\textwidth]{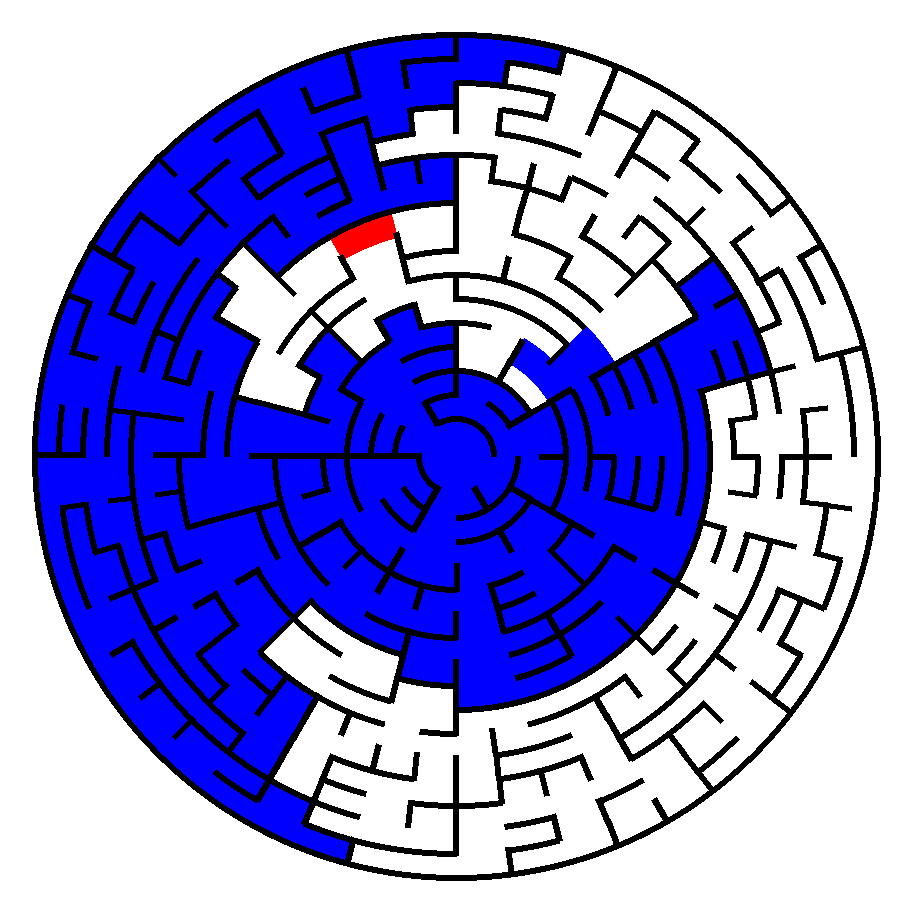}
     \end{subfigure}
     \hfill
     \begin{subfigure}[b]{0.13\textwidth}
         \centering
\includegraphics[width=\textwidth]{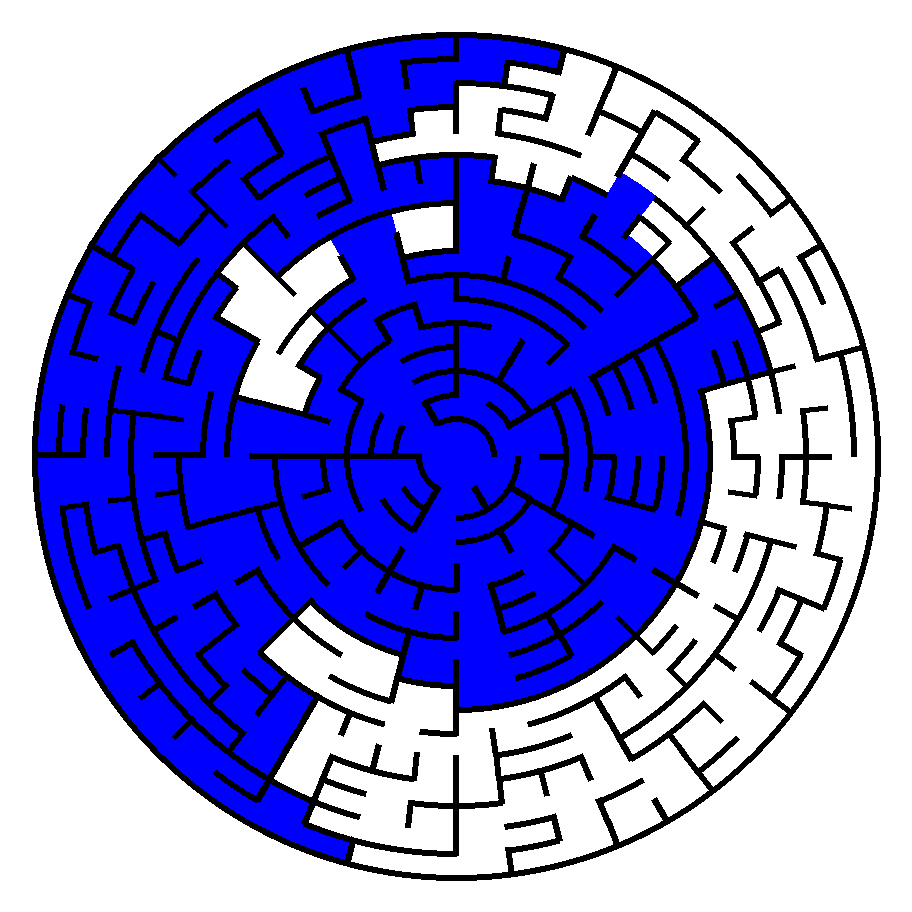}
     \end{subfigure}
    \vspace{-1mm}
    \caption{Example of sketches for the maze (circular) 16 dataset, solvable maze.}
\label{fig:scratchpad_mzc16_solvable}
    \vspace{-2mm}
\end{figure}

\begin{figure*}[htb]
     \centering
     \begin{subfigure}[b]{0.16\textwidth}
         \centering
         \includegraphics[width=\textwidth]{figures/examples/pvr4/0.png}
     \end{subfigure}
     \hfill
     \begin{subfigure}[b]{0.16\textwidth}
         \centering
         \includegraphics[width=\textwidth]{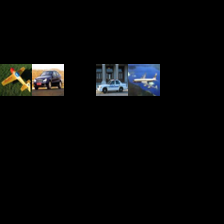}
     \end{subfigure}
     \hfill
     \begin{subfigure}[b]{0.16\textwidth}
         \centering
         \includegraphics[width=\textwidth]{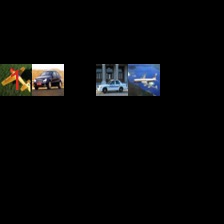}
     \end{subfigure}
     \hfill
     \begin{subfigure}[b]{0.16\textwidth}
         \centering
         \includegraphics[width=\textwidth]{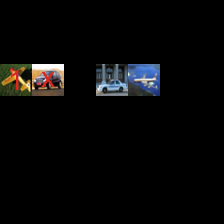}
     \end{subfigure}
     \hfill
     \begin{subfigure}[b]{0.16\textwidth}
         \centering
         \includegraphics[width=\textwidth]{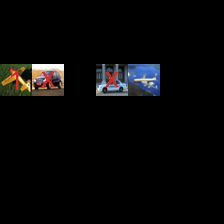}
     \end{subfigure}
     \hfill
     \begin{subfigure}[b]{0.16\textwidth}
         \centering
         \includegraphics[width=\textwidth]{figures/examples/pvr4/5.png}
     \end{subfigure}
    \vspace{-1mm}
    \caption{Example of sketches for PVR task on $7 \times 7$ grid and $k=4$ images per row where the parity function is used on the row. The final answer which is the parity of the airplane class in the indicated row is zero.}
    \label{fig:pvr-frames}
    \vspace{-2mm}
\end{figure*}

This section provides example chain-of-sketch instances for several tasks, demonstrating target frames for the model. The following figures illustrate chain-of-sketches for tasks like connected and disconnected cycles, strings, and solvable and non-solvable mazes. In Figure \ref{fig:scratchpad_cycles20}, we show examples of the cycles 20 dataset with connected cycles. In Figure \ref{fig:scratchpad_strings12_disconnected} and \ref{fig:scratchpad_strings12_connected}, sketches for the strings 12 dataset with disconnected and connected strings are shown. For maze tasks, Figures \ref{fig:scratchpad_mzr24_solvable} and \ref{fig:scratchpad_mzr24_non_solvable} display sketches for solvable and non-solvable rectangular mazes. Figures \ref{fig:scratchpad_mzc16_solvable} and \ref{fig:scratchpad_mzc16_non_solvable}, do the same for maze circular 16. Finally, Figure \ref{fig:pvr-frames} shows the sketches for a PVR task on $7 \times 7$ grid and $k=4$ images per row where the parity function is used on the row.

\clearpage
\subsection{Additional staircase figures}\label{app:add-staircase-figs}
\begin{figure}[htb]
     \centering
     \begin{subfigure}[b]{0.16\textwidth}
         \centering
         \includegraphics[width=\textwidth]{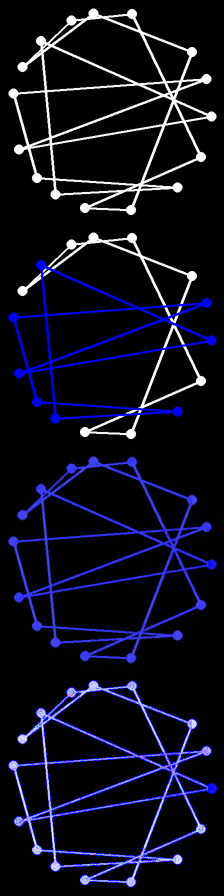}
        \caption{Iter = 2k}
     \end{subfigure}
     \hfill
      \begin{subfigure}[b]{0.16\textwidth}
     \centering
     \includegraphics[width=\textwidth]{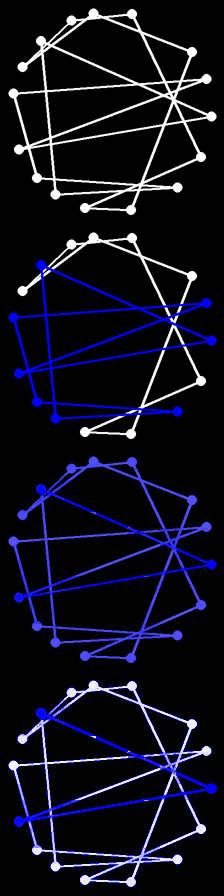}
     \caption{Iter = 6k}
     \end{subfigure}
     \hfill
     \begin{subfigure}[b]{0.16\textwidth}
         \centering
         \includegraphics[width=\textwidth]{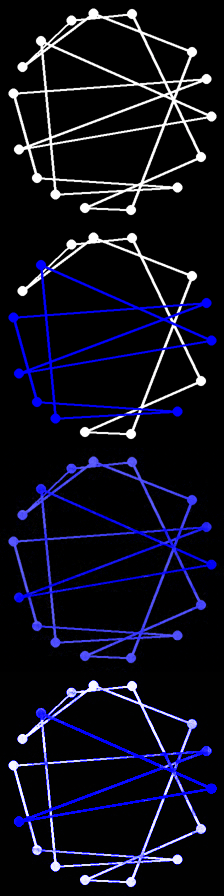}
         \caption{Iter = 8k}
     \end{subfigure}
     \hfill
     \begin{subfigure}[b]{0.16\textwidth}
         \centering
         \includegraphics[width=\textwidth]{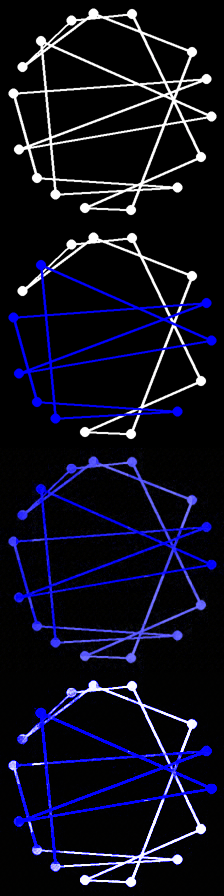}
         \caption{Iter = 10k}
     \end{subfigure}
     \hfill
     \begin{subfigure}[b]{0.16\textwidth}
         \centering
         \includegraphics[width=\textwidth]{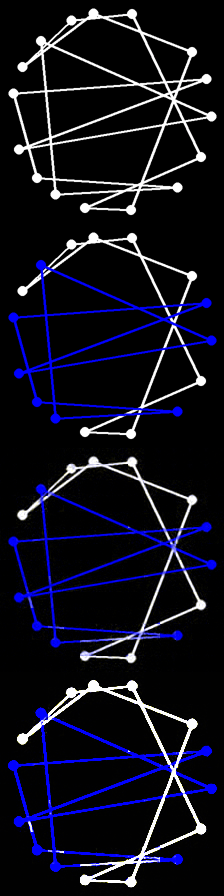}
         \caption{Iter = 13k}
     \end{subfigure}
     \hfill
     \begin{subfigure}[b]{0.16\textwidth}
         \centering
         \includegraphics[width=\textwidth]{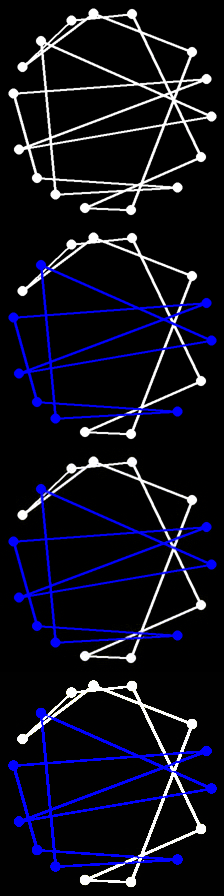}
         \caption{Iter = 50k}
     \end{subfigure}
    \caption{Expanded staircase examples for the cycles 16 task. The first row and second row represent the input image and the final ground truth CoS frame respectively. The third image represents the output of the model and the fourth image is similar to the third image with a higher contrast.}
    \label{fig:staircase_cycles_full}
\end{figure}

\begin{figure}[htb]
     \centering
     \begin{subfigure}[b]{0.16\textwidth}
         \centering
         \includegraphics[width=\textwidth]{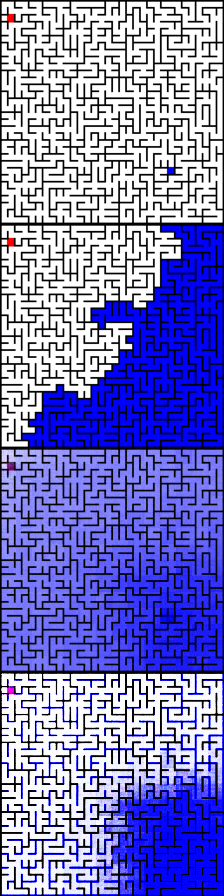}
         \caption{Iter = 3k}
     \end{subfigure}
     \hfill
      \begin{subfigure}[b]{0.16\textwidth}
     \centering
     \includegraphics[width=\textwidth]{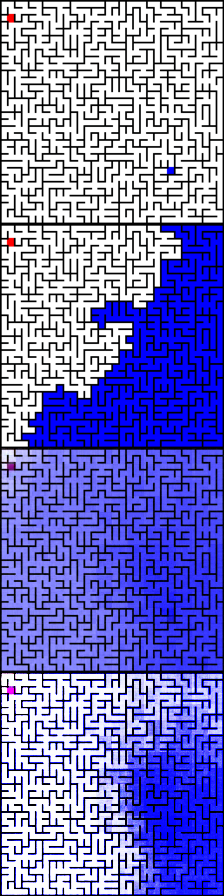}
     \caption{Iter = 5k}
     \end{subfigure}
     \hfill
     \begin{subfigure}[b]{0.16\textwidth}
         \centering
         \includegraphics[width=\textwidth]{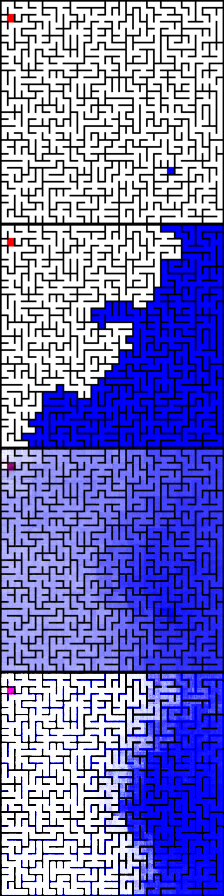}
         \caption{Iter = 6k}
     \end{subfigure}
     \hfill
     \begin{subfigure}[b]{0.16\textwidth}
         \centering
         \includegraphics[width=\textwidth]{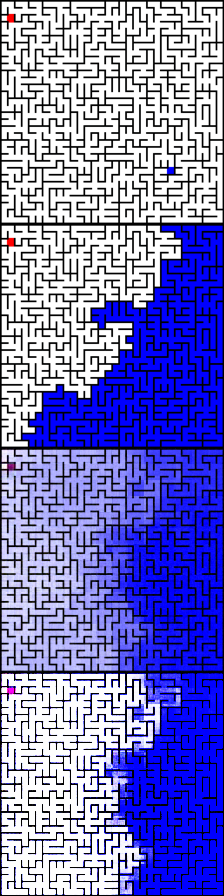}
         \caption{Iter = 8k}
     \end{subfigure}
     \hfill
     \begin{subfigure}[b]{0.16\textwidth}
         \centering
         \includegraphics[width=\textwidth]{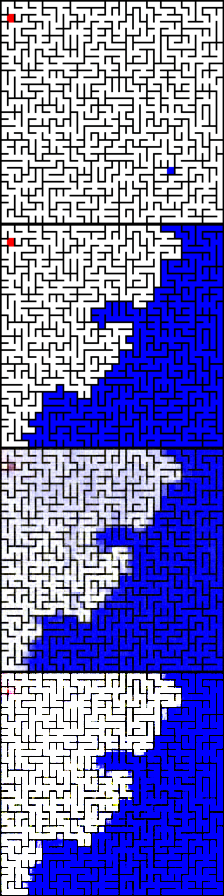}
         \caption{Iter = 11k}
     \end{subfigure}
     \hfill
     \begin{subfigure}[b]{0.16\textwidth}
         \centering
         \includegraphics[width=\textwidth]{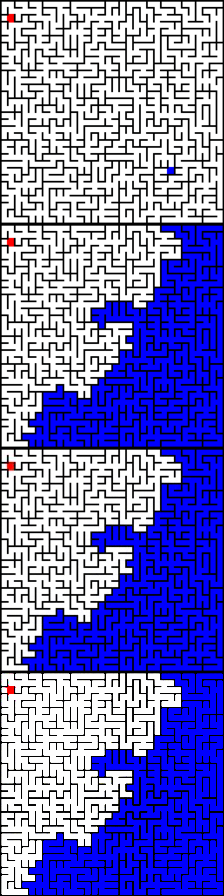}
         \caption{Iter = 50k}
     \end{subfigure}
    \caption{Additional staircase example for the maze (rectangular) task. The first row and second row show the input image and the final ground truth CoS frame respectively. The third image represents the output of the model and the fourth image is similar to the third image with a higher contrast.}
    \label{fig:staircase_maze}
\end{figure}

Figures \ref{fig:staircase_cycles_full} and \ref{fig:staircase_maze} present the staircase hierarchical learning phenomenon for cycles and maze datasets. In each figure, the first row presents the input image, the second row presents the ground truth (single-frame) CoS. The third row presets the model output (at a certain training iteration), and finally, the fourth row presents the same output with increased contrast for extra clarity. 

\clearpage
\subsection{CoS generations of the models}
\begin{figure}[htbp]
    \centering
    \includegraphics[width=\linewidth]{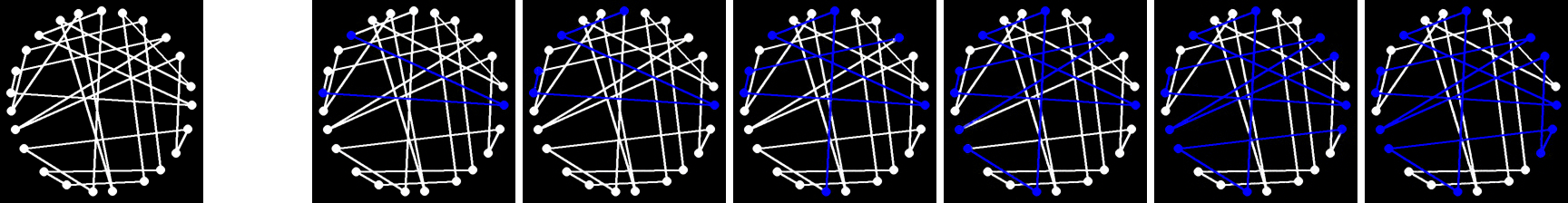}
    \caption{Generated CoS frames by our inductive CoS model for an example from the cycles task. The input is shown on the left, followed by the generated frames from left to right.}
    \label{fig:gen-cyc}
\end{figure}

\begin{figure}[htbp]
    \centering
    \includegraphics[width=\linewidth]{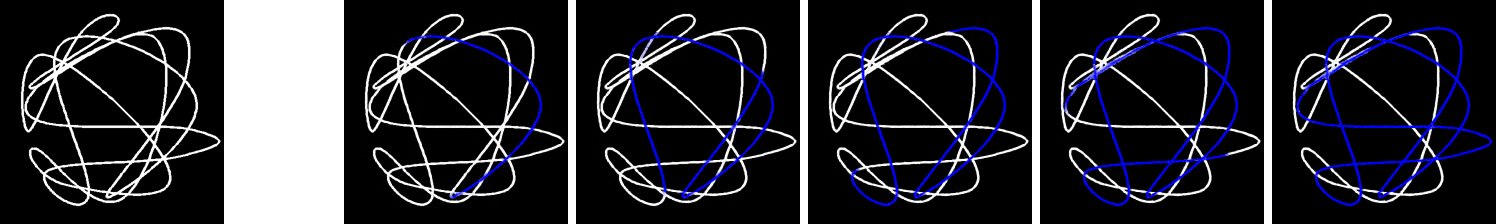}
    \caption{Generated CoS frames by our inductive CoS model for a strings task example. Input is shown on the left, followed by the generated frames are shown from left to right.}
    \label{fig:gen-str}
\end{figure}

\begin{figure}[htbp]
    \centering
    \includegraphics[width=\linewidth]{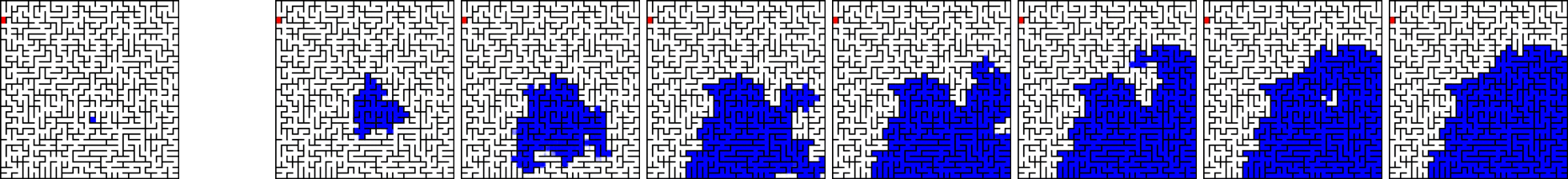}
    \caption{Generated CoS frames by our inductive CoS model for a maze (rect.) task example. Input is shown on the left, followed by the generated frames are shown from left to right.}
    \label{fig:gen-mzr}
\end{figure}

\begin{figure}[htbp]
    \centering
    \includegraphics[width=\linewidth]{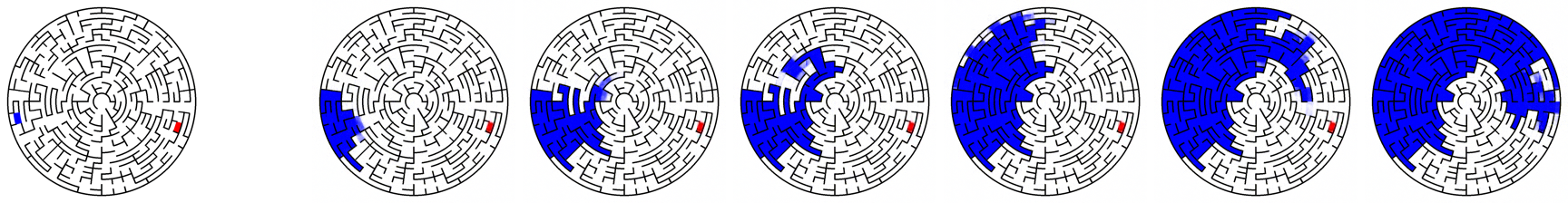}
    \caption{Generated CoS frames by our inductive CoS model for a maze (circ.) task example. Input is shown on the left, followed by the generated frames are shown from left to right.}
    \label{fig:gen-mzc}
\end{figure}

\begin{figure}[htbp]
    \centering
    \includegraphics[width=\linewidth]{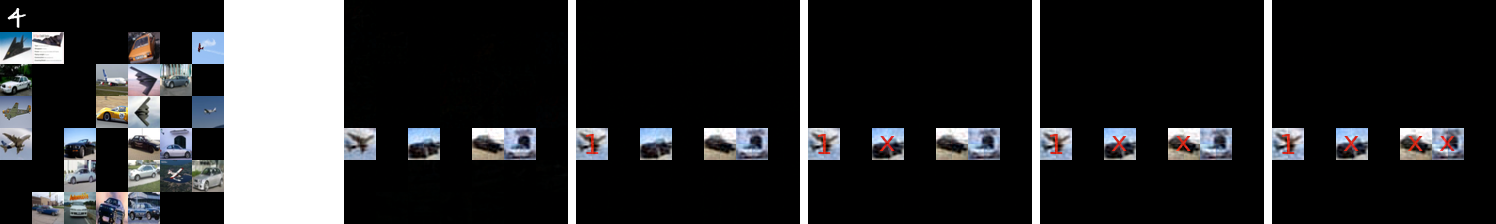}
    \caption{Generated CoS frames by our inductive CoS model for a PVR task example. Input is shown on the left, followed by the generated frames are shown from left to right.}
    \label{fig:gen-pvr}
\end{figure}

All of our CoS models that were large enough to learn the tasks were also able to generate the CoS frame(s) with a rather high quality. For reference, here we report some of the sketches that our inductive CoS model generated for different tasks in Figures \ref{fig:gen-cyc} to \ref{fig:gen-pvr}. 

\end{document}